\documentclass{article}

\usepackage[linesnumbered,vlined,ruled]{algorithm2e} 
\usepackage[bookmarksopen=true,bookmarksopenlevel=5,pagebackref]{hyperref}
\usepackage[a4paper, total={16cm, 26cm}]{geometry}
\usepackage[utf8]{inputenc}
\usepackage{amsmath}
\usepackage{amstext}
\usepackage{amssymb}
\usepackage{adjustbox}
\usepackage{bm,array}
\usepackage{tabularx}
\usepackage{booktabs}
\usepackage{multirow}
\usepackage{colortbl}
\usepackage[table]{xcolor}
\usepackage{graphicx,dblfloatfix}

\hypersetup
{
	pdftitle={POSNoise: An Effective Countermeasure Against Topic Biases in Authorship Analysis},
	pdfauthor={Oren Halvani and Lukas Graner},
	pdfsubject={Authorship Verification},
	pdfcreator={},
	pdfproducer={},
	pdfkeywords={Authorship analysis, topic masking, bias mitigation, preprocessing},
	pdfpagemode=UseNone,
	pdfstartview={XYZ null null null},
	colorlinks = true,
	linkcolor  = blue,
	citecolor  = blue,
	urlcolor  = blue,
	anchorcolor = blue
}

\newcommand*{\A}            {\mathcal{A}}

\newcommand*{\unknown}      {\mathcal{U}}

\newcommand*{\D}            {\mathcal{D}}
\newcommand*{\DA}           {{\D_{\A}}}

\newcommand*{\Dunk}         {\D_{\,\unknown}}

\newcommand*{\Dset}         {\mathbb{D}}

\newcommand*{\numberOfExistingBaselines}        {six\xspace}

\newcommand*{\Corpus}       {\mathcal{C}}
\newcommand*{\CorpusTrain}  {\Corpus_{\textrm{Train}}}
\newcommand*{\CorpusTest}   {\Corpus_{\textrm{Test}}}

\newcommand*{\CorpusACL}       {\Corpus_\mathrm{ACL}}
\newcommand*{\CorpusApricity}  {\Corpus_\mathrm{Apric}}

\newcommand*{\CorpusTelegraph} {\Corpus_\mathrm{Tel}}
\newcommand*{\CorpusReddit}    {\Corpus_\mathrm{Reddit}}

\newcommand*{\CorpusPeeJ}       {\Corpus_\mathrm{Perv}}
\newcommand*{\CorpusGutenberg}  {\Corpus_\mathrm{Gut}}
\newcommand*{\CorpusWikiSockpuppets}  {\Corpus_\mathrm{Wiki}}

\newcommand*{\CorpusAGNews} {\Corpus_\mathrm{AG}}
\newcommand*{\CorpusYahoo}  {\Corpus_\mathrm{Yahoo}}
\newcommand*{\CorpusBBC}    {\Corpus_\mathrm{BBC}}

\newcommand*{\Arefset}      {\Dset_{\A}}

\newcommand*{\Problem}      {\ensuremath{c}\xspace}  

\newcommand{\List}[1]
{
	\ifx&#1&%
	{\mathcal{L}}
	\else
	{\mathcal{L}_{#1}}
	\fi       
}

\newcommand*{\listStylePatterns}    {\List{}} 

\newcommand*{\coav}          {\textsf{COAV}\mbox{}\xspace}
\newcommand*{\occav}         {\textsf{OCCAV}\mbox{}\xspace}
\newcommand*{\stamatatosProf}{\textsf{ProfileAV}\mbox{}\xspace}
\newcommand*{\veenmanNNCD}   {\textsf{NNCD}\mbox{}\xspace}
\newcommand*{\spatium}       {\textsf{SPATIUM}\mbox{}\xspace}
\newcommand*{\koppelUnmask}  {\textsf{Unmasking}\mbox{}\xspace}

\renewcommand*{\AA}     {AA\mbox{}\xspace}

\renewcommand*{\aa}     {authorship attribution\mbox{}\xspace}

\newcommand*{\AV}       {AV\mbox{}\xspace}

\newcommand*{\av}       {authorship verification\mbox{}\xspace}

\newcommand*{\classY}        {\texttt{Y}\mbox{}\xspace} 
\newcommand*{\classN}        {\texttt{N}\mbox{}\xspace} 
\newcommand*{\classYdash}[1] {\texttt{Y-}#1\mbox{}\xspace} 
\newcommand*{\classNdash}[1] {\texttt{N-}#1\mbox{}\xspace} 

\newcommand*{\unanswered}  {\texttt{U}\mbox{}\xspace}

\newcommand*{\auc}         {AUC\mbox{}\xspace}

\newcommand*{\stateOfTheArt}  {state of the art\mbox{}\xspace}

\newcommand*{\posTag}         {POS tag\mbox{}\xspace}
\newcommand*{\posTags}        {POS tags\mbox{}\xspace}

\newcommand*{\charNgrams}     {character $n$\texttt{-}grams\mbox{}\xspace}

\newcommand*{\Dmasked}   {\D_{\,\mathrm{Masked}}}
\newcommand*{\Dpos}      {\D_{\,\mathrm{POS}}}
\newcommand*{\Dbitmask}  {\D_{\,\mathrm{Bitmask}}}
\newcommand*{\Dtokens}   {\D_{\,\mathrm{Tok}}}
\newcommand*{\Dtoklocs}  {\D_{\,\mathrm{Tokloc}}}
\newcommand*{\lWords}    {\ell_{\,\mathrm{Words}}}

\newcommand*{\originalCorpus} {\textsf{Original}\mbox{}\xspace} 
\newcommand*{\posNoise}       {\textsf{POSNoise}\mbox{}\xspace}  
\newcommand*{\textDistortion} {\textsf{TextDistortion}\mbox{}\xspace}

\makeatletter
\newcommand*{\rom}[1]{\expandafter\@slowromancap\romannumeral #1@}
\makeatother

\renewcommand  {\quote}[1]{{``#1''}}
\newcommand    {\quotetxt}[1]{{\texttt{"#1"}}}

\newcommand*{\topicMasking}  {topic masking\mbox{}\xspace}

\newcommand*{\eg}             {e.\,g.,\mbox{}\xspace}
\newcommand*{\ie}             {i.\,e.,\mbox{}\xspace}
 
\newcommand{\e}[1]{\emph{#1}} 
\makeatletter
\newcommand\footnoteref[1]{\protected@xdef\@thefnmark{\ref{#1}}\@footnotemark}
\makeatother

\begin{document}
\title{\textbf{POSNoise: An Effective Countermeasure Against Topic Biases in Authorship Analysis}}

\author{\textbf{Oren Halvani\footnote{Corresponding author.} and Lukas Graner} \\ 
	Fraunhofer Institute for Secure Information Technology SIT,\\ Rheinstr. 75, 64295 Darmstadt, Germany\\
	\{\texttt{FirstName.LastName\}@SIT.Fraunhofer.de}}
\date{\vspace{-2ex}}
\maketitle

\begin{abstract}
Authorship verification (AV) is a fundamental research task in digital text forensics, which addresses the problem of whether two texts were written by the same person. In recent years, a variety of AV methods have been proposed that focus on this problem and can be divided into two categories: The first category refers to such methods that are based on explicitly defined features, where one has full control over which features are considered and what they actually represent. The second category, on the other hand, relates to such AV methods that are based on implicitly defined features, where no control mechanism is involved, so that any character sequence in a text can serve as a potential feature. However, AV methods belonging to the second category bear the risk that the topic of the texts may bias their classification predictions, which in turn may lead to misleading conclusions regarding their results. To tackle this problem, we propose a preprocessing technique called POSNoise, which effectively masks topic-related content in a given text. In this way, AV methods are forced to focus on such text units that are more related to the writing style. Our empirical evaluation based on six AV methods (falling into the second category) and seven corpora shows that POSNoise leads to better results compared to a well-known topic masking approach in 34 out of 42 cases, with an increase in accuracy of up to 10\%. 

\end{abstract}
\section{Introduction} \label{Introduction} 
Texts are written for a variety of purposes and appear in numerous digital and non-digital forms, including emails, websites, chat logs, office documents, magazines and books. They can be categorized according to various aspects including language, genre, topic, sentiment, readability or writing style. The latter is particularly relevant when the question regarding the authorship of a certain document (\eg ghostwritten paper, blackmail letter, suicide note, letter of confession or testament) arises. 
Stylometry is the quantitative study of writing style, especially with regard to questions of authorship, and can be dated back to the 19th century \cite{HolmesEvolutionStylometryHumanities:1998}. Stylometry uses statistical methods to analyze style on the basis of measurable features and has historical, literary and forensic applications. The underlying assumption in stylometry is that authors tend to write in a recognizable and unique way \cite{EderStyloSystemMultilevelTextAnalysis:2017,BrennanAdversarialStylometry:2012}. 
Over the years, a number of authorship analysis disciplines have been established, of which \aa (\AA) is the most widely researched. 
The task \AA is concerned with is to assign an anonymous text to the most likely author based on a set of sample documents from candidate authors. 
\AA relies on the so-called \quote{closed-set assumption} \cite{HitschlerAAWithCNNs:2017,SavoyBookAAandAP:2020}, which states that the true author of the anonymous text is indeed in this candidate set. However, if for some reason this assumption cannot be met, then an \AA method will necessarily fail to select the true author of the anonymous text.  
\\
\\
A closely related discipline to \AA is \av (\AV), which deals with the \textbf{fundamental problem} of whether two given documents $\D_1$ and $\D_2$ were written by the same person \cite{KoppelFundamentalProblemAA:2012}. If this problem can be solved, almost any conceivable \AA problem can be solved \cite{KoppelFundamentalProblemAA:2012}, which is the reason why \AV is particularly attractive for practical use cases. Based on the fact that any \AA problem can be broken down into a series of \AV problems \cite{KoppelWinter2DocsBy1:2014}, we have decided to focus in this paper on the \AV problem.
From a machine learning point of view, \AV represents a similarity detection problem, where the focus lies on the \textbf{writing style} rather than the \textbf{topic} of the documents. 
In spite of this, it can be observed in the literature that a large number of AV methods, including  \cite{AgbeyangiAVYorubaBlogPostsCharNgrams:2020,BrocardoStylometryAV:2013,BrocardoDeepBeliefAV:2017,CastroAVAverageSimilarity:2015,KoppelSeidmanEMNLP:2013,KoppelWinter2DocsBy1:2014,LitvakAVwithCNNs:2019,NealAVviaIsolationForests:2018,StamatatosPothaImprovedIM:2017,PothaStamatatosExtrinsicAV:2019}, are based on implicit\footnote{A feature category is \textbf{implicit}, if it is not clear, which type of features will be indeed captured. Whenever a sliding window is moved over a stream of text units such as characters or tokens, the captured features are necessarily implicit, as it cannot be determined beforehand what exactly they represent (in contrast to explicit features).} feature categories such as character/word or token $n$-grams. 
However, often it remains unclear which specific \quote{linguistic patterns} they cover in contrast to explicit\footnote{A feature category is \textbf{explicit}, if the specification of the underlying features is known beforehand. For example, function words are explicit, as we not only know how the extracted features will look like (\eg \quotetxt{while}, \quotetxt{for} or \quotetxt{thus}) but also what they represent (words that express grammatical relationships regarding other words).} feature categories such as punctuation marks, function words or part-of-speech (POS) tags, which can be interpreted directly. 
Since in general one has no control over implicitly defined features, it is important to ensure (for example, through a post-hoc analysis) what they actually capture. Otherwise, predictions made with \AV methods based on such features may be influenced by the topic rather than the writing style of the documents. This, in turn, can prevent \AV methods from achieving their intended goal.   
\\
\\
To counter this problem, we propose a simple but effective technique that deprives \AV methods of the ability to consider topic-related features with respect to their predictions. 
The basic idea is to retain stylistically relevant words and phrases using a predefined list, while replacing topic-related text units with their corresponding \posTags. The latter represent word classes such as nouns, verbs or adjectives and thus provide grammatical information which refer to the content of the corresponding words. \posTags have been widely used in \AA, \AV and many other disciplines related to authorship analysis. They have been confirmed to be effective stylistic features, not only for documents written in English \cite{BrocardoDeepBeliefAV:2017,HitschlerAAWithCNNs:2017,PatchalaAAConsensusAmongFeatures:2018} but also in other languages such as Russian \cite{LitvinovaPOSNgramsAP:2015}, Estonian \cite{PetmansonAVEstonian:2014} and German \cite{DiederichAAwithSVMs:2003}. 
While many \AV and \AA approaches consider simple \posTags \cite{PetmansonAVEstonian:2014}, other variants are also common in the literature including \posTag $n$-grams \cite{HirstAVUnmaskingAlzheimer:2012}, \posTag one-hot encodings \cite{HitschlerAAWithCNNs:2017}, \posTags combined with function words \cite{DiederichAAwithSVMs:2003} and probabilistic \posTag structures \cite{DingAuthorshipAnalysisRepresentations:2019}. 
\\
\\
The remainder of the paper is organized as follows: Section~\ref{ExistingApproaches} discusses previous work that served as an motivation for our approach, which is proposed in Section~\ref{ProposedApproach}. Section~\ref{Evaluation} describes our experiments and Section~\ref{Conclusions} concludes the work and provides suggestions for future work. 

\section{Previous Work} \label{ExistingApproaches}
A fundamental requirement of any \AV method is the choice of a suitable data representation, which aims to model the writing style of the investigated documents. 
The two most common representations that can be used for this purpose are (1) \textbf{vector space models} and (2) \textbf{language models}. 
A large part of existing \AV approaches fall into category (1). Kocher and Savoy \cite{KocherSavoySpatiumL1:2017}, for example, as well as 
Koppel and Schler \cite{KoppelAVOneClassClassification:2004,KoppelUnmasking:2007}, proposed \AV methods that consider the most frequent words occurring in the documents. 
Other approaches, that also make use of vector space models, are those of Potha and Stamatatos \cite{StamatatosPothaImprovedIM:2017}, 
Koppel and Winter \cite{KoppelWinter2DocsBy1:2014}, Hürlimann et al. \cite{GLAD:2015}, Barbon et al. \cite{BarbonAV4CompromisedAccountsSocialNetworks:2017} 
and Neal et al. \cite{NealAVviaIsolationForests:2018} which, among others, consider the most frequent character $n$-grams. 
On the other hand, \AV methods based on neural networks such as the approaches of Hosseinia and Mukherjee \cite{HosseiniaNeuralAV:2018}, 
Boenninghoff et al. \cite{BoenninghoffSocialMediaAV:2019}, Bagnall \cite{BagnallRNN:2015} and Jasper et al. \cite{JasperStylometricEmbeddingsAV:2018} fall into category (2). 
These approaches employ continuous space\footnote{Continuous-space language models are not limited to a fixed-size context. In contrast, count-based $n$-gram models, which represent the core of compression-based \AV methods (\eg \cite{HalvaniOCCAV:2018,HalvaniARES:2017,VeenmanPAN13:2013}) have restricted contexts, where $n$ is the limit.} language models that operate on the word and character level of the documents. 
In contrast to these, \AV approaches as those proposed by Veenman and Li \cite{VeenmanPAN13:2013} or Halvani et al. \cite{HalvaniOCCAV:2018,HalvaniARES:2017} are based on compression-based language models, where internally a probability distribution for a given document is estimated based on all characters and their preceding contexts. Regarding the latter, Bevendorff et al. \cite{BevendorffUnmasking:2019} have shown that this type of \AV methods are effective compared to the current \stateOfTheArt in \AV. 
Regardless of their strengths and effectiveness, all the above-mentioned \AV approaches suffer from the same problem. They lack a control mechanism which ensures that their decision with respect to the questioned authorship, is not inadvertently distorted by the topic of the documents. In the absence of such a control mechanism, \AV methods can (in the worst case) degenerate from style to simple topic classifiers. 
\\
\\
To address this problem, Stamatatos \cite{StamatatosTextDistortion:2017} proposed a technique that we refer to in this paper as \textDistortion\footnote{An implementation of \textDistortion is available under \newline \url{https://paperswithcode.com/paper/authorship-attribution-using-text-distortion}.}. The method aims to mask topic-specific information in documents, before passing them further to \AA or \AV methods. The topic-specific information is not related to the author's personal writing style, which is why masking helps to maintain the correct objective 
(classifying documents by their \textbf{writing style} rather than by their \textbf{content}). 
To achieve this, occurrences of infrequent words are substituted entirely by uniform symbols. In addition, numbers are masked such that their structure is retained while hiding their specific value. Given these transformations, most of the syntactical structure of the text is retained (including capitalization and punctuation marks) which is more likely to be associated with the authors writing style \cite{StamatatosTextDistortion:2017}. 
Stamatatos introduced the following two variants of \textDistortion, which require as a prerequisite a word list $W_{k}$ containing the $k$ most frequent words\footnote{\textDistortion uses the \e{British National Corpus} (BNC)  word list available under \newline \url{https://www.kilgarriff.co.uk/bnc-readme.html}.} in the English language: 
\begin{itemize}
	\item \textbf{Distorted View - Single Asterisk (\textbf{DV-SA}):} Every word $w \notin W_{k}$ in a given document $\D$ is masked by replacing each word occurrence with a single asterisk \texttt{*}. Every sequence of digits in $\D$ is replaced by a single hashtag \texttt{\#}. 
		
	\item \textbf{Distorted View - Multiple Asterisks (\textbf{DV-MA}):} Every word $w \notin W_{k}$ in $\D$ is masked by replacing each of its characters with \texttt{*}. Every digit in $\D$ is replaced by \texttt{\#}.
\end{itemize} 
Both variants can be applied to any given document $\D$ written in English, without the need for specific NLP tools or linguistic resources (besides the word list $W_{k}$). 
However, in order to apply \textDistortion to $\D$, the hyperparameter $k$, which regulates how much content is going to remain in $\D$, must be carefully specified beforehand. 
Moreover, one must take into account that the replacement of each potentially topic-related word $w$ in $\D$ is performed uniformly without any further distinction as to what $w$ represents. 
Consequently, the masking procedure may necessarily miss relevant information associated with $w$ that could serve as a useful stylistic feature. 
\section{Proposed Approach} \label{ProposedApproach} 
Inspired by the approach of Stamatatos \cite{StamatatosTextDistortion:2017}, we propose an alternative \topicMasking technique called $\bm{\mathsf{POSNoise}}$ (\quote{\underline{POS}-Tag-based \underline{Noise} smoothing}), which addresses the two issues of \text distortion mentioned in Section~\ref{ExistingApproaches}. 
The core idea of our approach is to keep stylistically relevant words and phrases in a given document $\D$ using a predefined list $\listStylePatterns$, while replacing topic-related words with their corresponding \posTags represented by a set $\mathcal{S}$. Text units not covered by $\listStylePatterns$ and $\mathcal{S}$, such as punctuation marks and idiosyncratic words, are further retained. 
In what follows, we first describe the requirements of \posNoise and explain how it differs from the existing \textDistortion approach. 
Afterwards, we present the respective steps of our \topicMasking algorithm, which is listed in Algorithm~\ref{POSNoiseAlgo} as Python-like pseudocode. 
\\
\\
Similarly to \textDistortion, our approach also relies on a predefined list\footnote{The list is available under \url{http://bit.ly/ARES-2021}.} $\listStylePatterns$ of specific words that should not be masked. 
However, unlike \textDistortion, which uses a list of words ordered by frequency of occurrence in the \e{British National Corpus} (BNC), our list $\listStylePatterns$ is structured by grammatical factors. 
More precisely, $\listStylePatterns$ comprises different categories of function words, phrases, contractions, generic adverbs and empty verbs. 
Regarding the function word categories, we consider conjunctions, determiners, prepositions, pronouns and quantifiers, which are widely known in the literature (\eg \cite{PavelecAAConjunctionsAndAdverbs:2008,StolermanPhD:2015}) to be content and topic independent. 
With respect to the phrases, we use different categories of transitional phrases including \e{causation}, \e{contrast}, \e{similarity}, \e{clarification}, \e{conclusion}, \e{purpose} and \e{summary}. 
As generic adverbs, we consider \e{conjunctive}, \e{focusing}, \e{grading} and \e{pronominal} adverbs, while as empty verbs, we take \e{auxiliary} and \e{delexicalised} verbs into account, as these have no meaning on their own. The tenses\footnote{For this purpose, we used \e{pattern} available under \url{https://github.com/clips/pattern}.} of verbs are additionally considered so that AV methods operating at the character level of documents can benefit from morphological features occurring in the inflected form of such words. 
Table~\ref{table:POSNoiseFeatures} lists all the categories of words and phrases considered by \posNoise, together with a number of examples. 
Note that the comparison of the text units with $\listStylePatterns$ is case-insensitive (analogous to \textDistortion), \ie the original case of the text units is retained. We also wish to emphasize that in our list $\listStylePatterns$ (in contrast to the word list $W_{k}$ used by \textDistortion) \textbf{topic-related} nouns and pronouns, verbs, adverbs and adjectives are not present. According to Sundararajan and Woodard \cite{SundararajanWhatIsStyle:2018}, especially the former two are strongly influenced by the content of the documents. 
\begin{table} 
	\centering\footnotesize  
		\begin{tabular}{ll} 
			\toprule 
			\textbf{Category} & \textbf{Examples} \\ \midrule
			Contractions         & $\{\texttt{i'm, i'd, i'll, i've, it's, we're, how's,}\,\e{...}\,\}$  \\
			Auxiliary verbs      & $\{\texttt{can, could, might, must, ought, shall, will,}\,\e{...}\,\}$  \\
			Delexicalised verbs  & $\{\texttt{get, go, take, make, do, have, give, set,}\,\e{...}\,\}$  \\
			Conjunctions         & $\{\texttt{and, as, because, but, either, for, however,}\,\e{...}\,\}$  \\
			Determiners          & $\{\texttt{a, an, both, either, every, no, other, some,}\,\e{...}\,\}$  \\
			Prepositions         & $\{\texttt{above, below, beside, between, beyond, inside,}\,\e{...}\,\}$  \\
			Pronouns             & $\{\texttt{all, another, anyone, anything, everything,}\,\e{...}\,\}$  \\
			Quantifiers          & $\{\texttt{any, certain, each, either, lots, neither,}\,\e{...}\,\}$  \\
			Generic adverbs      & $\{\texttt{only, almost, just, again, yet, therefore,}\,\e{...}\,\}$  \\
			Transitional phrases & $\{\texttt{of course, because of, in contrast,}\,\e{...}\,\}$  \\
			\bottomrule
		\end{tabular}
	\caption{All categories of function words and phrases contained in our list $\listStylePatterns$. \label{table:POSNoiseFeatures}}
\end{table} 
\\
\\
Besides our statically defined list $\listStylePatterns$, we also make use of certain dynamically generated \posTags to retain stylistic features. 
For this, we apply a POS tagger\footnote{For this, we used \e{spaCy} (model \quotetxt{en\_core\_web\_lg}) available under \url{https://spacy.io}.} to $\D$ so that a sequence of pairs $\langle t_i, p_i \rangle$ is created, where $t_i$ denotes a token and $p_i$ its corresponding \posTag. 
Here, we decided to restrict ourselves to the \e{Universal POS Tagset}\footnote{The list of all \posTags is available under \url{https://universaldependencies.org/u/pos}.} 
so that each $p_i$ falls into a coarse-grained POS category (cf. Table~\ref{table:SubstitutionPOSTags}) of the token $t_i$. 
There are two reasons why we decided to use this tagset. First, universal \posTags allow a better adaptation of \posNoise to other languages, 
as it can be observed that the cardinality of fine-grained \posTags differ from language to language \cite{PetrovUniversalPOSTagsa:2012}. 
Second, the POS tagger might lead to more misclassified \posTags if the fine-grained tagset is used instead. 
Note that we do not use the original form of the tags as they appear in the tagset such as \texttt{PROPN} (proper noun) or \texttt{ADJ} (adjective). 
Instead, we use individual symbols as \textbf{representatives} that are more appropriate with respect to \AV methods that operate at the character level of documents. However, for readability reasons, we refer to these symbols as \quote{tags}. 
Once $\D$ has been tagged, \posNoise substitutes all adjacent pairs $\langle t_i, p_i \rangle,  \langle t_{i + 1}, p_{i + 1} \rangle, \ldots, \langle t_{i + n}, p_{i + n} \rangle$, whose tokens $t_i, \ldots, t_n$ form an element in $\listStylePatterns$, with their corresponding pos tags $p_i, \ldots, p_n$, respectively. However, the replacement is only performed if $p_i \in \mathcal{S} = \{$\textsf{\#}, \textsf{§}, \textsf{Ø}, \textsf{@}, \textsf{©}, \textsf{$\mu$}, \textsf{\$}, \textsf{\yen}$\}$ holds (cf. Table~\ref{table:SubstitutionPOSTags}). 
More precisely, every token $t_i$ for which $p_i \notin \mathcal{S}$ applies is retained, in addition to all words and phrases in the document $\D$ that occur in $\listStylePatterns$. The retained tokens are, among others, \textbf{punctuation marks} and \textbf{interjections} (\eg \quotetxt{yes}, \quotetxt{no}, \quotetxt{okay}, \quotetxt{hmm}, \quotetxt{hey}, etc.), where the latter represent highly personal and thus idiosyncratic stylistic features according to Silva et al. \cite{SilvaMicroBlogAA:2011}. 
\\
\\
Regarding numerals, we keep written numbers unmasked as such words and their variations may reflect stylistic habits of certain authors (\eg \quotetxt{one hundred} / \quotetxt{one-hundred}). Digits, numbers and roman numerals, on the other hand, are masked by the \posTag \textsf{$\mu$}. 
\begin{table} 
	\centering\small
	\begin{tabular}{lcl} 
		\toprule 
		\textbf{Category} & \textbf{Tag} & \textbf{Examples} \\\midrule   
		Noun        & \textsf{\#} &  $\{\texttt{house, music, bird, tree, air, }\,\e{...}\,\}$  \\
		Proper noun & \textsf{§}  &  $\{\texttt{David, Vivien, London, USA, COVID-19,}\,\e{...}\,\}$  \\
		Verb        & \textsf{Ø}  &  $\{\texttt{eat, laugh, dance, travel, hiking,}\,\e{...}\,\}$  \\
		Adjective   & \textsf{@}  &  $\{\texttt{red, shiny, fascinating, phenomenal,}\,\e{...}\,\}$  \\
		Adverb      & \textsf{©}  &  $\{\texttt{financially, foolishly, angrily,}\,\e{...}\,\}$  \\
		Numeral     & \textsf{$\mu$}  &  $\{\texttt{0, 5, 2013, 3.14159, III, IV, MMXIV,}\,\e{...}\,\}$  \\
	    Symbol      & \textsf{\$} &  $\{\texttt{\pounds, \copyright, §, \%, \#,}\,\e{...}\,\}$  \\  
		Other       & \textsf{\yen}   &  $\{\texttt{xfgh, pdl, jklw, }\,\e{...}\,\}$  \\  
		\bottomrule		
	\end{tabular}
	\caption{\posTags considered by \posNoise that aim to replace topic-related words, after all words and phrases listed in Table~\ref{table:POSNoiseFeatures} have been retained. \label{table:SubstitutionPOSTags}}
\end{table}
In a subsequent step, we adjust punctuation marks that were separated from their adjacent words (\eg \quotetxt{however ,} $\rightsquigarrow$ \quotetxt{however,}) as a result of the tokenization process of the POS tagger. Our intention is to retain the positional information of the punctuation marks, as certain \AV methods might use standard tokenizers that split by white-spaces so that \quotetxt{however ,} would result in two different tokens. As a final step, all tokens are concatenated into the topic-masked representation $\Dmasked$. 
In Section~\ref{Evaluation}, we compare the resulting representations of \posNoise and \textDistortion. For the latter, we also explain how a suitable setting of the hyperparameter $k$ (\ie one that suppresses topic-related words but retains style-related words as best as possible) was determined. 
\begin{algorithm} [h!]
	\footnotesize
	\SetKwInput{KwData}{Input}
	\SetKwInput{KwResult}{Output}
	\SetKw{KwBreak}{break}
	\SetKw{KwContinue}{continue}
	\SetKw{KwThrow}{throw}
	\SetKwIF{If}{ElseIf}{Else}{if}{}{else if}{else}{end if}%
	\SetKwFor{For}{for}{}{end for}%
	\SetKwFor{ForEach}{foreach}{}{end foreach}%
	\SetKwFor{While}{while}{}{end while}%
	\SetNoFillComment
	\DontPrintSemicolon
	\KwData{Document $\D$, pattern list $\listStylePatterns$}
	\KwResult{Topic-masked representation $\Dmasked$}
	\BlankLine
	
	$\Dtokens \leftarrow$ tokenize($\D$) \tcc{Segment $\D$ into a list of tokens $\Dtokens$.}    
	$\Dtoklocs \leftarrow$ token\_locations($\D$) \tcc{Store for all tokens their respective start locations in $\D$.}  
	$\Dpos \leftarrow$ pos\_tag($\Dtokens$) \tcc{Classify for each $t_i$ in $\Dtokens$ its corresponding \posTag.}
	$n \leftarrow$ length($\Dtokens$) \\	
	\BlankLine
	
	\tcc{Initialize a bitmask array for the $n$ tokens in $\D$. Activated bits ('1') reflect tokens that should be retained.}
	$\Dbitmask \leftarrow [0, 0, 0, \ldots, 0]$  	
	\BlankLine
	\ForEach{$\ell \in \listStylePatterns$}
	{
		$\lWords \leftarrow$ tokenize($\ell$) \tcc{Note that $\ell$ might be a phrase such as \texttt{"apart from this"}.} 
		$m \leftarrow \mathrm{length}(\lWords)$ \\
		$x \leftarrow$ 0 \\
		$i \leftarrow$ 0
		
		\While{$i < n$}
		{
			\If{$\mathrm{lowercase(}\Dtokens[i]\mathrm{) == lowercase(}\lWords[x]\mathrm{)}$}
			{
				$x \leftarrow x + 1$ \\				
				
				\If{$x == m$}
				{
					\For{$j \leftarrow i - m + 1; j < i + 1; j \leftarrow j + 1$}
					{
						$\Dbitmask[j] \leftarrow 1$
					}
					$x \leftarrow 0$
				}
			}
			\Else
			{
				$i \leftarrow i - x$ \\
				$x \leftarrow 0$ 
			}
			$i \leftarrow i + 1$			
		}
	}

	\BlankLine
	\tcc{Define POS tags that aim to replace topic-related words.} 		
	$ \mathcal{S} = \{$ \texttt{"\#"}, \texttt{"§"}, \texttt{"Ø"}, \texttt{"@"}, \texttt{"©"}, \texttt{"$\mu$"}, \texttt{"\$"}, \texttt{"\yen"} $\}$
	
	\BlankLine

	$\Dmasked \leftarrow \D$ \\	
	\For{$i \leftarrow n - 1; i \geq 0; i \leftarrow i - 1$}
	{
		\tcc{Retain truncated contraction tokens.}
		\If{$\Dtokens[i] \in \{$  \e{\texttt{"'m"}}, \e{\texttt{"'d"}}, \e{\texttt{"'s"}}, \e{\texttt{"'t"}}, \e{\texttt{"'ve"}}, \e{\texttt{"'ll"}}, \e{\texttt{"'re"}}, \e{\texttt{"'ts"}} $\}$ }
		{
			$\Dbitmask[i] \leftarrow 1$
		}	
		
		\If{$\Dtokens[i]$ \e{represents a written-out number (\eg \texttt{"four"} or \texttt{"twelve"})}}
		 {
		 	$\Dbitmask[i] \leftarrow 1$
		 }
	 
		\If{ $\Dbitmask[i] == 0$}
		{
			sub $ \leftarrow $ \texttt{""} \\
			\lIf{$\Dpos[i] \in \mathcal{S} $}
			{
				sub $ \leftarrow \Dpos[i]$ 
			}
			\lElse
			{
				sub $ \leftarrow \Dtokens[i]$ 
			}		 	
		 	$\Dmasked \leftarrow \Dmasked[:\Dtoklocs[i]] + \mathrm{sub} + \Dmasked[(\Dtoklocs[i] + \mathrm{length(}\Dtokens[i]\mathrm{)}):]$
		}
	}	
	\KwRet{$\Dmasked$}\;
	\caption{POSNoise}
	\label{POSNoiseAlgo}
\end{algorithm}
\section{Experimental Evaluation} \label{Evaluation} 
In the following, we present our experimental evaluation. 
First, we present our seven compiled corpora and explain for them where the documents were obtained and how they were preprocessed, followed by a summary of their main statistics. 
Next, we mention which existing \AV methods were chosen for the evaluation as well as how they were trained and optimized. 
Afterwards, we explain how an appropriate setting was chosen with regard to the topic-regularization hyperparameter of \textDistortion to allow a fair comparison between it and our \posNoise approach. 
Finally, we present the results and describe our analytical findings.  

\subsection{Corpora} 
To compare our approach against \textDistortion, we compiled seven English corpora covering a variety of challenges, 
such as texts (of the same authors) written at different periods of time, texts with an excessive use of slang and texts with no cohesion. 
In total, the corpora comprise 4.742 verification cases, were each corpus $\Corpus = \{ \Problem_1, \Problem_2, \ldots \}$ is split into author-disjunct training and test partitions based on a 40/60 ratio. 
Each $\Problem \in \Corpus$ denotes a verification case $(\Dunk, \Arefset)$, where $\Dunk$ represents an unknown document and $\Arefset$ a set of sample documents of the known author $\A$. 
To counteract \e{population homogeneity}, a form of an \AV bias described by Bevendorff \cite{BevendorffBiasAV:2019}, 
we ensured that for each $\A$ there is one \textbf{same-authorship} (\classY) and one \textbf{different-authorship} (\classN) verification case. 
Furthermore, we constructed all corpora so that the number of \classY and \classN cases is equal (in other words, all training and test corpora are \textbf{balanced}). 
In the Sections~\ref{Corpora_Gutenberg}$\,$-$\,$\ref{Corpora_Reddit}, we introduce the corpora and summarize their key statistics in Table~\ref{tab:CorpusStatistics}. 
\newcolumntype{R}{>{\raggedleft\arraybackslash}X}
\begin{table} 
	\centering \footnotesize 
	\begin{tabularx}{11cm}{lXllrrRR} \toprule
		\bfseries\boldmath Corpus           & \textbf{Genre} & \textbf{Topic} & \textbf{Partition} & \boldmath$|\Corpus|$ & \boldmath$|\Arefset|$ & \bfseries\boldmath avg$|\DA|$ & \bfseries\boldmath avg$|\Dunk|$  \\\midrule
        \multirow{2}{*}{$\CorpusGutenberg$} & Fiction    & Mixed & $\CorpusTrain$ & 104  & 1 & 21660 & 21787 \\
                                            & books      & topics  & $\CorpusTest$  & 156  & 1 &	21879 &	21913 \\\midrule
        \multirow{2}{*}{$\CorpusWikiSockpuppets$} & Wikipedia  & Related & $\CorpusTrain$ & 150 & 3 & 2280 & 2983 \\
        										  & talk pages & topics  & $\CorpusTest$  & 226 & 3 & 2363 & 3042 \\\midrule
		\multirow{2}{*}{$\CorpusACL$}       & Scientific & Related & $\CorpusTrain$ & 186 & 1 &	9497 &	14196 \\
		                                    & papers     & topics  & $\CorpusTest$  & 280 & 1 &	8250 &	13913 \\\midrule
		\multirow{2}{*}{$\CorpusPeeJ$}      & Chat       & Related & $\CorpusTrain$ & 208 &  $\approx\,$2 &	2329 & 2946 \\
		                                    & logs       & topics  & $\CorpusTest$  & 312 &  $\approx\,$2 &	2324 & 2967 \\ \midrule
		\multirow{2}{*}{$\CorpusTelegraph$} & News       & Related & $\CorpusTrain$ & 220 & 2 & 3329 &	4946 \\
		                                    & articles   & topics  & $\CorpusTest$  & 332 & 2 & 3068 &	4702 \\\midrule
		\multirow{2}{*}{$\CorpusApricity$}  & Forum      & Related & $\CorpusTrain$ & 228 & $\approx\,$4 & 3900 & 4023 \\
		                                    & posts      & topics  & $\CorpusTest$  & 340 & $\approx\,$4 & 3921 & 4020 \\\midrule
		\multirow{2}{*}{$\CorpusReddit$}    & Social     & Mixed   & $\CorpusTrain$ &  800 &  3 & 5735 & 6785 \\
		                                    & news       & topics  & $\CorpusTest$  & 1200 &  3 & 5854 & 6794 \\\bottomrule
	\end{tabularx}	
	\caption{Overview of our corpora and their key statistics. Notation: $\bm{|\Corpus|}$ denotes the \textbf{number of verification cases} in each corpus $\bm{\Corpus}$, while $\bm{|\Arefset|}$ denotes the \textbf{number of the known documents}. The \textbf{average character length} of $\bm{\Dunk}$ and $\bm{\DA}$ is denoted by avg$\bm{|\Dunk|}$ and avg$\bm{|\DA|}$, respectively. Note that in this context $\bm{\DA}$ represents the \textbf{concatenation} of all documents in $\bm{\Arefset}$. \label{tab:CorpusStatistics}}
\end{table}

\subsubsection{Gutenberg Corpus ($\CorpusGutenberg$)} \label{Corpora_Gutenberg} 
This corpus comprises 260 texts taken from the \e{Webis Authorship Verification Corpus 2019} dataset released by Bevendorff et al. \cite{BevendorffBiasAV:2019}. 
The texts in $\CorpusGutenberg$ represent fragments extracted from fiction books that stem from the \e{Project Gutenberg} portal. 
$\CorpusGutenberg$ is the only corpus where the contained documents have not been preprocessed by us, as they have already gone through a clean and well-designed preprocessing routine, which is described in detail by Bevendorff et al.  \cite{BevendorffBiasAV:2019}. The only action we took was to re-arrange the training and test partitions and to balance the \classYdash{} and \classNdash{cases}, since the original partitions were both imbalanced. 

\subsubsection{Wikipedia Sockpuppets Corpus ($\CorpusWikiSockpuppets$)} \label{Corpora_WikipediaSockpuppets} 
This corpus comprises 752 excerpts of 288 Wikipedia talk page editors taken from the \e{Wikipedia Sockpuppets} dataset released by Solorio et al. \cite{SolorioSockpuppet:2014}. 
The original dataset contains two partitions comprising sockpuppets and non-sockpuppets cases, where for $\CorpusWikiSockpuppets$ we considered only the latter subset. 
In addition, we did not make use of the full range of authors within the non-sockpuppets cases, as appropriate texts of sufficient length were not available for each of the authors.     
From the considered texts, we removed Wiki markup, timestamps, URLs and other types of noise. 
Besides, we discarded sentences with many digits, proper nouns and near-duplicate string fragments as well as truncated sentences. 

\subsubsection{ACL Anthology Corpus ($\CorpusACL$)} \label{Corpora_AclAnthology} 
This corpus comprises 466 paper excerpts from 233 researchers, which were collected from the computational linguistics archive \e{ACL Anthology}\footnote{\url{https://www.aclweb.org/anthology}}. The corpus was constructed in such a way that for each author there are exactly two papers\footnote{Note that we ensured that each paper was single-authored.}, stemming from different periods of time. From the original papers we tried, as far as possible, to restrict the content of each text to only such sections that mostly comprise natural-language text (\eg \e{abstract}, \e{introduction}, \e{discussion}, \e{conclusion} or \e{future work}). 
To ensure that the extracted fragments met the important \AV bias cues of Bevendorff et al. \cite{BevendorffBiasAV:2019}, we preprocessed each paper extract in $\CorpusACL$ manually. Among others, we removed tables, formulas, citations, quotes, references and sentences that include non-language content such as mathematical constructs or specific names of researchers, systems or algorithms. The average time span between both documents of each author is $\approx$12 years, whereas the minimum and maximum time span are 8 and 31 years, respectively. 
Besides the temporal aspect of $\CorpusACL$, another characteristic of this corpus is the formal (scientific) language, where the usage of stylistic devices (\eg repetitions, metaphors or rhetorical questions) is more restricted, in contrast to other genres such as chat logs. 

\subsubsection{Perverted Justice Corpus ($\CorpusPeeJ$)} \label{Corpora_PeeJ}
This corpus comprises 738 chat logs of 260 sex offenders, which have been crawled from the \e{Perverted-Justice} portal\footnote{\url{http://www.perverted-justice.com}}. 
The chat logs stem from older instant messaging clients (\eg \e{MSN}, \e{AOL} or \e{Yahoo}), where we ensured for each conversation that only chat lines from the offender were extracted. 
To obtain as much language variability as possible regarding the content of the conversations, we selected chat lines from different messaging clients and different time spans (where possible) and considered such lines that differed mostly from each other using a similarity measure\footnote{For this purpose, we used the \e{FuzzyWuzzy} library available under \url{https://github.com/seatgeek/fuzzywuzzy} and selected the \e{Levenshtein} distance as a similarity measure.}. 
One characteristic of the chats in $\CorpusPeeJ$ is an excessive use of slang, a variety of specific abbreviations and other forms of noise. 
As further preprocessing steps, we discarded chat lines with less than 5 tokens (or 20 characters) and such lines containing usernames, timestamps, URLs as well as words with repetitive characters (\eg \quotetxt{yoooo} or \quotetxt{jeeeez}). 

\subsubsection{The Telegraph Corpus ($\CorpusTelegraph$)} 
This corpus consists of 828 excerpts of news articles from 276 journalists, crawled from \e{The Telegraph} website. 
Due to their nature, the original articles contain many verbatim quotes, which can distort the writing style of the author of the article. 
To counter this problem, we sampled from each article such sentences that did not contain quotations and other types of noise including headlines and URLs. 
As a result, the underlying characters in each preprocessed article are solely restricted to (case-insensitive) letters, spaces and common punctuation marks. 
Finally, we concatenated the preprocessed sentences from each article into a single document. 
Note that due to this procedure the coherence of the resulting document is distorted.  
Consequently, \AV methods that make use of character and/or word $n$-grams may capture \quote{artificial features} that occur across sentence boundaries. 

\subsubsection{The Apricity Corpus ($\CorpusApricity$)} 
This corpus comprises 1,395 posts from 284 users that were obtained from \e{The Apricity - A European Cultural Community}\footnote{\url{https://theapricity.com}} portal. The postings are distributed across different subforums with related topics (\eg \e{anthropology}, \e{genetics}, \e{race and society} or \e{ethno cultural discussion}). 
To construct $\CorpusApricity$, we ensured that all documents within each verification case stem from different subforums. 
The crawled postings have been cleaned from markup tags, URLs, signatures, quotes, usernames and other forms of noise.  
 
\subsubsection{Reddit Corpus ($\CorpusReddit$)} \label{Corpora_Reddit}
This corpus consists of 4,000 posts from 1,000 users, which were crawled from the \e{Reddit} community network. 
Each document in $\CorpusReddit$ has been aggregated from multiple posts from the same so-called \emph{subreddit} to obtain a sufficient length. 
However, all documents within each verification case originate from different \emph{subreddits} with unrelated topics. 
Hence, in contrast to the $\CorpusApricity$ corpus, $\CorpusReddit$ represents a mixed-topic corpus. 
In total, $\CorpusReddit$ covers exactly 1,388 different topics including \e{politics}, \e{science}, \e{books}, \e{news} and \e{movies}. 

\subsection{Considered AV Methods} 
To compare the effectiveness of \posNoise and \textDistortion, we selected six well-known \AV methods that have shown their potential in various studies 
(\eg \cite{HalvaniARES:2017,BevendorffUnmasking:2019,StamatatosTextDistortion:2017}). 
All of the chosen methods are based on \textbf{implicit feature categories} and are thus susceptible to topic influences. 
Four of these (\coav \cite{HalvaniARES:2017}, \occav \cite{HalvaniOCCAV:2018}, \veenmanNNCD \cite{VeenmanPAN13:2013} and \stamatatosProf \cite{StamatatosProfileCNG:2014}) rely on \charNgrams, while the remaining two (\spatium \cite{KocherSavoySpatiumL1:2017} and \koppelUnmask \cite{KoppelAVOneClassClassification:2004}) are based on frequent words/tokens. In the following, we describe some design decisions we have made with regard to these methods and explain which and how their hyperparameters were set. 

\subsubsection{Source of Impostor Documents} 
\veenmanNNCD and \spatium represent binary-extrinsic \AV methods (cf. \cite{HalvaniAssessingAVMethods:2019}), meaning that they rely on so-called \quote{impostor documents} (external documents outside the respective verification case) for their classification of whether or not there is a matching authorship. In the original paper, Veenman and Li \cite{VeenmanPAN13:2013} did not provide an automated solution to generate the impostor documents, but collected them in a manual way using a search engine. 
However, since this manual approach is not scalable, we opted for an alternative idea in which the impostor documents were taken directly from the test corpora. 
This strategy has been also considered by Kocher and Savoy \cite{KocherSavoySpatiumL1:2017} with respect to their \spatium approach. 
Although using static corpora is not as flexible as using search engines, it has the advantage that due to the available metadata (for instance, user names of the authors) the true author of the unknown document is likely not among the impostors\footnote{However, we cannot guarantee if different user names in fact refer to different persons (multiple accounts might refer to the same person).}. Furthermore, the documents contained in the test corpora are already cleaned and therefore do not require additional preprocessing. 

\subsubsection{Uniform (Binary) Predictions} 
In their original form, \spatium and \stamatatosProf allow the three possible prediction outputs \classY (same-author), \classN (different-author) and \unanswered (unanswered), whereas for the remaining approaches only binary predictions (\classY/\classN) are considered. To enable a fair comparison, we therefore decided to unify the predictions of all involved \AV methods to the binary case. 
In this context, verification cases for which the \AV methods determined similarity values greater than 0.5 were classified as \classY, otherwise as \classN. Here, all similarity values were normalized into the range $[0, 1]$, so that 0.5 marks the decision threshold. 

\subsubsection{Settings for \coav, \occav, and \veenmanNNCD} 
All these three represent compression-based \AV methods based on the PPMd compression algorithm, as specified in the original papers \cite{HalvaniARES:2017,HalvaniOCCAV:2018,VeenmanPAN13:2013}. However, in these papers it has not been mentioned how the \e{model-order} hyperparameter of PPMd has been set. We therefore decided to set this hyperparameter to 7 for all three methods, based on our observation\footnote{Regarding the \e{model-order}  hyperparameter we experimented with values in $[1,10]$.} that this value led to the best accuracy across all training corpora. 
Moreover, we used the same dissimilarity functions that were specified in the original papers (CDM for \veenmanNNCD as well as CBC for \coav\footnote{An implementation of \coav is available under \newline \url{https://paperswithcode.com/paper/authorship-verification-based-on-compression}.} and \occav\footnote{An implementation of \occav is available under \newline \url{https://paperswithcode.com/paper/authorship-verification-in-the-absence-of}.}). Apart from these, there are no other hyperparameters for these approaches. 

\subsubsection{Counteracting Non-Deterministic Behavior} 
The two \AV methods \spatium and \koppelUnmask involve different sources of randomness (\eg impostor selection and chunk generation) and, due to this, cause non-deterministic behavior regarding their predictions. In other words, applying these methods multiple times to the same verification case can result in different predictions which, in turn, can lead to a biased evaluation (cf. \cite{HalvaniAssessingAVMethods:2019}). 
To address this issue, we performed 11 runs for each method and selected the run where the accuracy score represented the median. The reason we avoided averaging multiple runs (as was the case, for example, in \cite{StamatatosPothaImprovedIM:2017}) was to obtain more precise numbers with respect to our analysis.   

\subsubsection{Model and Hyperparameters} 
Apart from \spatium\footnote{We used the original implementation of \spatium available under \url{https://github.com/pan-webis-de}.} and \occav, the remaining four \AV methods involve model and adjustable hyperparameters. The former represent parameters that are estimated directly from the data, while in contrast hyperparameters must be set manually.  
Of the four respective \AV methods considered in our experiments, model parameters represent the weights that form the SVM-hyperplanes (used by \koppelUnmask) or the thresholds required to accept or reject the questioned authorships (used by \coav and \stamatatosProf). To obtain the model parameters of \stamatatosProf and \coav, we trained both methods on the \originalCorpus, \posNoise and \textDistortion training corpora, respectively. 
The hyperparameters involved in our selected \AV approaches represent, among others, the number of $k$ cross-validation folds (used by \koppelUnmask) or the $n$-order of the \charNgrams (used by \stamatatosProf) and have been optimized as follows. Regarding \koppelUnmask, an adjustment was needed to fit our experimental setting. In the original definition of this method, Koppel and Schler \cite{KoppelAVOneClassClassification:2004,KoppelUnmasking:2007} considered entire books to train and evaluate \koppelUnmask, which differ in lengths from the documents used in our experiments. Therefore, instead of using the original fixed hyperparameter settings (which would make \koppelUnmask inapplicable in our evaluation setting), we decided to consider individual hyperparameter ranges with values that are more appropriate for shorter documents as available in our corpora. The customized hyperparameter ranges are listed in Table~\ref{tab:UnmaskingHyperparams}. 
For \stamatatosProf, on the other hand, we employed the same hyperparameter ranges described in the original paper \cite{StamatatosProfileCNG:2014}. 
Based on the original and adjusted hyperparameter ranges of \stamatatosProf and \koppelUnmask, we optimized both methods on the training partitions of \textbf{\originalCorpus}, \textbf{\posNoise} and \textbf{\textDistortion} using grid search guided by accuracy as a performance measure. The resulting hyperparameters are listed in Table~\ref{tab:BaselinesHyperparameters}. 

\subsection{Comparison Between Representations} \label{TextDistortion_Setup}  
\begin{figure} [h!]
	\centering
	\includegraphics[width=0.45\linewidth]{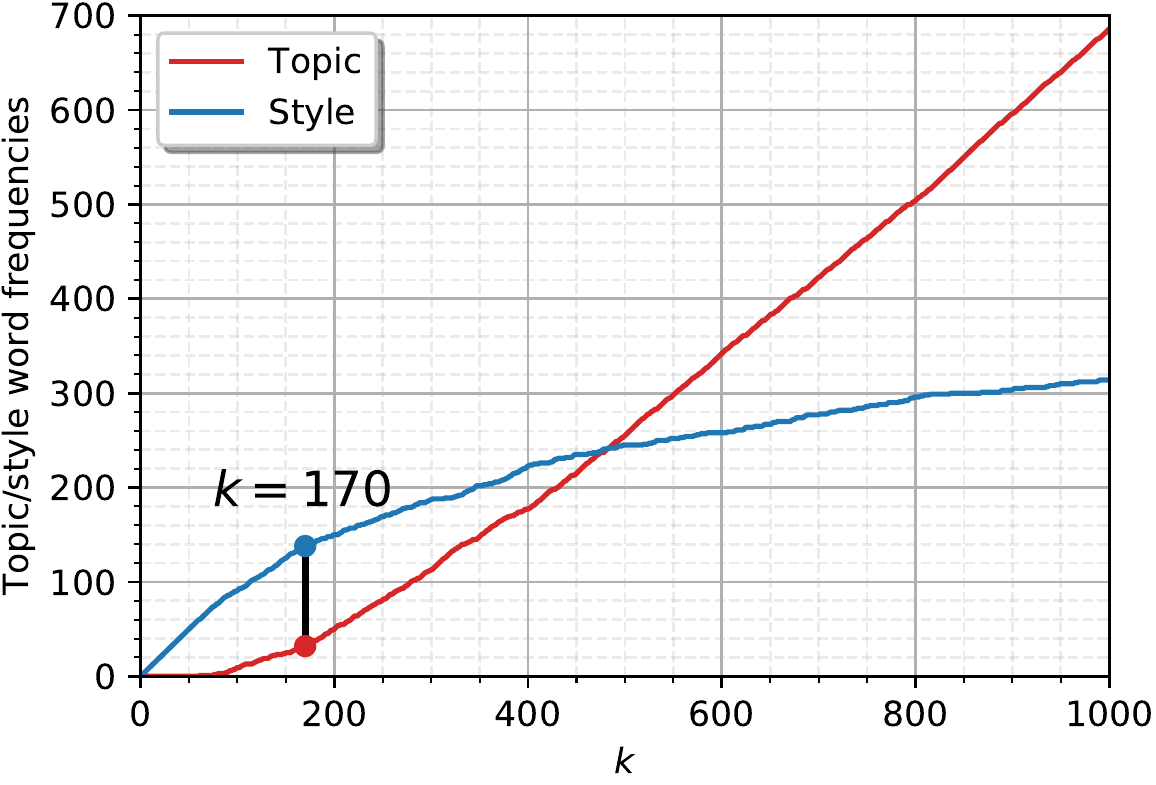}
	\caption{Choice of the topic regularization hyperparameter $\bm{k}$ of \textDistortion. \label{TextDistortion_ChoiceOfK}}
\end{figure}

To allow a reasonable comparison between \posNoise and \textDistortion, a suitable and fixed configuration for the latter is necessary. 
For this, we opted for the DV-SA variant on the basis of the following considerations: 
\begin{itemize}
	\item Stamatatos \cite{StamatatosTextDistortion:2017} conducted a number of \AA and \AV experiments with the two variants DV-SA and DV-MA, where no differences were observed in the context of \AA. However, with regard to the \AV experiments Stamatatos found that the DV-SA variant was more competitive than DV-MA.
		
	\item Both \posNoise and the \textDistortion variant DV-SA substitute topic-related text units token-wise, which allows a better comparability. 
	
	\item None of the selected \AV methods consider word length as a separate feature, so there is no advantage with regard to this possibility that is only maintained by the DV-MA variant. 	   	
\end{itemize}
Besides the choice between the two variants DV-SA and DV-MA an adjustable hyperparameter $k$ must be set for \textDistortion, which regulates which tokens (not necessarily words) are retained, while all other tokens are masked. 
However, finding a suitable value for $k$ can be seen as a problematic trade-off, since a low value suppresses too many style-related tokens, while in contrast a higher value leaves many topic-related tokens unmasked. To address this trade-off, we therefore pursued the following systematic approach.   
Given $W_k$ (the list on which \textDistortion is based) we divided all the words contained in it with respect to their index in $W_k$ into two groups, 
which include style-related tokens (\eg conjunctions, determiners, prepositions, pronouns, etc.) and topic-related tokens (\eg nouns, verbs and adjectives), respectively. 
Then, we visualized the distributions of the tokens in both groups, where it can be seen in Figure~\ref{TextDistortion_ChoiceOfK} that the distribution 
of the style-related tokens increase at a decreasing rate, while the topic-related tokens increase linearly as the $k$ value increases. 
In other words, as $k$ increases, fewer style-related tokens occur in $W_k$, while at the same time a higher number of topic-related tokens are present in $W_k$. 
\\
\\
To determine an appropriate $k$ that (1) suppresses topic-related words and (2) retains stylistically relevant patterns as much as possible, 
we chose $k=170$ which represents the value at which the style-related tokens outnumber the topic-related tokens the most in terms of absolute frequency. 
For $k \in \{ 1, 2, \ldots, |W_k| \}$, the setting $k = 170$ satisfies conditions (1) and (2). 
Note that in Section~\ref{Results} we perform a more in-depth analysis showing which topic-related tokens occur in the \textDistortion-preprocessed documents 
and the consequences of considering higher values for $k$. 
Based on the setting $k=170$, we examined the extent to which the two topic-masked representations generated by \posNoise and \textDistortion differ from each other. For this, we list in Table~\ref{table:ComparisonTopicMasking} several example sentences taken from the documents in our test corpora, 
which show the differences regarding the outputs of both approaches. 
\begin{table} 
	\centering \footnotesize  
	\begin{tabular}{ll} \toprule   
		\textbf{Repres.} & \textbf{Original / topic-masked sentences} \\ \midrule		
		\originalCorpus    & \texttt{As an example, let us analyze the following English sentence.} \\
		\posNoise          & \texttt{As an example, let us Ø the following @ \#.} \\
		\textsf{TextDist.} & \texttt{As an *, * us * the * * *.} \\\midrule
		\originalCorpus    & \texttt{Like before, further improvements to this section are welcome.} \\
		\posNoise          & \texttt{Like before, further \# to this \# are @.} \\
		\textsf{TextDist.} & \texttt{Like *, * * to this * are *.} \\\midrule
		\originalCorpus & \texttt{I'd like to see some other editors' opinions on this question.} \\
		\posNoise       & \texttt{I'd like to see some other \#' \# on this \#.} \\
		\textsf{TextDist.} & \texttt{*'* like to see some other *' * on this *.} \\\midrule
		\originalCorpus & \texttt{Therefore we add another operator to erase this function.} \\
		\posNoise       & \texttt{Therefore we Ø another \# to Ø this \#.} \\
		\textsf{TextDist.} & \texttt{* we * another * to * this *.} \\\midrule	
		\originalCorpus & \texttt{Regarding the lexicon, the model allows for clusters.} \\
		\posNoise       & \texttt{Regarding the \#, the \# Ø for \#.} \\
		\textsf{TextDist.} & \texttt{* the *, the * * for *.} \\\midrule			
		\originalCorpus & \texttt{Look, most people have been lied to, most are...} \\		
		\posNoise       & \texttt{Look, most \# have been Ø to, most are...} \\		
		\textsf{TextDist.} & \texttt{*, most people have been * to, most are...} \\		
		\bottomrule	
	\end{tabular}
	\caption{Comparison between the resulting topic-masked representations generated by \posNoise and \textDistortion. \label{table:ComparisonTopicMasking}}
\end{table} 
It can be clearly seen that both approaches entirely mask topic-related words. 
However, in contrast to \textDistortion, our approach retains a greater number of syntactic structures including multi-word expressions 
(\quotetxt{As an example} and \quotetxt{Like before}), contractions (\quotetxt{I'd}) or sentence openers (\quotetxt{Regarding} and \quotetxt{Therefore}) that represent important stylistic features. 
Another difference that can be seen in Table~\ref{table:ComparisonTopicMasking} is that \posNoise not only \textbf{retains} stylistically relevant words and phrases occurring in the documents but also \textbf{generates} additional features \ie \posTags that increase the diversity of the documents feature space. Depending on the considered \AV method, a variety of feature compositions can be derived from \posNoise representations, which include, for instance, 
\posTags with preceding/succeeding punctuation marks or \posTags surrounded by function words. Such feature compositions can play a decisive role in the prediction of an \AV method and are therefore desirable. 

\subsection{Results} \label{Results}
\begin{figure*} [t]
	\centering
	\includegraphics[width=0.3\textwidth]{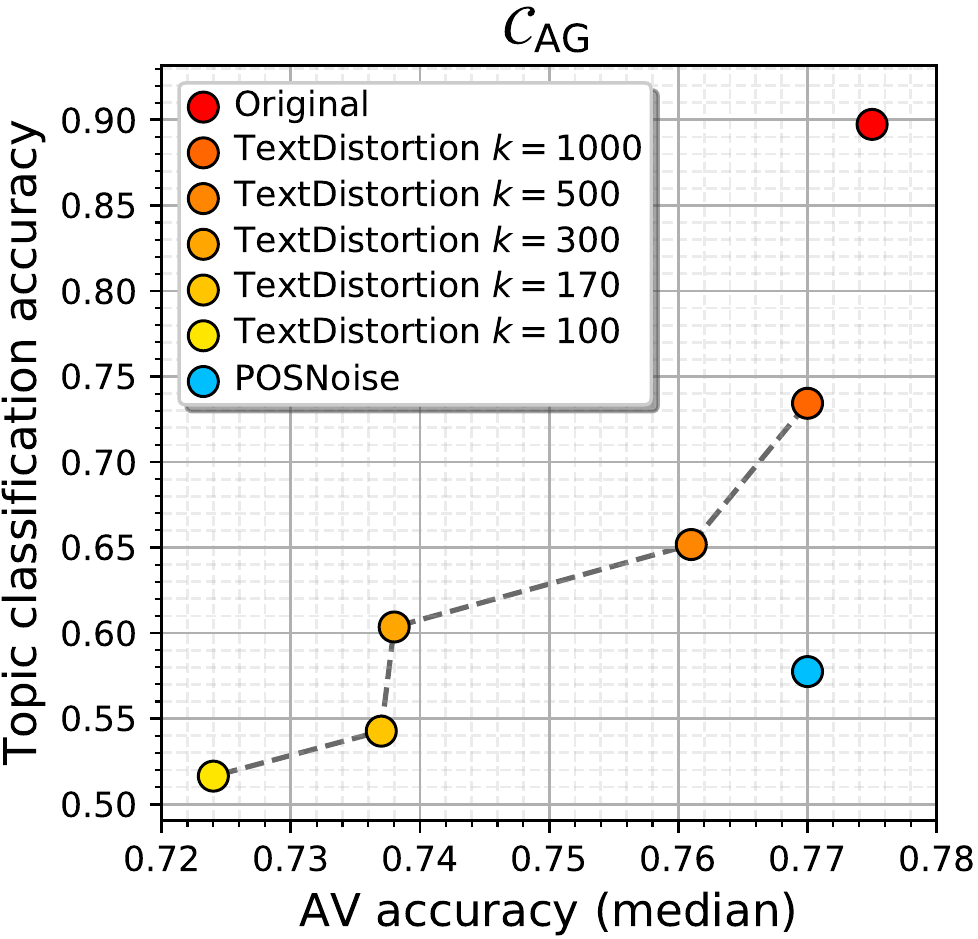}
	\hspace{.5cm}
	\includegraphics[width=0.3\textwidth]{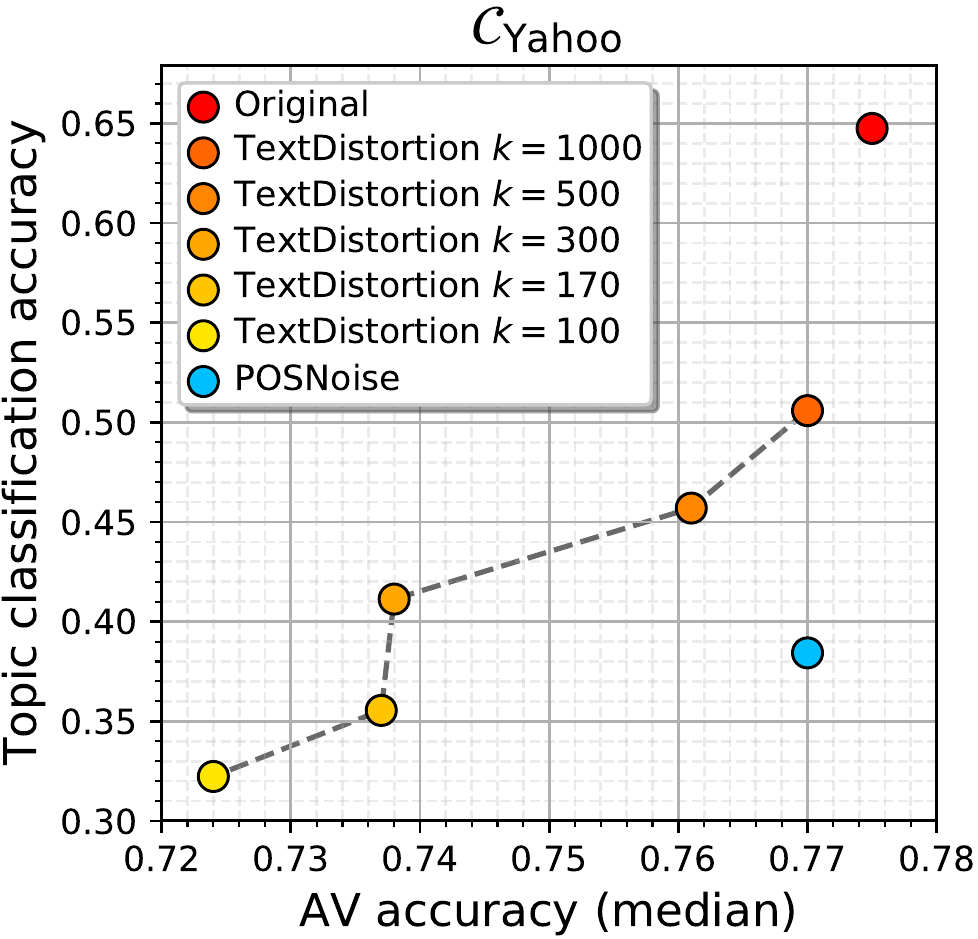}
	\hspace{.5cm}
	\includegraphics[width=0.3\textwidth]{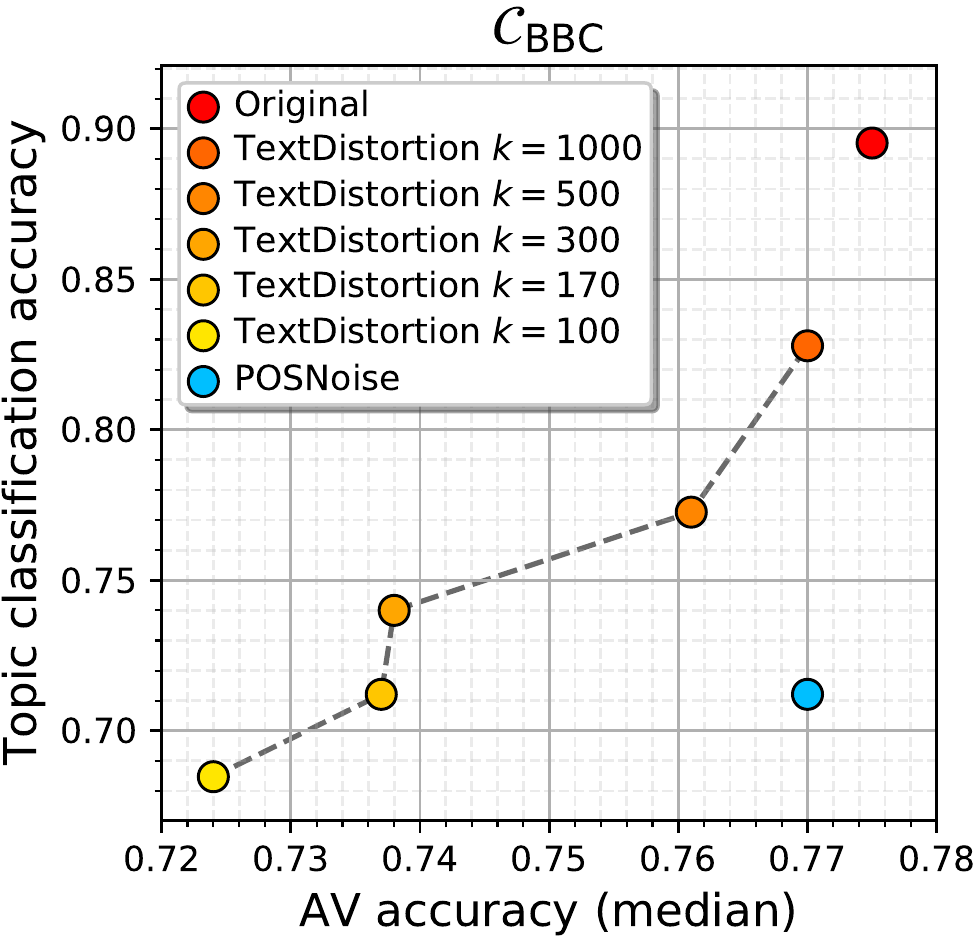}
	\caption{Comparison between topic and \AV classification results. \label{Accuracy_Topic_vs_AV}}
\end{figure*}

After training and optimizing all \AV methods, we applied these to the respective test partitions of the \originalCorpus, \posNoise and \textDistortion corpora. The overall results are shown in Table~\ref{tab:ExperimentComparisonResults}, where a compact summary with respect to the average/median improvements of \posNoise over \textDistortion is provided in Table~\ref{table:ImprovementsSummary}. 
In what follows, we focus on the results between \posNoise and \textDistortion (using $k=170$) in Table~\ref{tab:ExperimentComparisonResults}, which are highlighted in yellow. 
As can be seen from this table, \posNoise leads to better results than \textDistortion in terms of accuracy and \auc in 34 and 32 of 42 cases (81\% and 76\%, respectively). In total, there are 4 ties in terms of accuracy, where in two cases \posNoise leads to better \auc results, which we selected as a secondary performance measure. 
Besides the setting $k=170$, we further evaluated the six \AV methods on additional \textDistortion-preprocessed corpora using the settings 100, 300, 500 and 1000. A closer look at the results in Table~\ref{tab:ExperimentComparisonResults} shows that \textDistortion leads to increasingly larger accuracy improvements the higher the setting of $k$ is. What is not reflected in these results, however, is at what price the \quote{improvements} occur. 
\\
\\
To gain a better understanding of how higher settings for $k$ affect the performance of the methods, we first show how much topic-related text units remain in the documents preprocessed by \textDistortion. First, we concatenated all documents in each \textDistortion corpus into a single text $\D$. Next, we tokenized $\D$ and subtracted from the resulting list all words that appear in our pattern list $\listStylePatterns$. 
The remaining words and patterns can be inspected in detail in the word clouds illustrated in Table~\ref{tbl:TextDistortion_Influence_of_K}. 
As can be seen from these word clouds, the document representations masked by \textDistortion retain a wide variety of topic-related words 
(\eg \quotetxt{system}, \quotetxt{data}, \quotetxt{information}, etc.) the greater $k$ increases, which are not present in the \posNoise representations. 
The presence of these topic-related text units provides a first indication that the improvement in verification results can be attributed to them. 
To investigate this further, we proceed on the basis of the following assumption: If higher $k$-values are related to the topic of the texts, this should be reflected in the context of a topic classification with correspondingly high results. Low $k$-values, on the other hand, should lead to low topic classification results. 
As a first step, we selected three well-known benchmark corpora for topic classification namely: 
\e{AG's News Topic Classification Dataset} \cite{GulliAGNewsArticles:2005} (denoted by $\CorpusAGNews$), 
\e{BBC News Dataset} \cite{GreeneCunninghamBBCDataset:2006} (denoted by $\CorpusBBC$) and 
\e{Yahoo! Answers Topic Classification Dataset} \cite{ZhangCNNtextClassifiction:2015} (denoted by $\CorpusYahoo$), 
where all corpora were left in their original form (\ie no subsampling or subsets were considered) in order to allow reproducibility of our results.
Next, we trained a standard logistic regression classifier\footnote{We used the \e{Logistic Regression} classifier implementation available at \url{https://scikit-learn.org} with the default settings.} (on the basis of tokens as features) using the \originalCorpus, \posNoise and \textDistortion representations of $\CorpusAGNews$, $\CorpusYahoo$ and $\CorpusBBC$, where for \textDistortion we selected $k \in \{ 100, 170, 300, 500, 1000 \}$. 
The results of the topic classifier (based on 5-fold cross-validation) for each $\Corpus \in \{ \CorpusAGNews, \CorpusYahoo, \CorpusBBC \}$ along with its \originalCorpus, \posNoise and \textDistortion representations are visualized against the median of the \AV results of the six methods in the scatter plots shown in Figure~\ref{Accuracy_Topic_vs_AV}. 
As can be seen from the three scatter plots, the \textDistortion representations lead to higher \AV classification results, 
but at the same time also to higher topic classification results as $k$ increases. 
Conversely, lower $k$-values lead to both lower topic and \AV results. 
In contrast, \posNoise approaches more closely the best possible compromise between topic and style (across $\CorpusAGNews, \CorpusYahoo, \CorpusBBC$), which is located at the very bottom right. 
\\
\\
When comparing the results of topic classification with respect to the \originalCorpus and the \posNoise corpora, one can further observe that the degree of topic-related text units is significantly high regarding the former (up to $\approx$35\% higher). This shows that \posNoise leads to a substantial reduction of topic-related text units without strongly affecting the style-related classification performance. 
It should be noted that we consider the absolute accuracy values as not significant for this experiment, since in all three topic corpora the function words alone already represent a strong classification signal. This is reflected by the fact that restricting the topic classifier exclusively on function words still yields high accuracy scores of 56\% ($\CorpusAGNews$), 35\% ($\CorpusYahoo$) and 70\% ($\CorpusBBC$), which are comparable to \posNoise.
Separately, one can see from the three scatter plots that regarding \textDistortion, the setting $k=170$ in fact represents a good compromise between topic and style. While the median \AV classification results are almost identical ($\approx$1\% variance in terms of accuracy) the topic classification results vary from 3 to 6\%. Overall, this observation justifies our decision of choosing this setting for \textDistortion to be able to compare both approaches. 
\begin{table*} [h!]  
	\centering \small
	\setlength\tabcolsep{-0.05cm}
	\begin{tabularx}{\textwidth}{c@{\hspace{0.3cm}}ccccccc}
		\toprule  
		$\bm{k}$ &  $\bm{\CorpusGutenberg}$ & $\bm{\CorpusWikiSockpuppets}$ & $\bm{\CorpusACL}$ & $\bm{\CorpusPeeJ}$ & $\bm{\CorpusTelegraph}$ & $\bm{\CorpusApricity}$ & $\bm{\CorpusReddit}$  \\\midrule
		100 &
		\begin{minipage}{.142\textwidth} \includegraphics[width=0.9\textwidth,trim=0cm 3.4cm 0cm 0cm,clip]{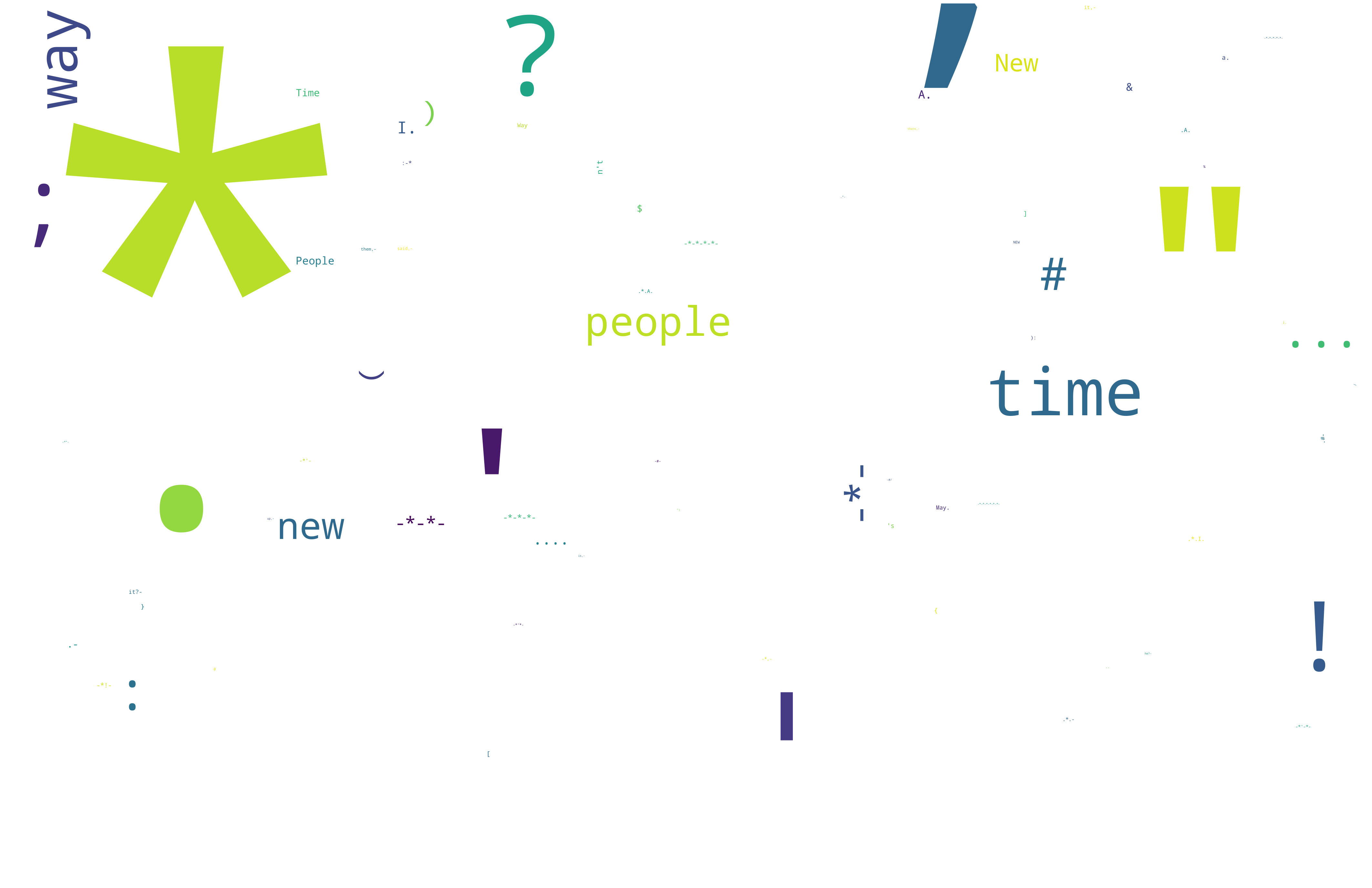} \end{minipage}  &
		\begin{minipage}{.142\textwidth} \includegraphics[width=0.9\textwidth,trim=0cm 3.4cm 0cm 0cm,clip]{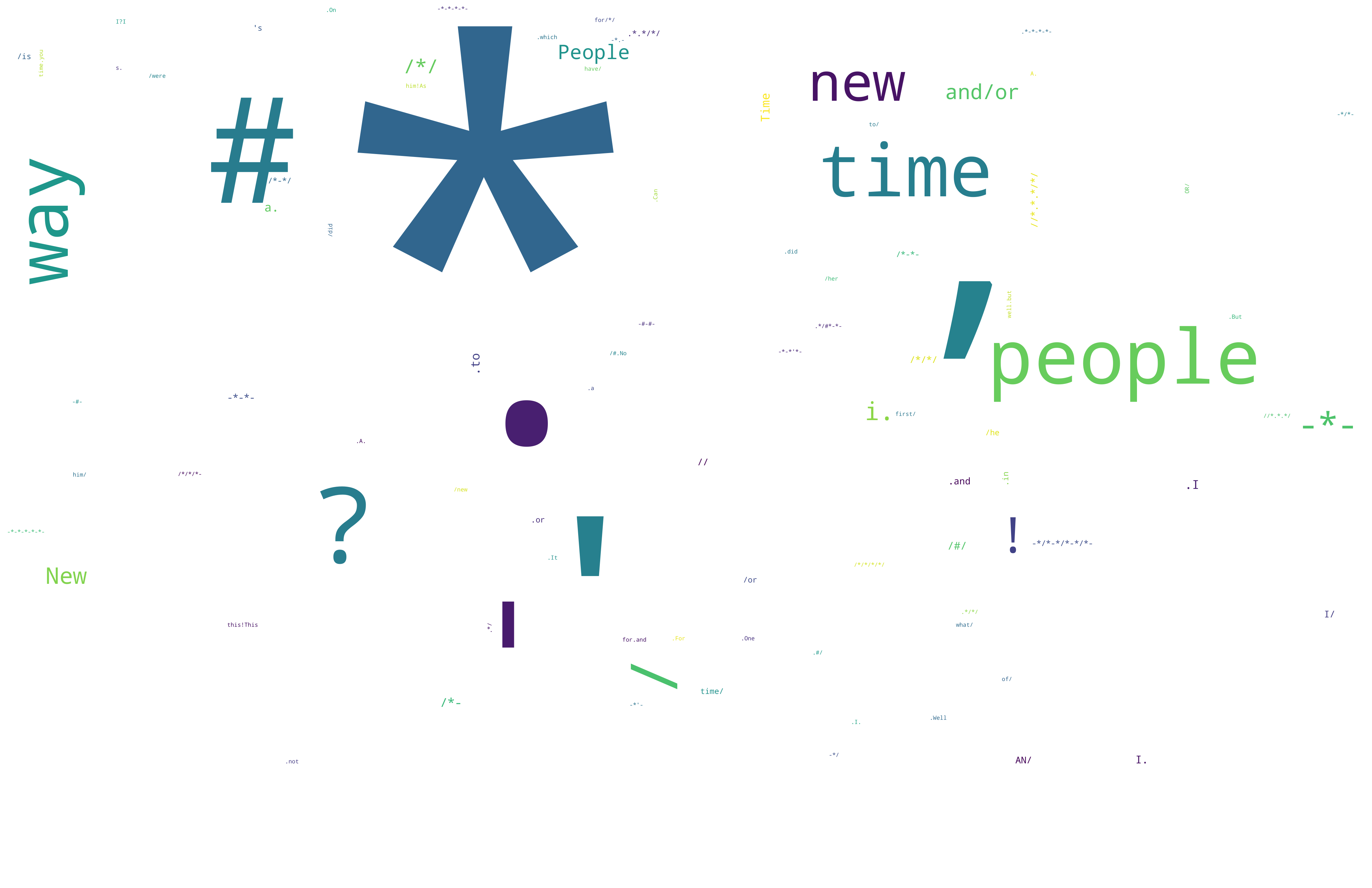} \end{minipage}  &	
		\begin{minipage}{.142\textwidth} \includegraphics[width=0.9\textwidth,trim=0cm 3.4cm 0cm 0cm,clip]{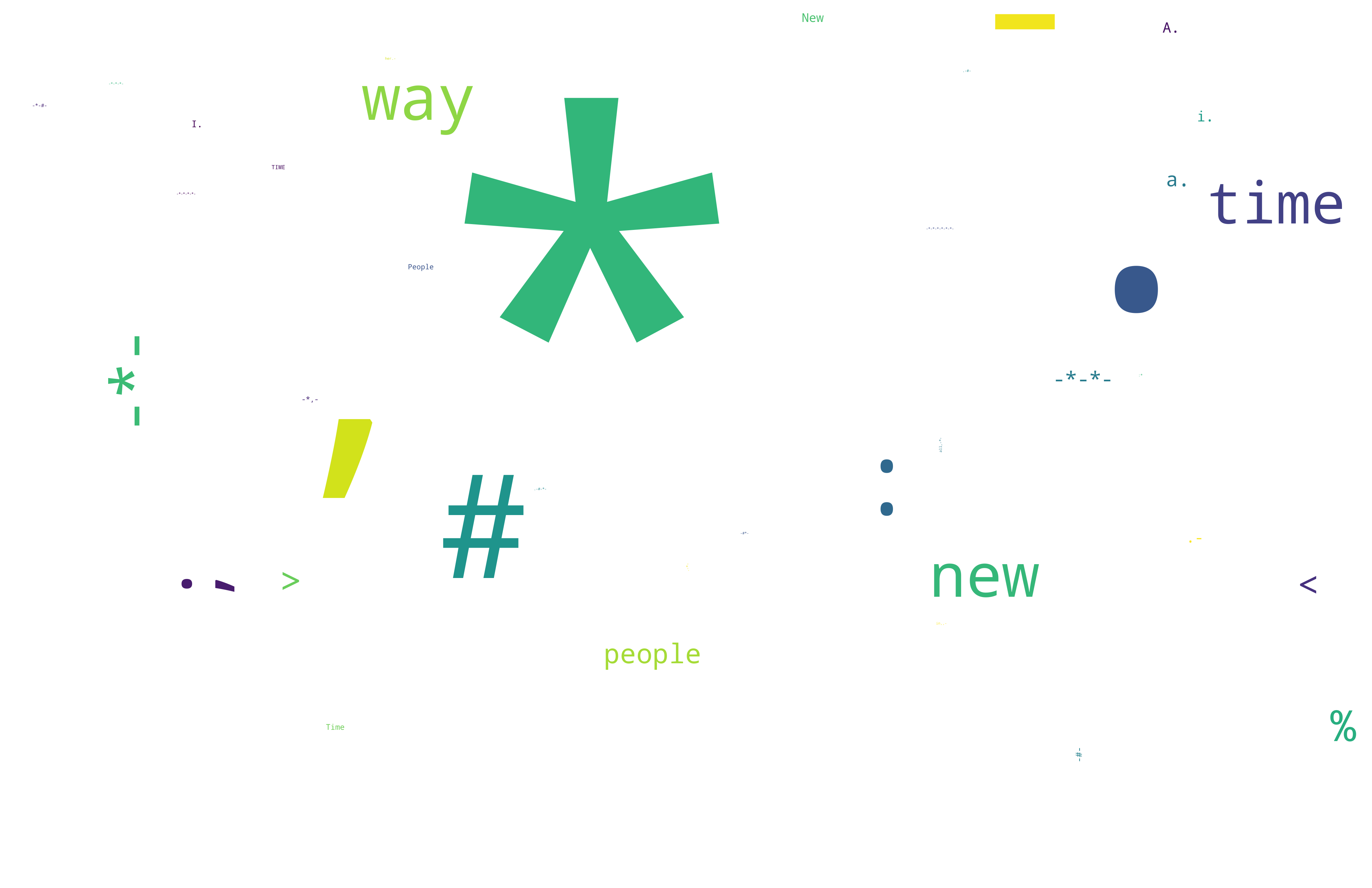} \end{minipage}  &
		\begin{minipage}{.142\textwidth} \includegraphics[width=0.9\textwidth,trim=0cm 3.4cm 0cm 0cm,clip]{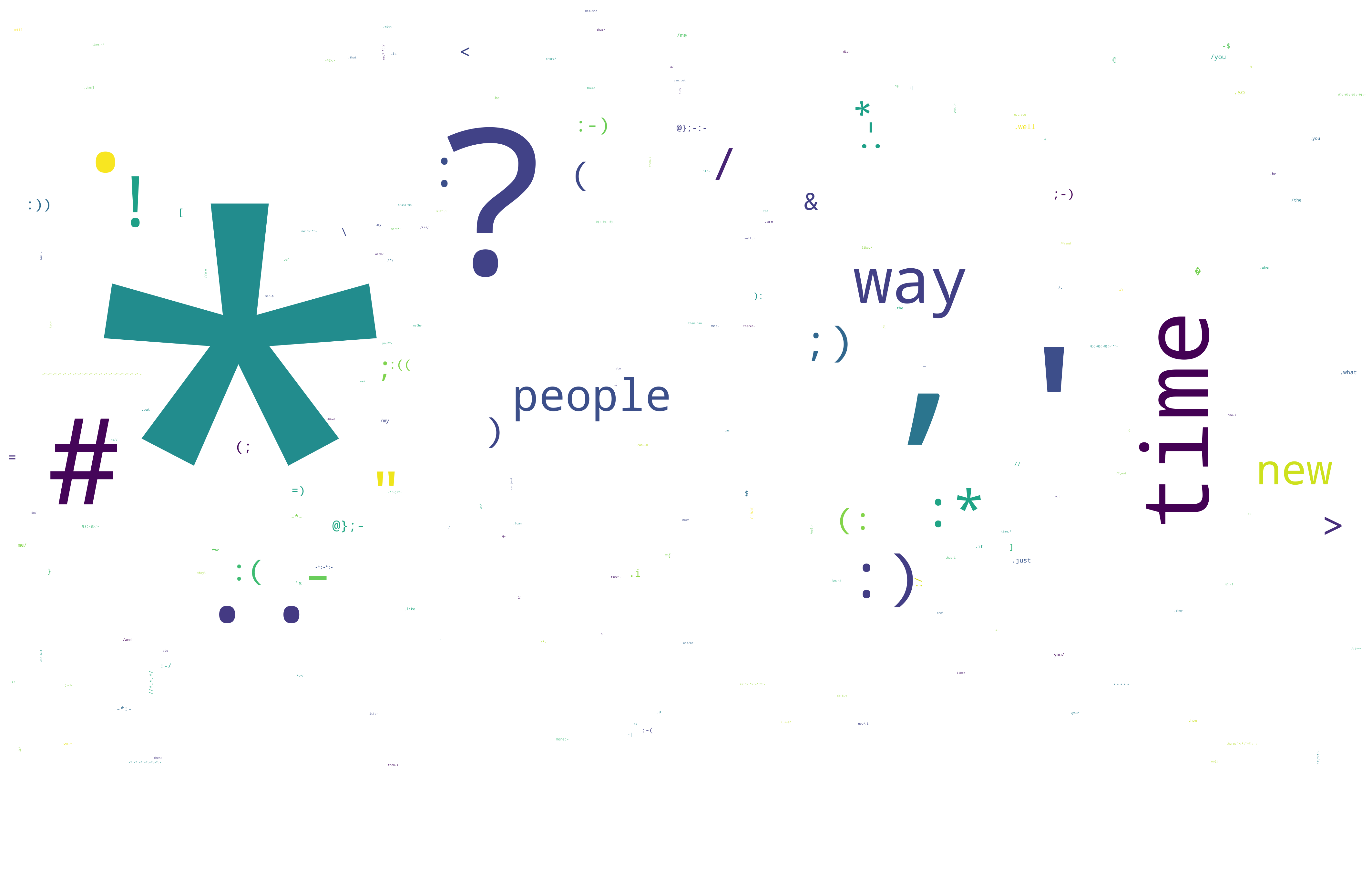} \end{minipage}  &         
		\begin{minipage}{.142\textwidth} \includegraphics[width=0.9\textwidth,trim=0cm 3.4cm 0cm 0cm,clip]{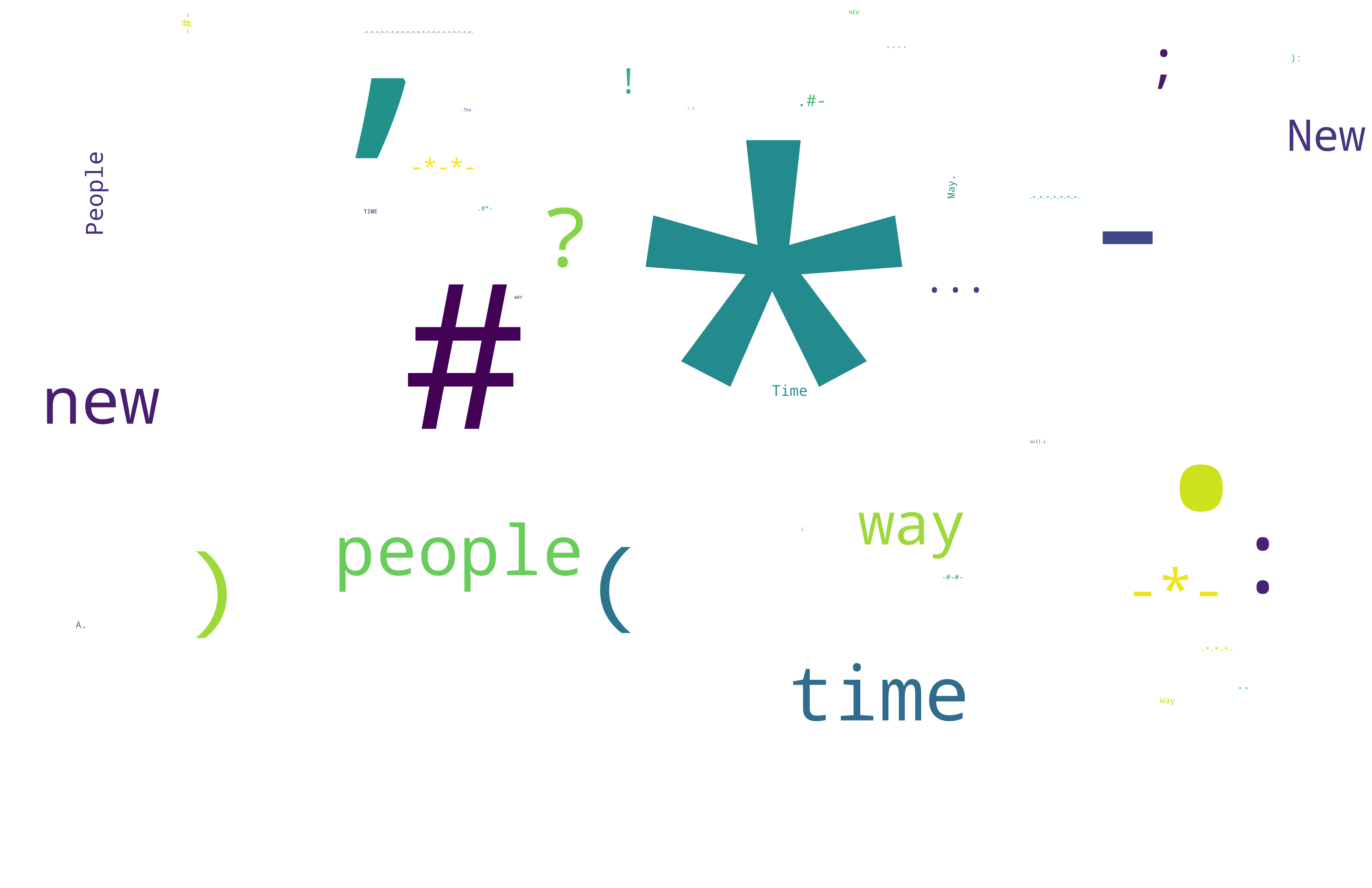} \end{minipage} &
		\begin{minipage}{.142\textwidth} \includegraphics[width=0.9\textwidth,trim=0cm 3.4cm 0cm 0cm,clip]{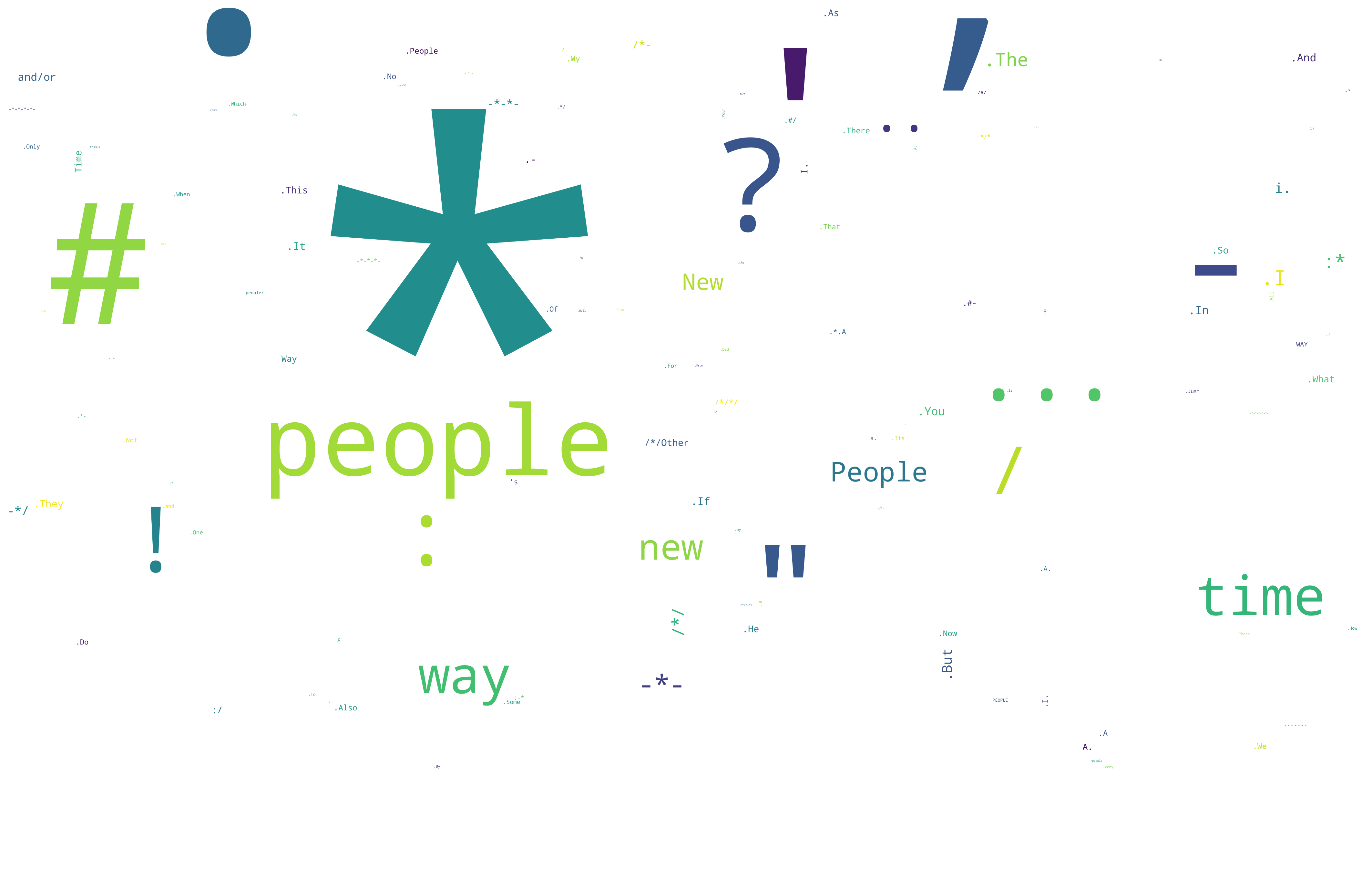} \end{minipage} &  
		\begin{minipage}{.142\textwidth} \includegraphics[width=0.9\textwidth,trim=0cm 3.4cm 0cm 0cm,clip]{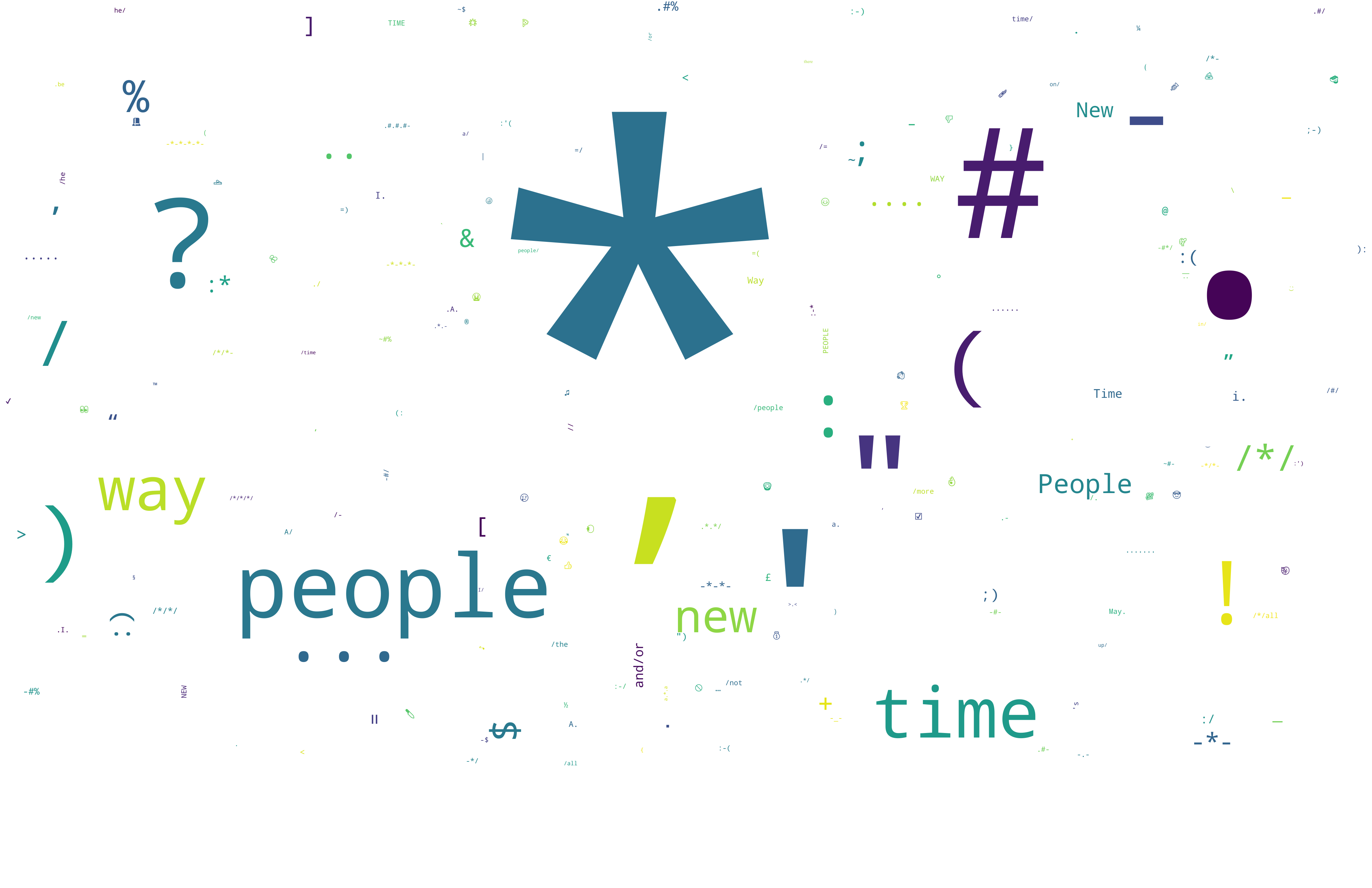} \end{minipage} \\
		\midrule		
		170 &
		\begin{minipage}{.142\textwidth} \includegraphics[width=0.9\textwidth,trim=0cm 3.4cm 0cm 0cm,clip]{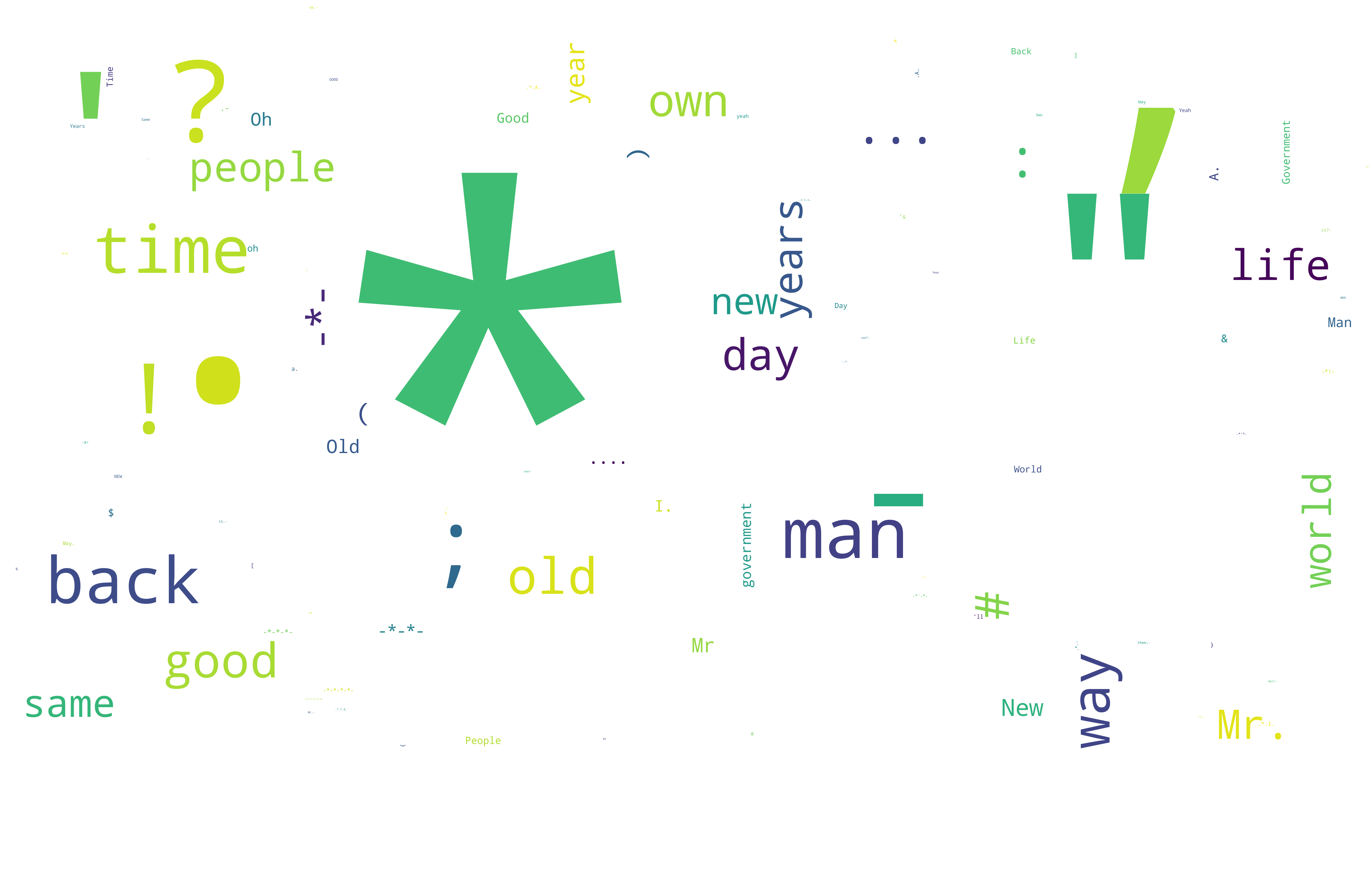} \end{minipage}  &
		\begin{minipage}{.142\textwidth} \includegraphics[width=0.9\textwidth,trim=0cm 3.4cm 0cm 0cm,clip]{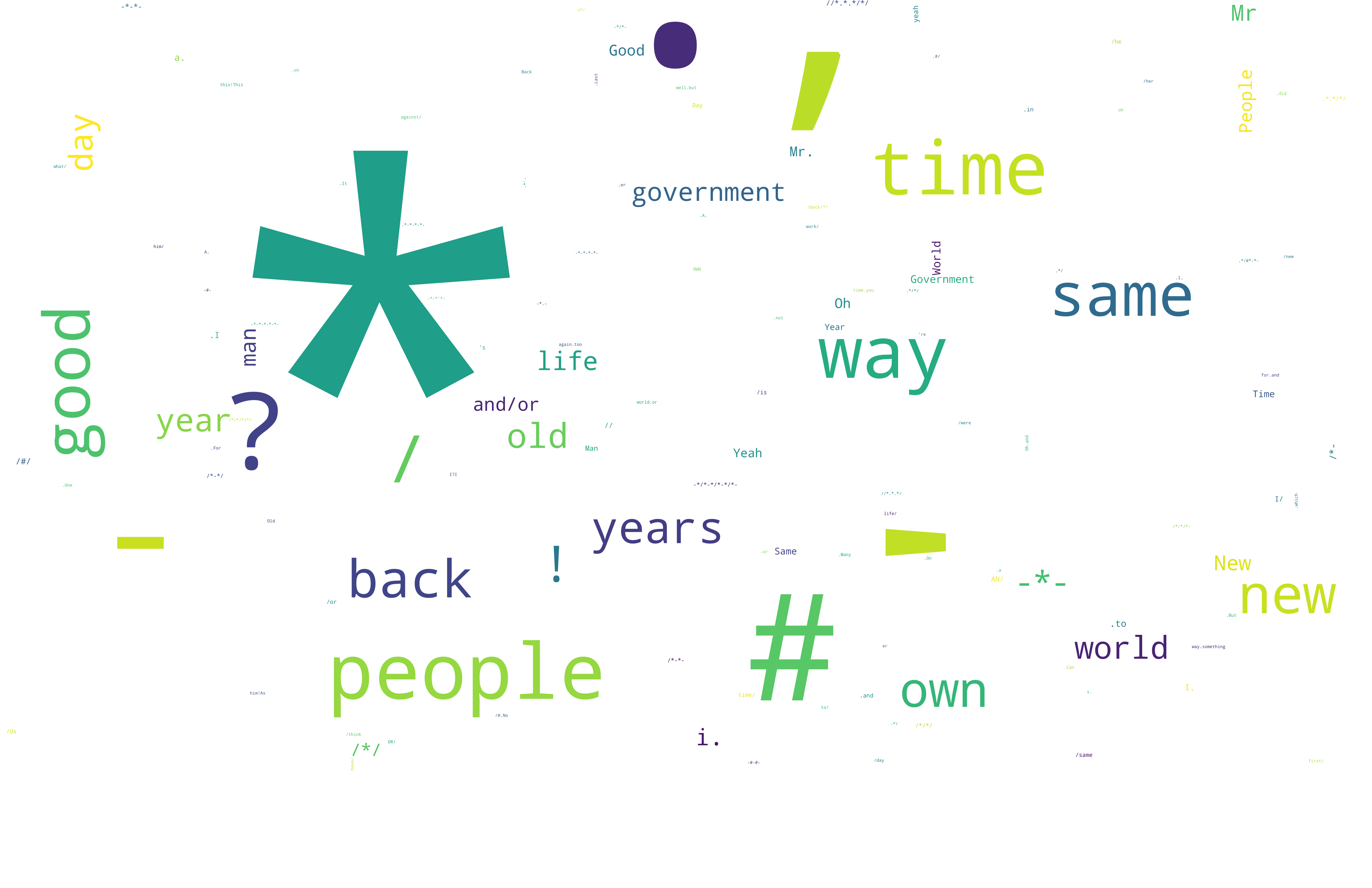} \end{minipage}  &
		\begin{minipage}{.142\textwidth} \includegraphics[width=0.9\textwidth,trim=0cm 3.4cm 0cm 0cm,clip]{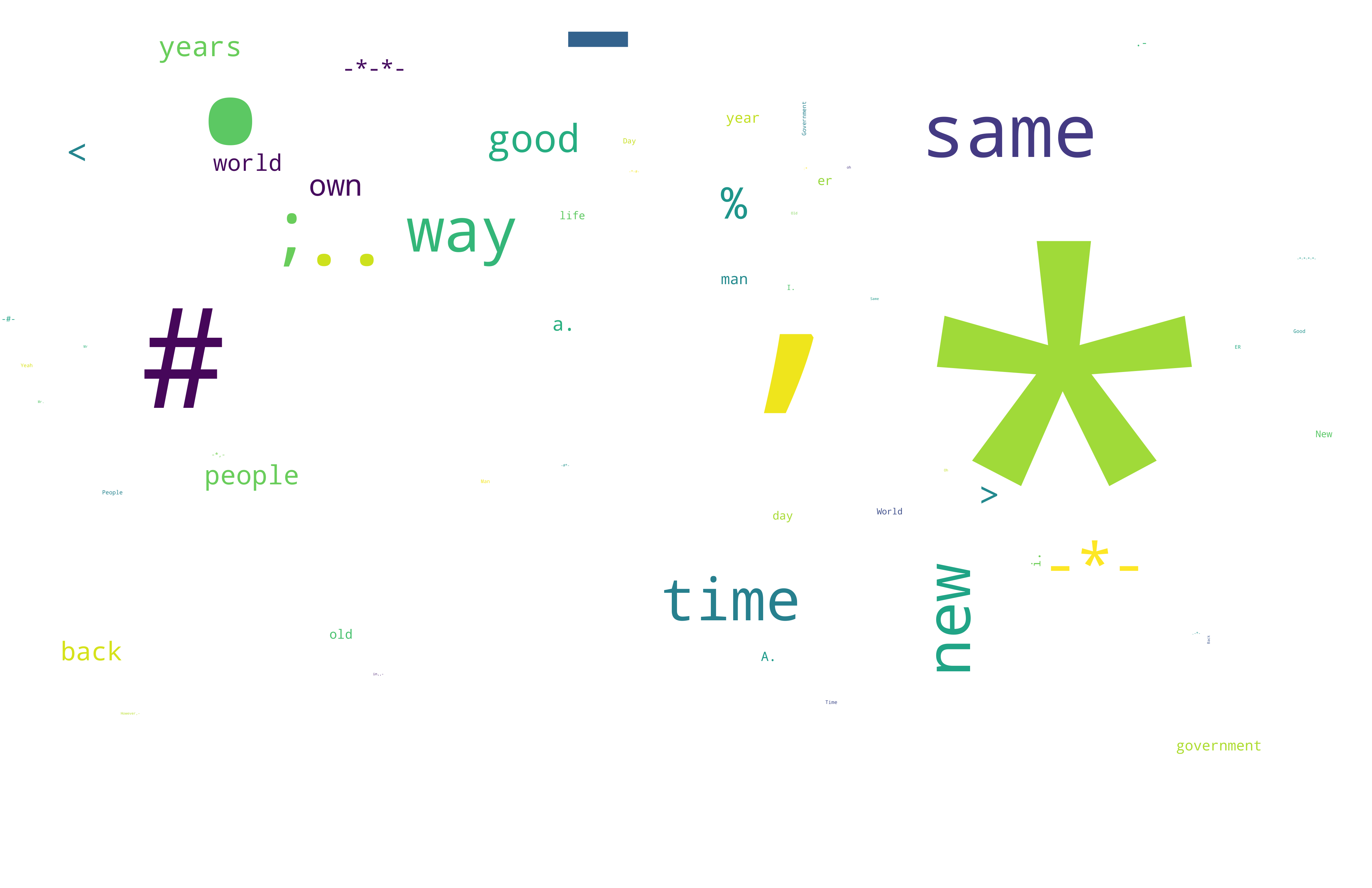} \end{minipage}  &
		\begin{minipage}{.142\textwidth} \includegraphics[width=0.9\textwidth,trim=0cm 3.4cm 0cm 0cm,clip]{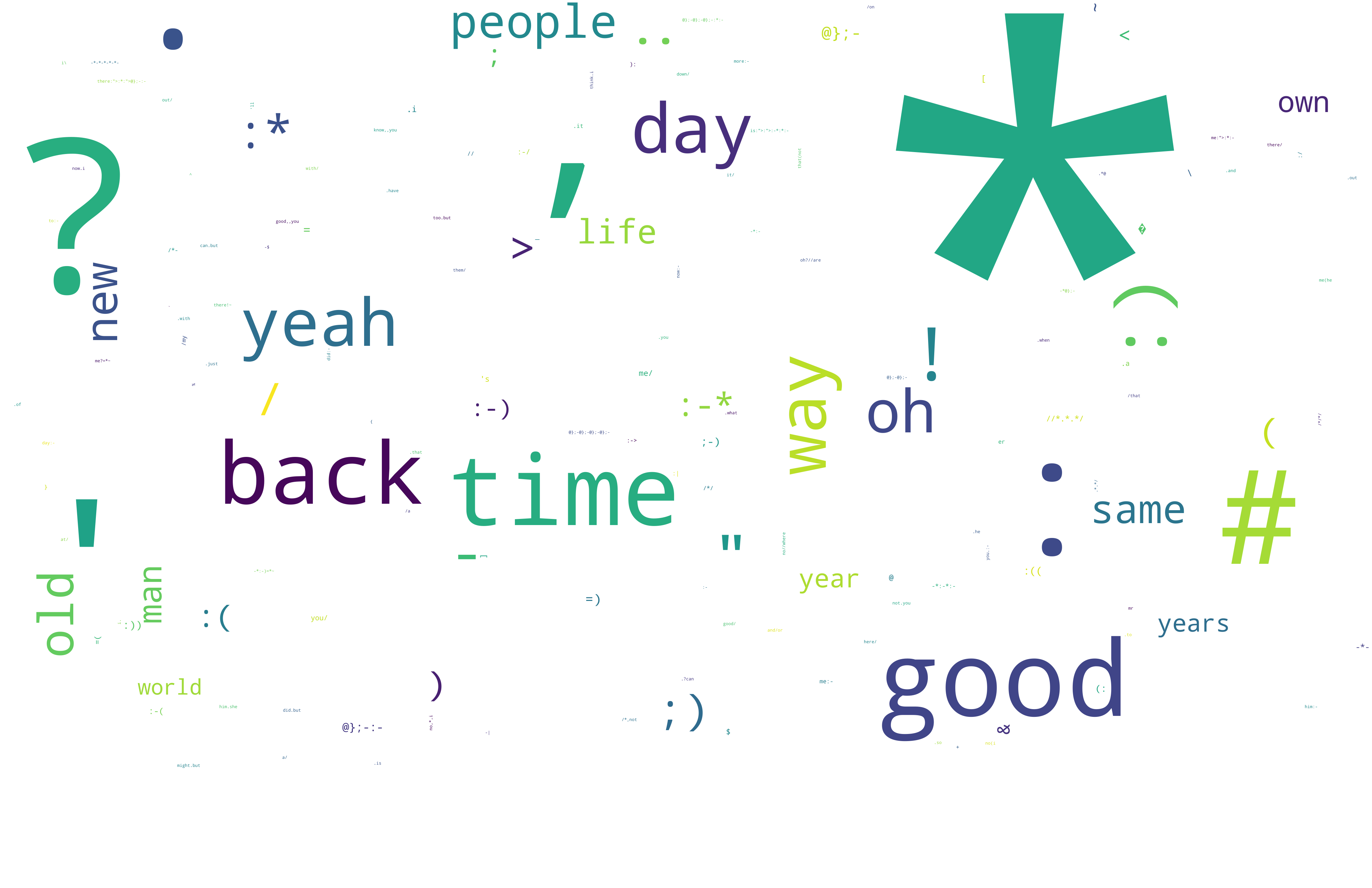} \end{minipage}  &
		\begin{minipage}{.142\textwidth} \includegraphics[width=0.9\textwidth,trim=0cm 3.4cm 0cm 0cm,clip]{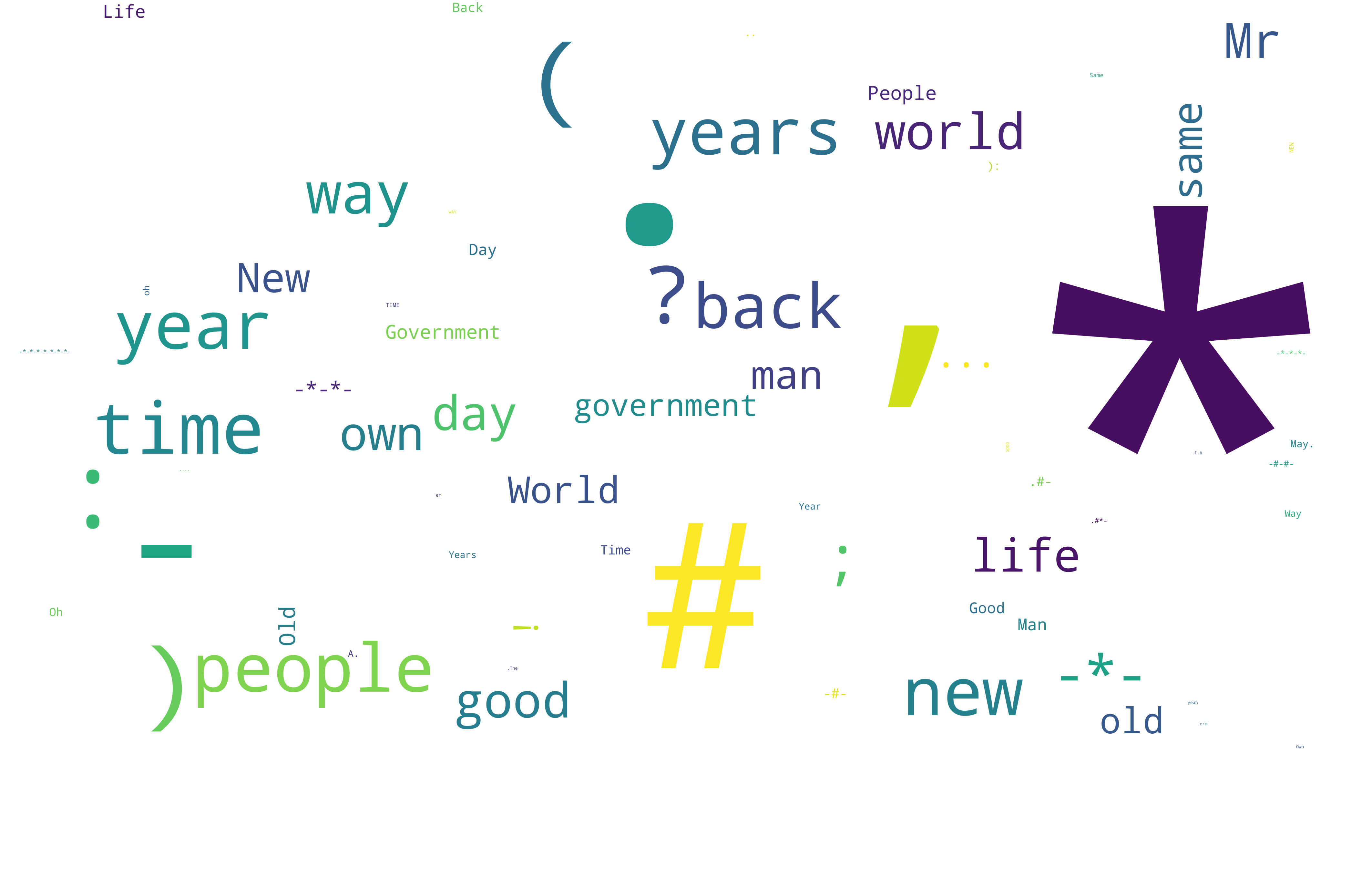} \end{minipage}  &	
		\begin{minipage}{.142\textwidth} \includegraphics[width=0.9\textwidth,trim=0cm 3.4cm 0cm 0cm,clip]{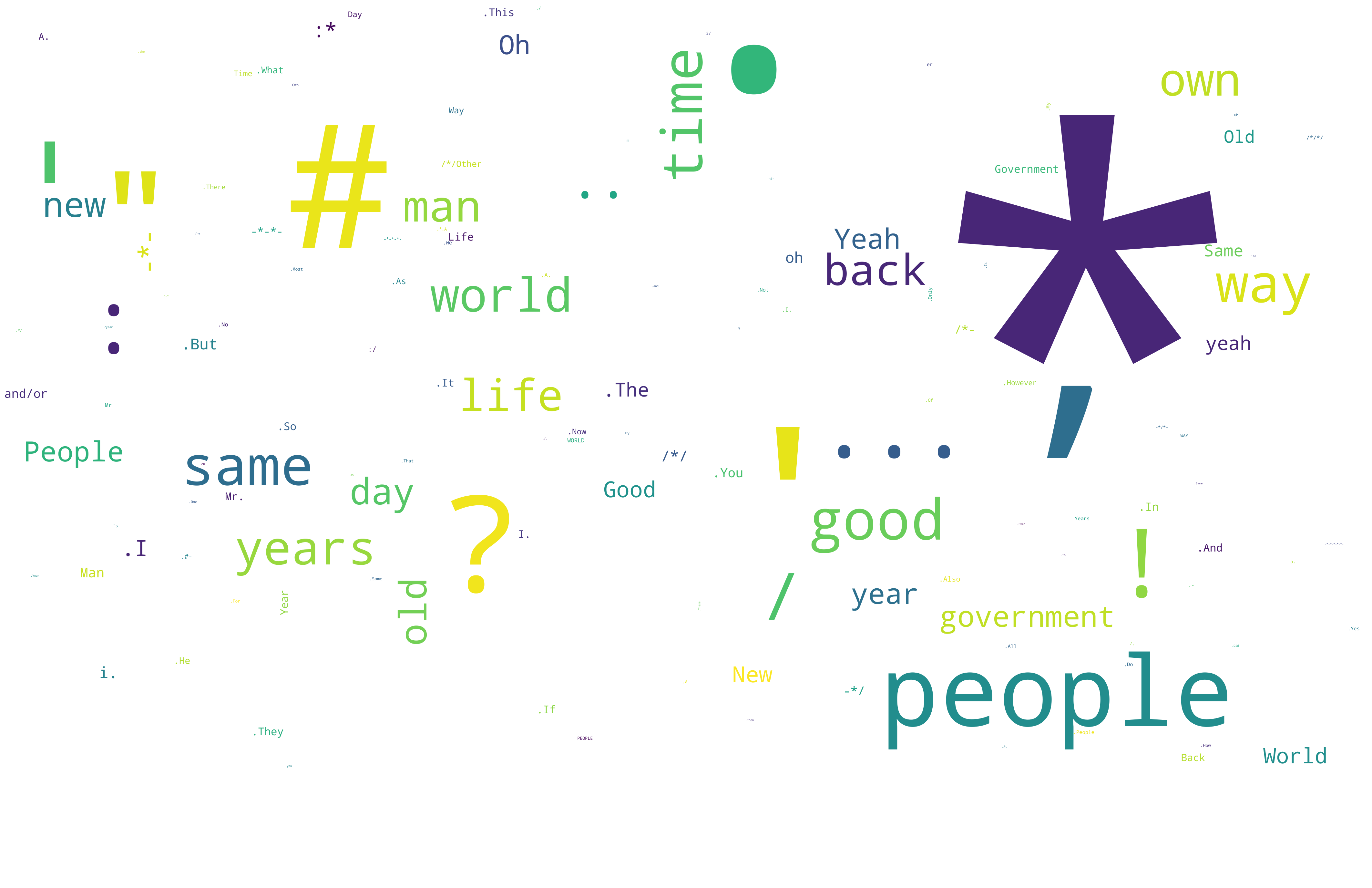} \end{minipage}  & 
		\begin{minipage}{.142\textwidth} \includegraphics[width=0.9\textwidth,trim=0cm 3.4cm 0cm 0cm,clip]{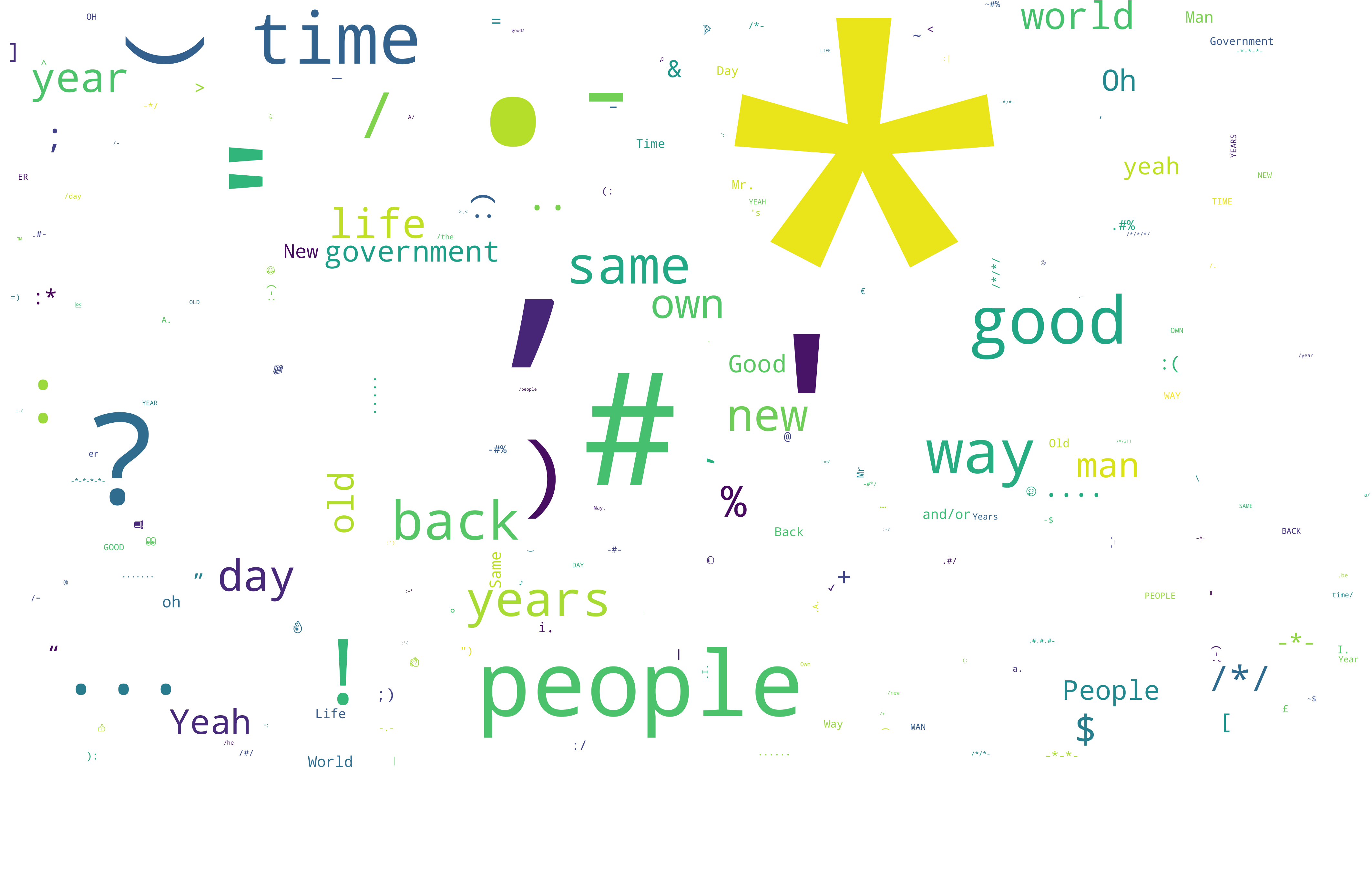} \end{minipage} \\	
		\midrule		
		300 &
		\begin{minipage}{.142\textwidth} \includegraphics[width=0.9\textwidth,trim=0cm 3.4cm 0cm 0cm,clip]{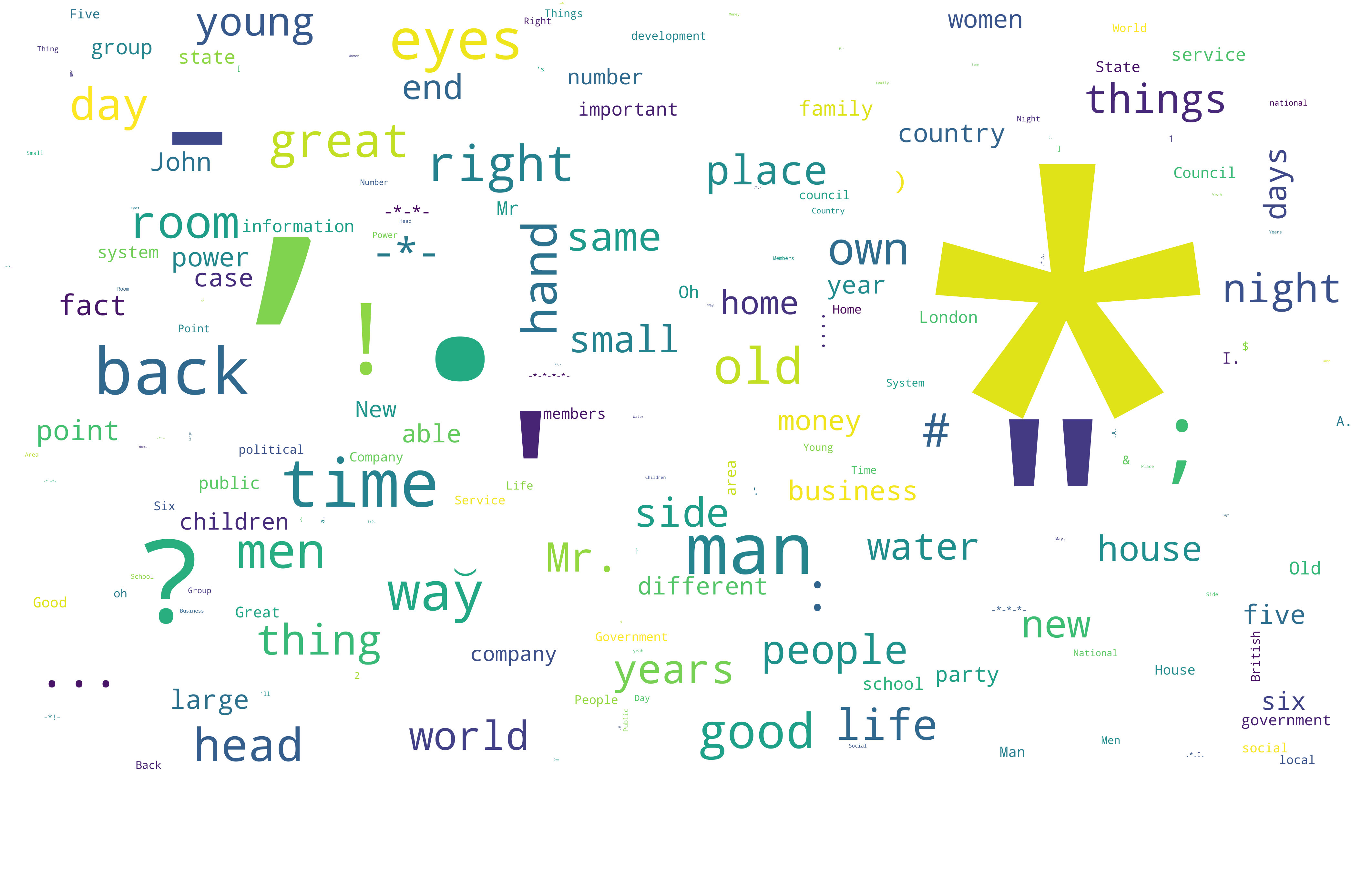} \end{minipage}  &
		\begin{minipage}{.142\textwidth} \includegraphics[width=0.9\textwidth,trim=0cm 3.4cm 0cm 0cm,clip]{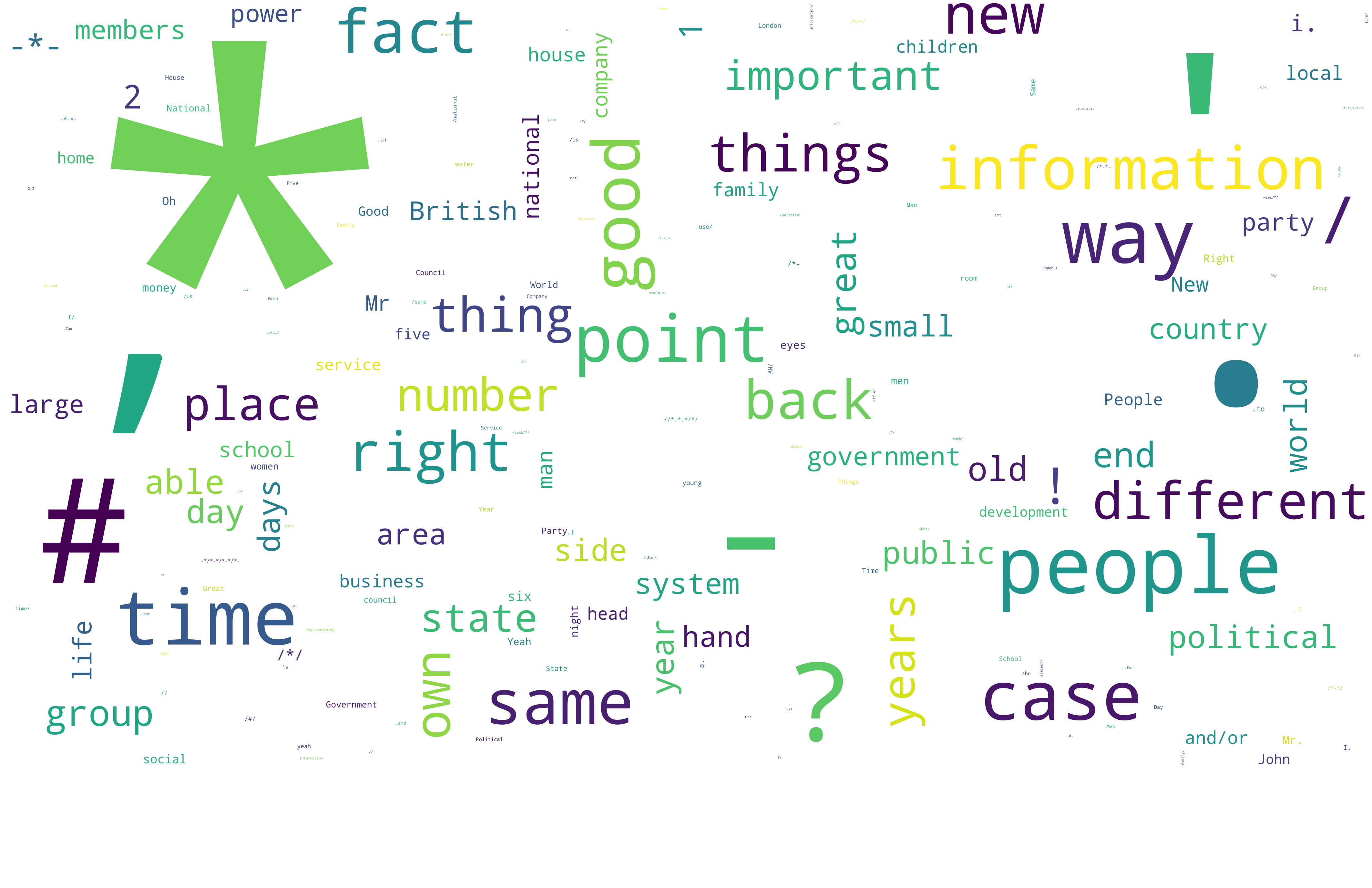} \end{minipage}  &
		\begin{minipage}{.142\textwidth} \includegraphics[width=0.9\textwidth,trim=0cm 3.4cm 0cm 0cm,clip]{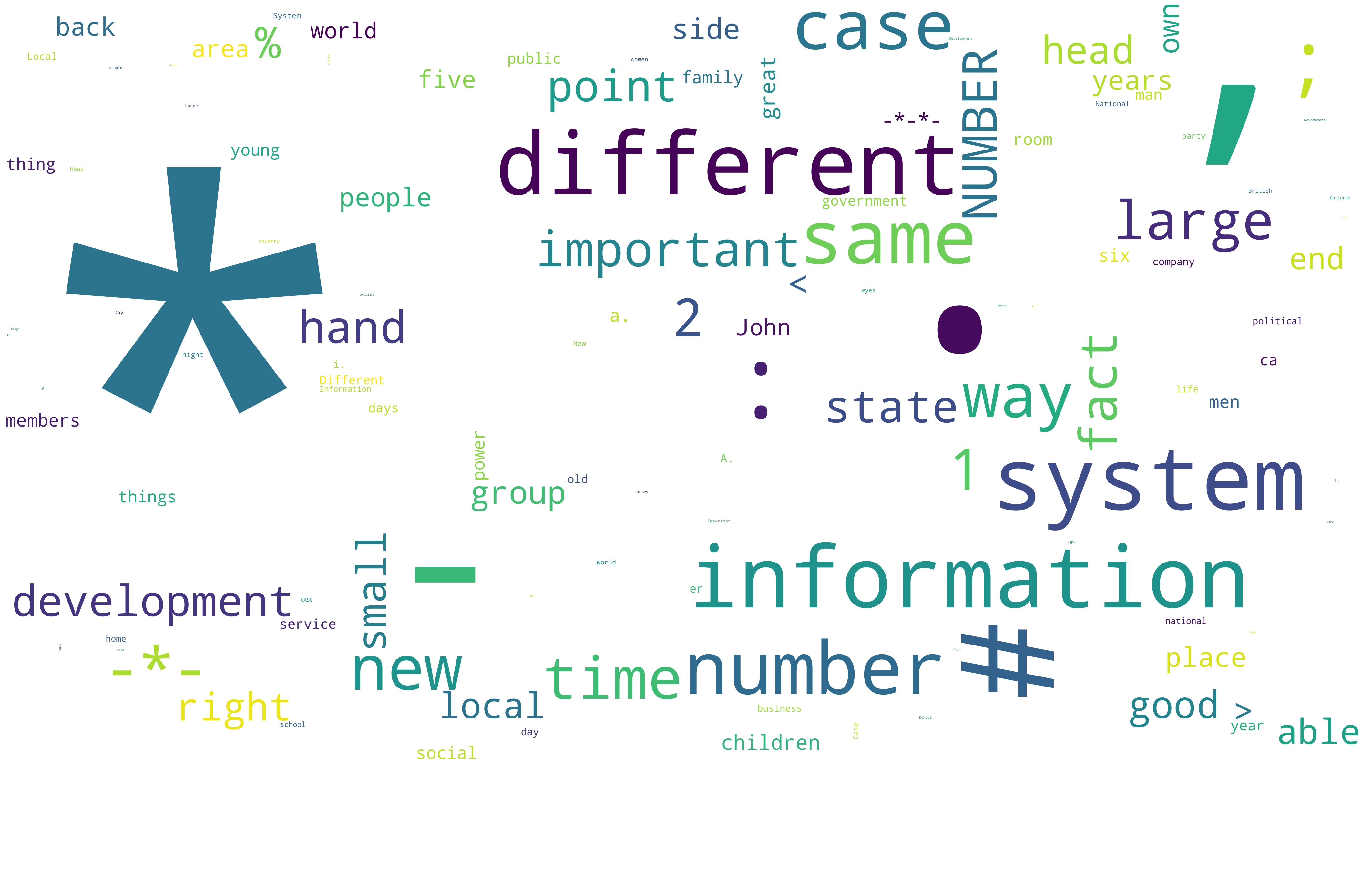} \end{minipage}  &
		\begin{minipage}{.142\textwidth} \includegraphics[width=0.9\textwidth,trim=0cm 3.4cm 0cm 0cm,clip]{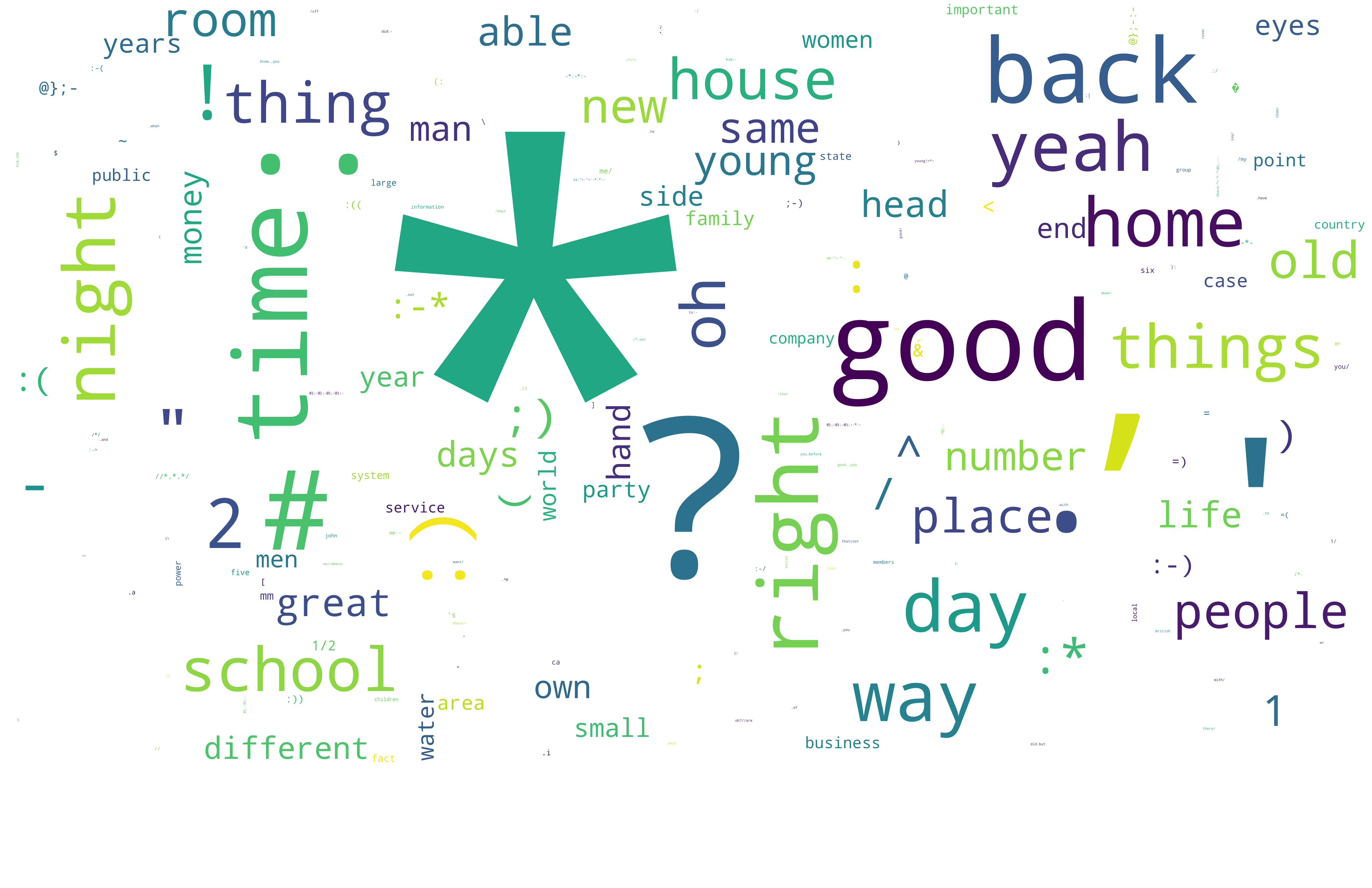} \end{minipage}  &
		\begin{minipage}{.142\textwidth} \includegraphics[width=0.9\textwidth,trim=0cm 3.4cm 0cm 0cm,clip]{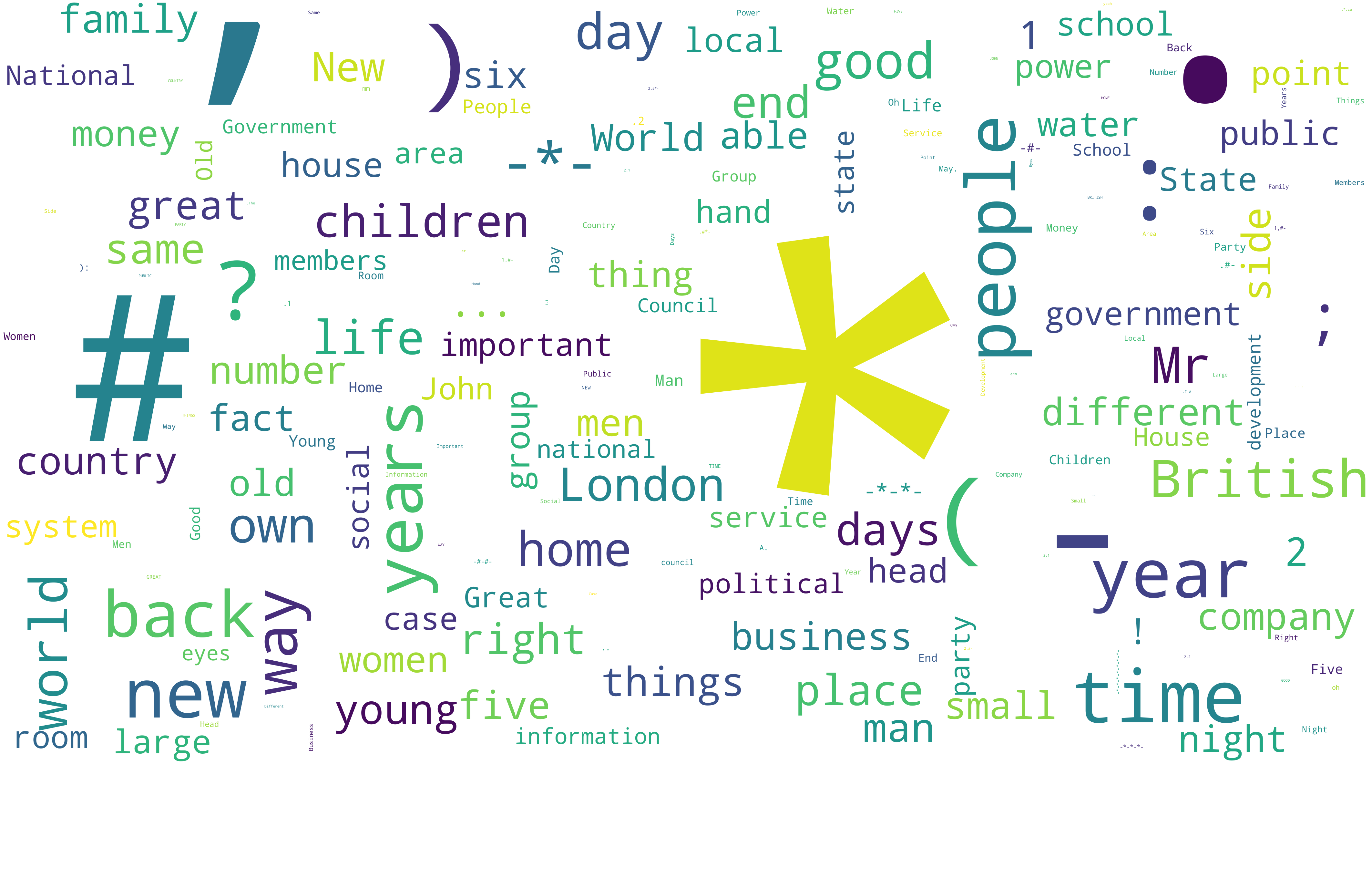} \end{minipage}  &	
		\begin{minipage}{.142\textwidth} \includegraphics[width=0.9\textwidth,trim=0cm 3.4cm 0cm 0cm,clip]{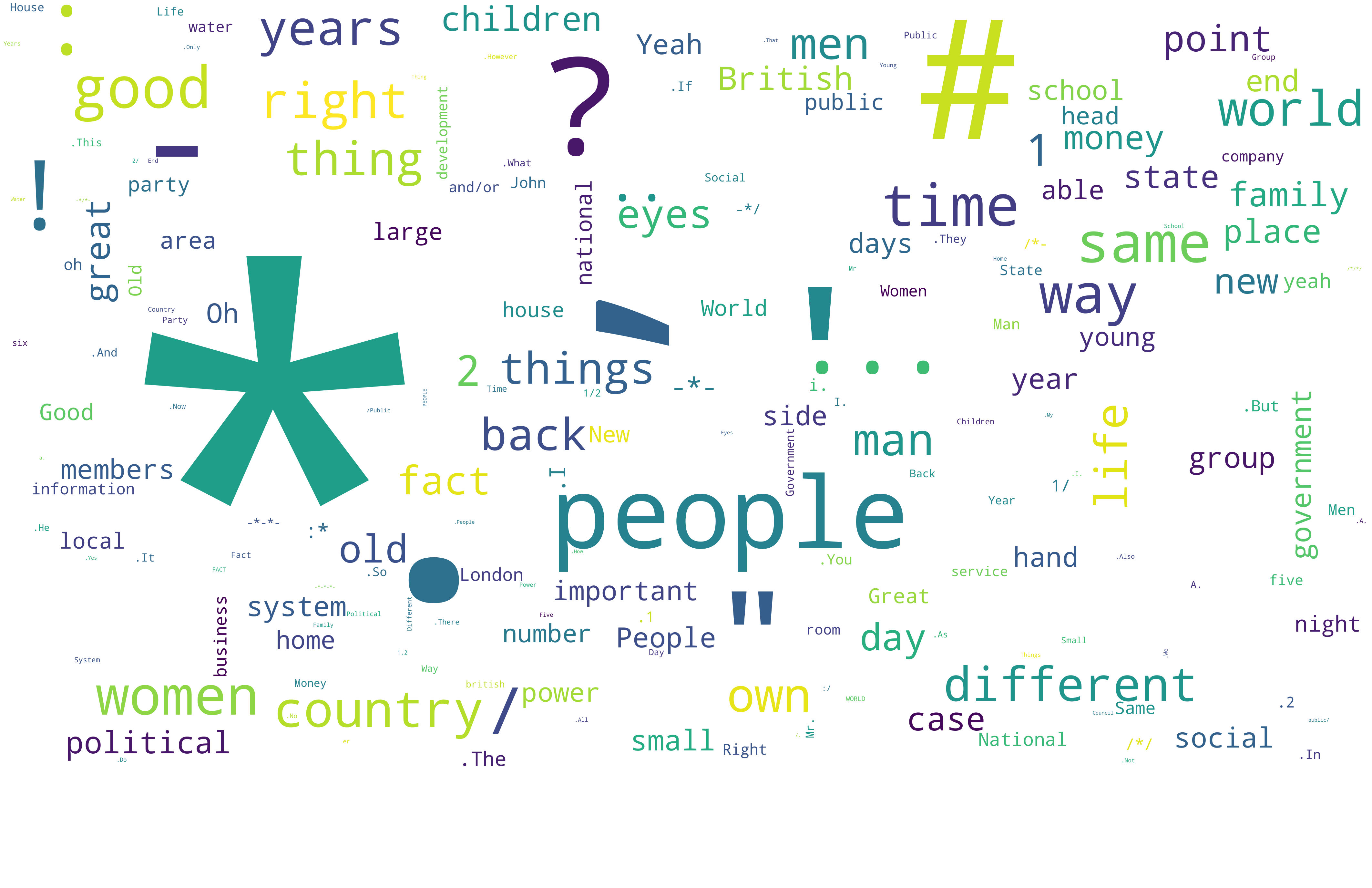} \end{minipage}  & 
		\begin{minipage}{.142\textwidth} \includegraphics[width=0.9\textwidth,trim=0cm 3.4cm 0cm 0cm,clip]{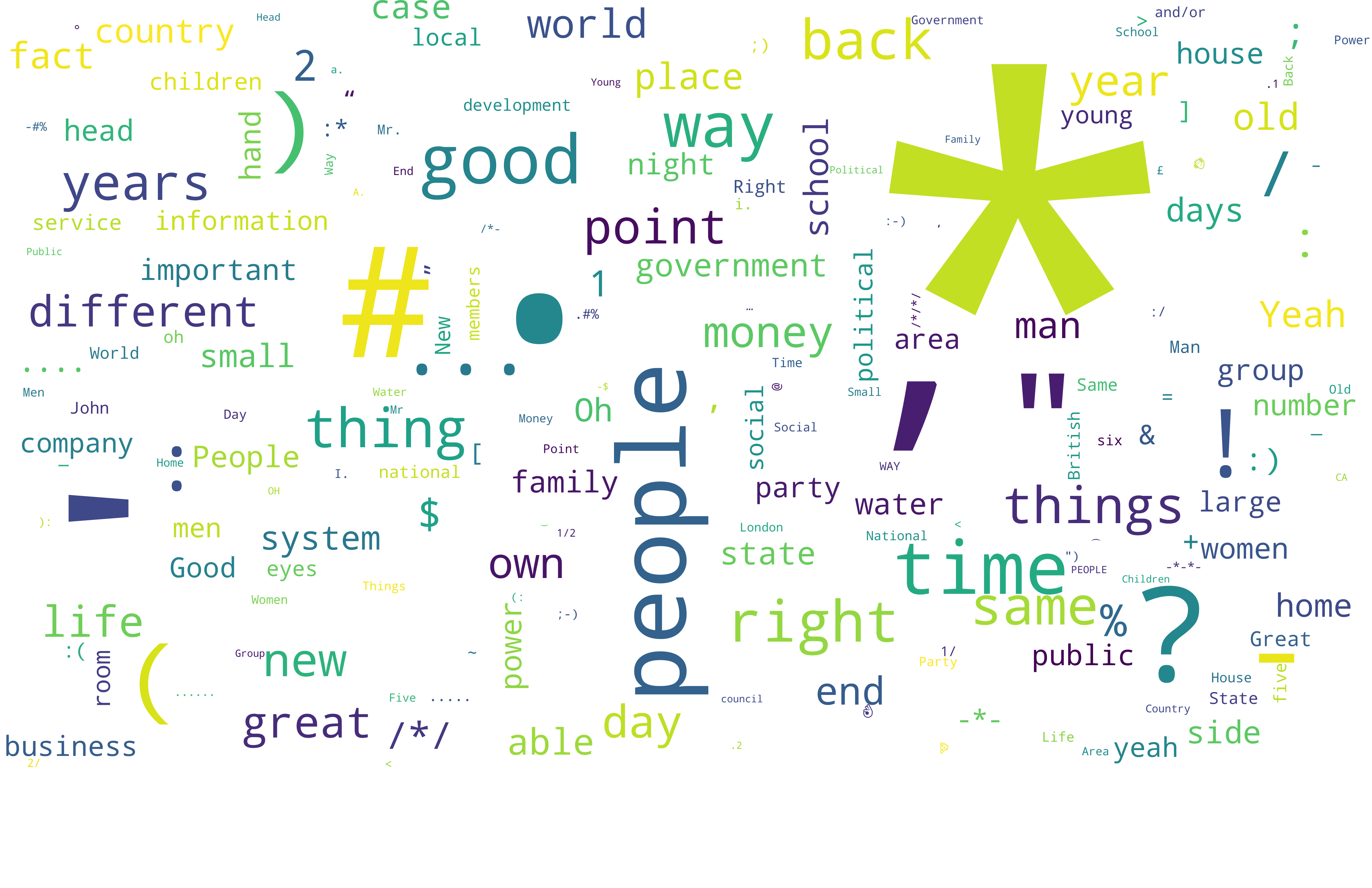} \end{minipage} \\	
		\midrule		
		500 &
		\begin{minipage}{.142\textwidth} \includegraphics[width=0.9\textwidth,trim=0cm 3.4cm 0cm 0cm,clip]{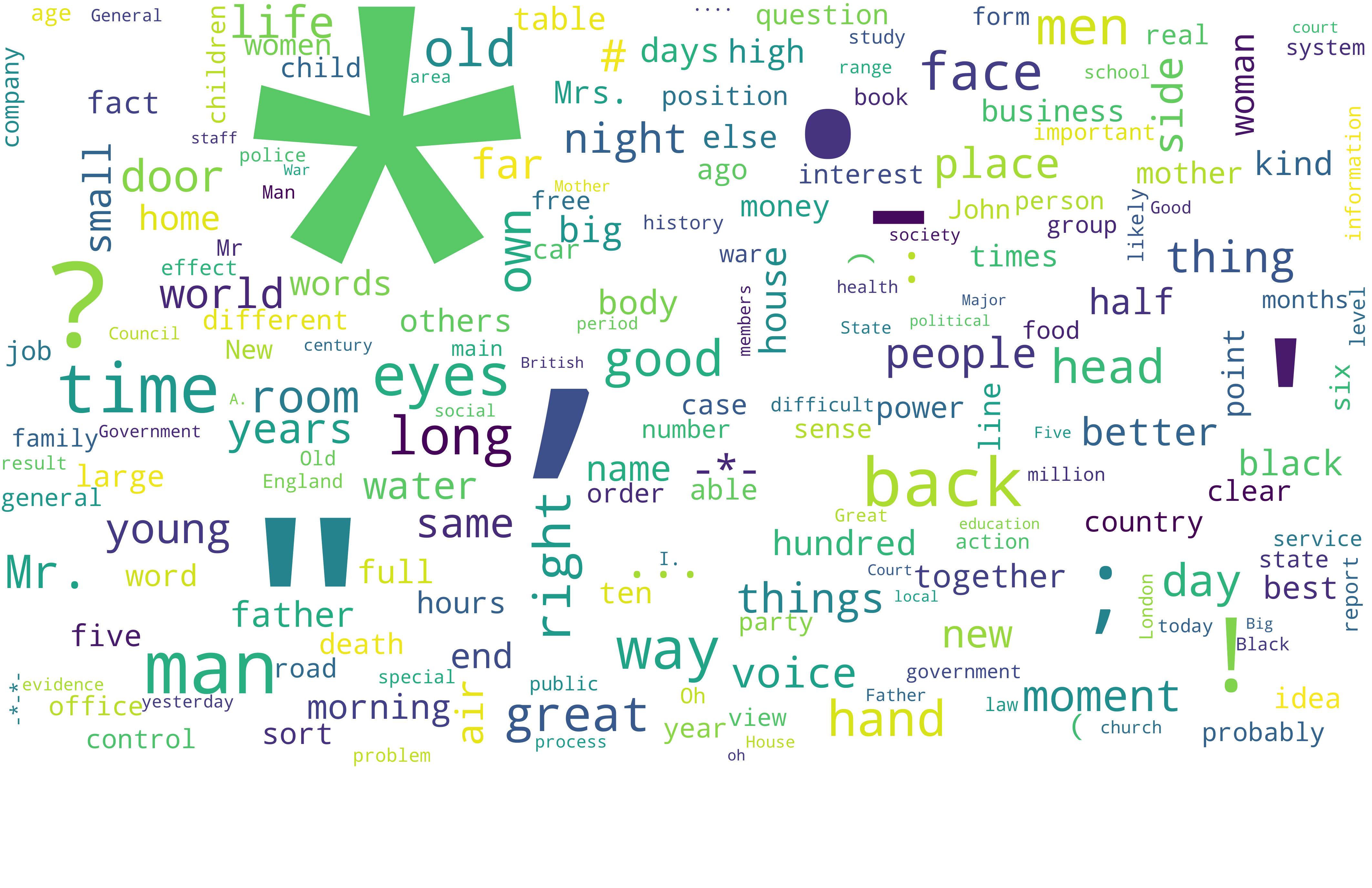} \end{minipage}  &
		\begin{minipage}{.142\textwidth} \includegraphics[width=0.9\textwidth,trim=0cm 3.4cm 0cm 0cm,clip]{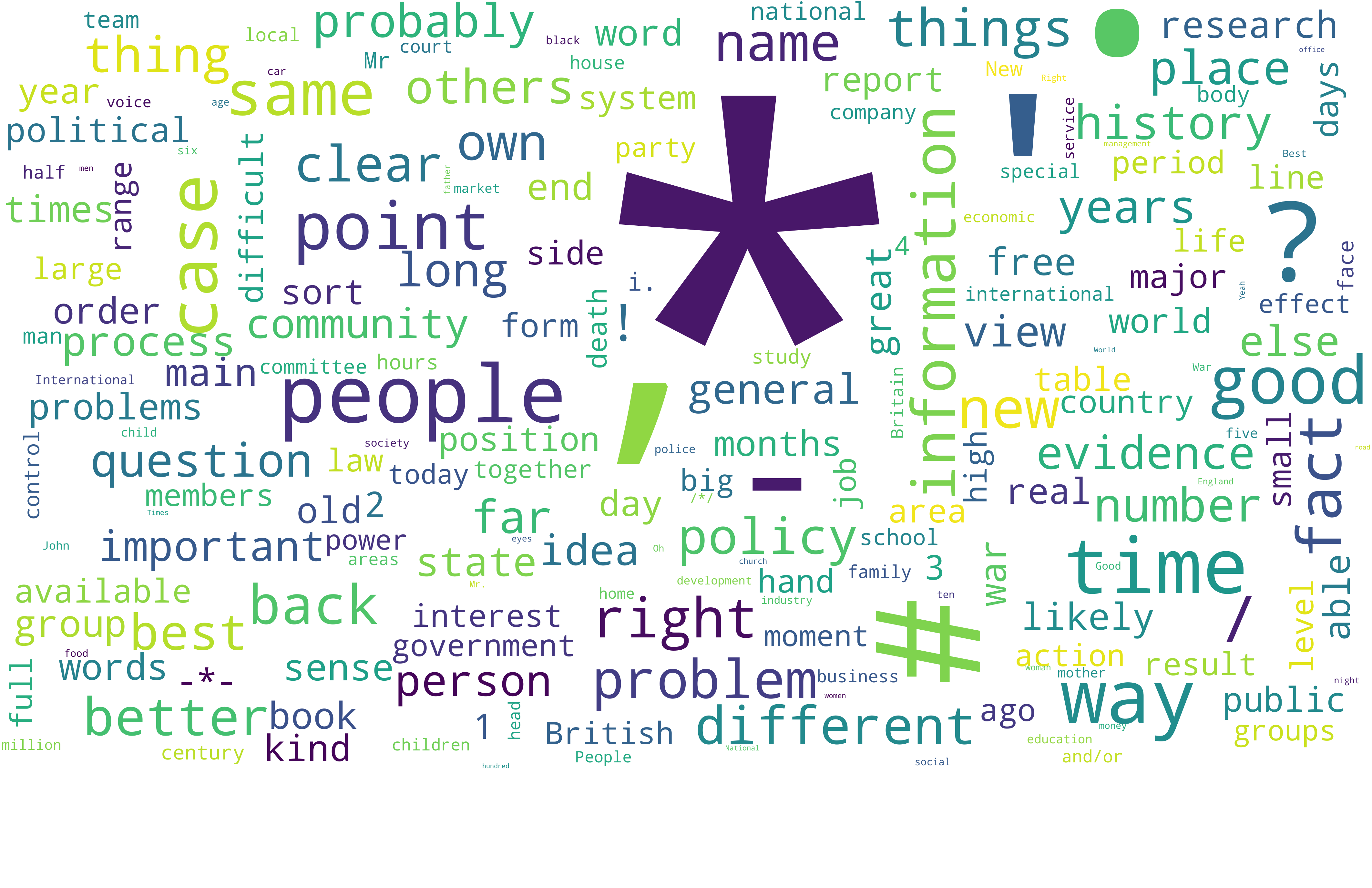} \end{minipage}  &	
		\begin{minipage}{.142\textwidth} \includegraphics[width=0.9\textwidth,trim=0cm 3.4cm 0cm 0cm,clip]{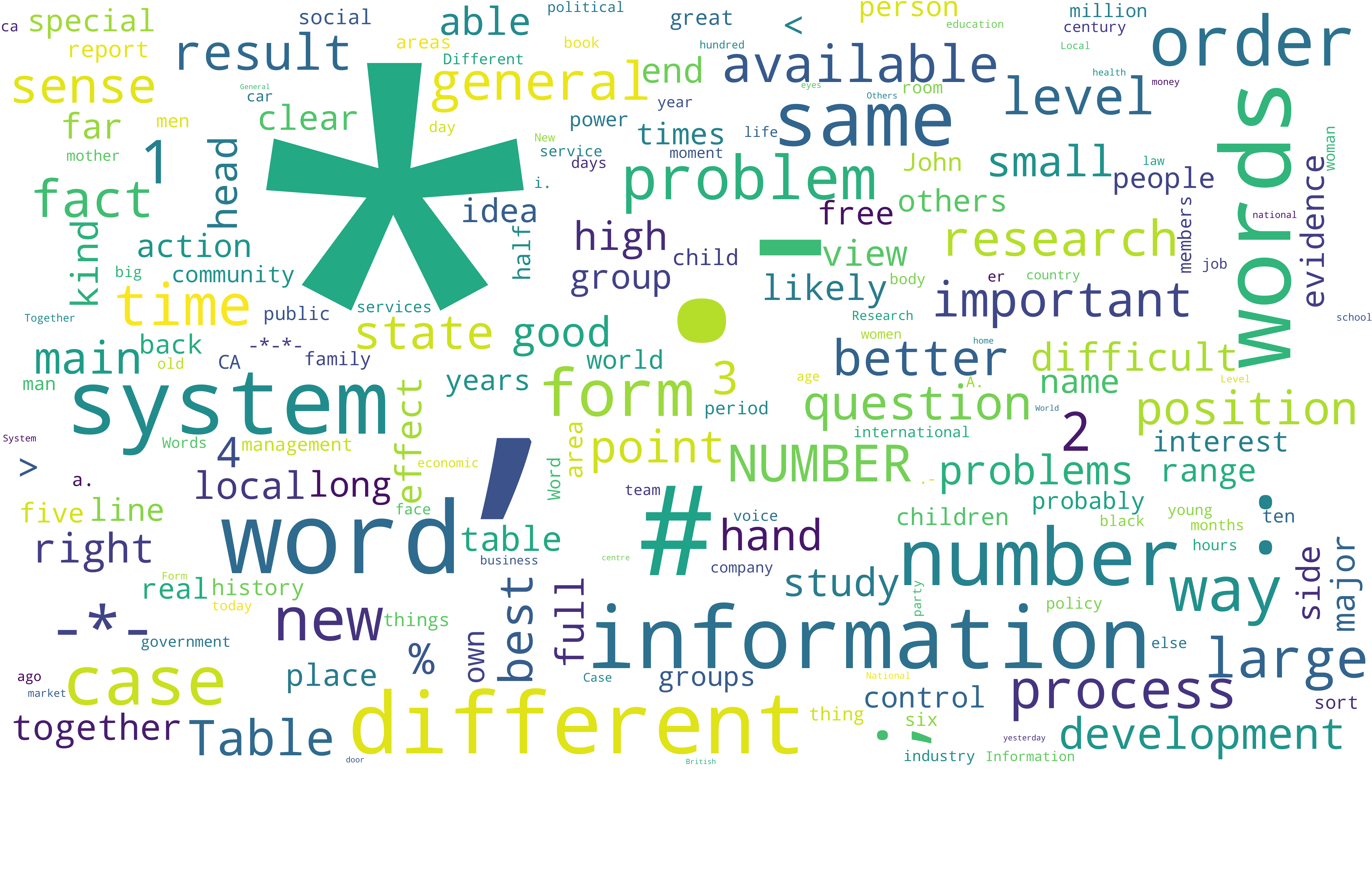} \end{minipage}  &
		\begin{minipage}{.142\textwidth} \includegraphics[width=0.9\textwidth,trim=0cm 3.4cm 0cm 0cm,clip]{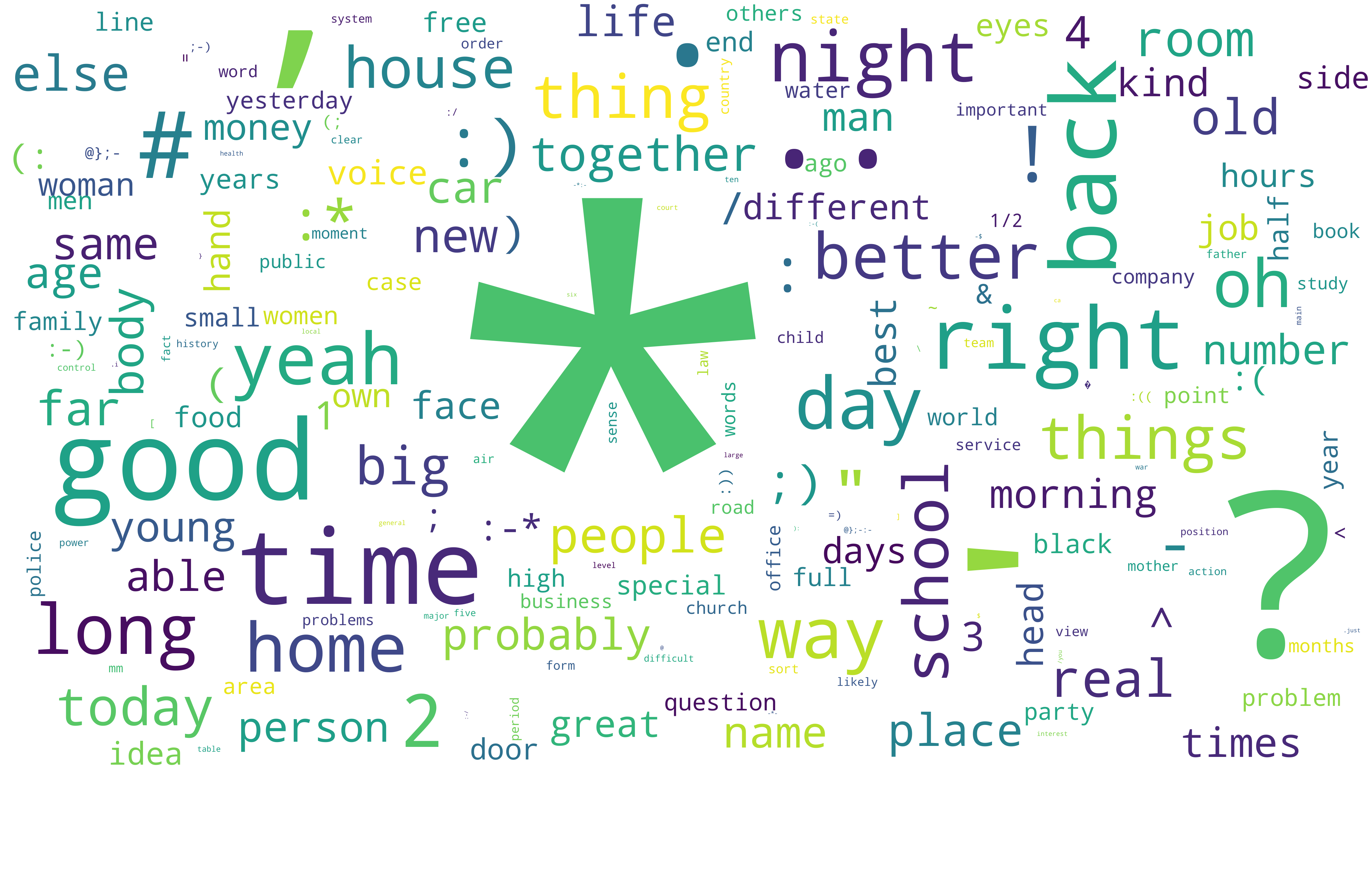} \end{minipage}  &         
		\begin{minipage}{.142\textwidth} \includegraphics[width=0.9\textwidth,trim=0cm 3.4cm 0cm 0cm,clip]{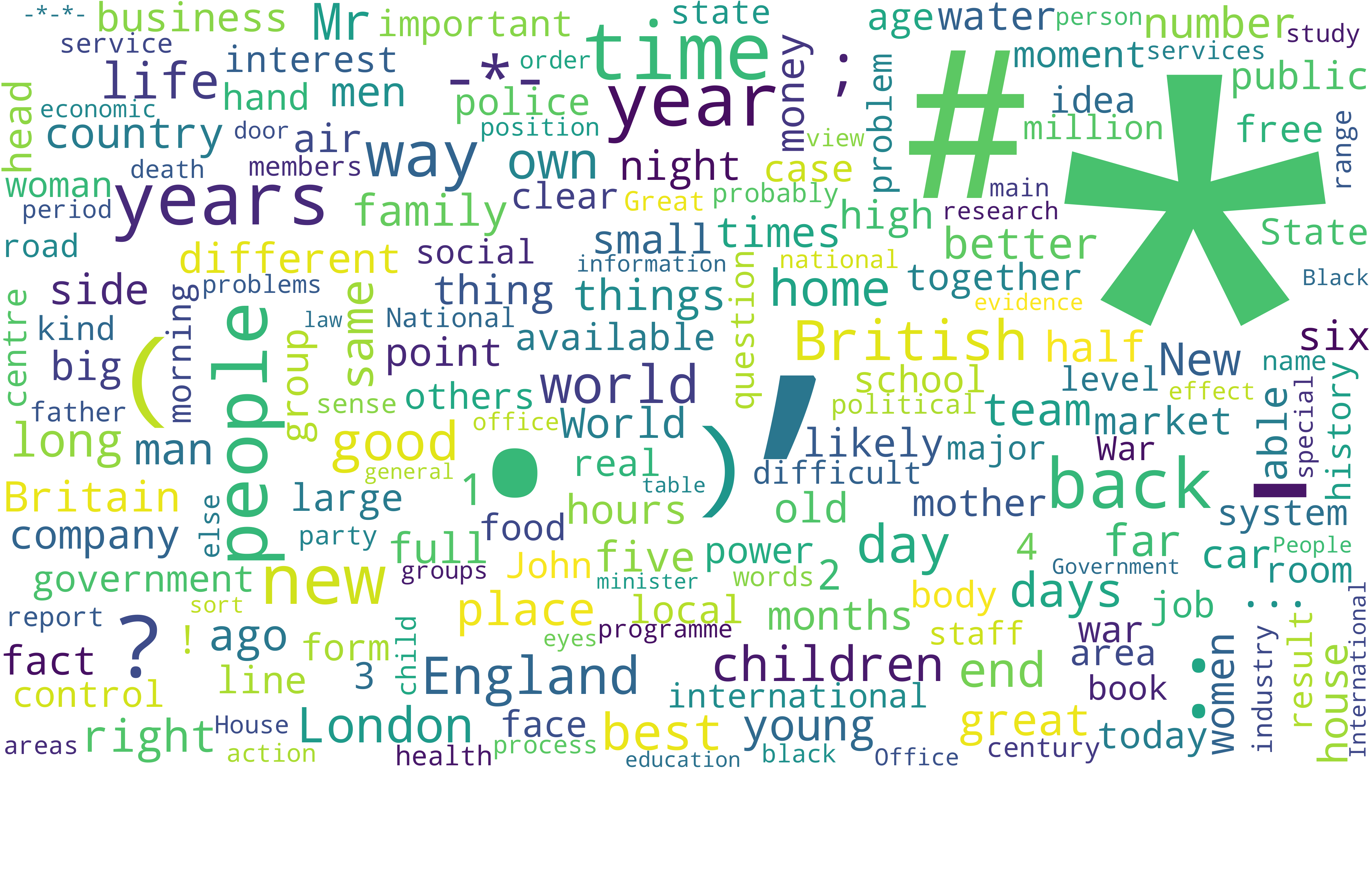} \end{minipage} & 
		\begin{minipage}{.142\textwidth} \includegraphics[width=0.9\textwidth,trim=0cm 3.4cm 0cm 0cm,clip]{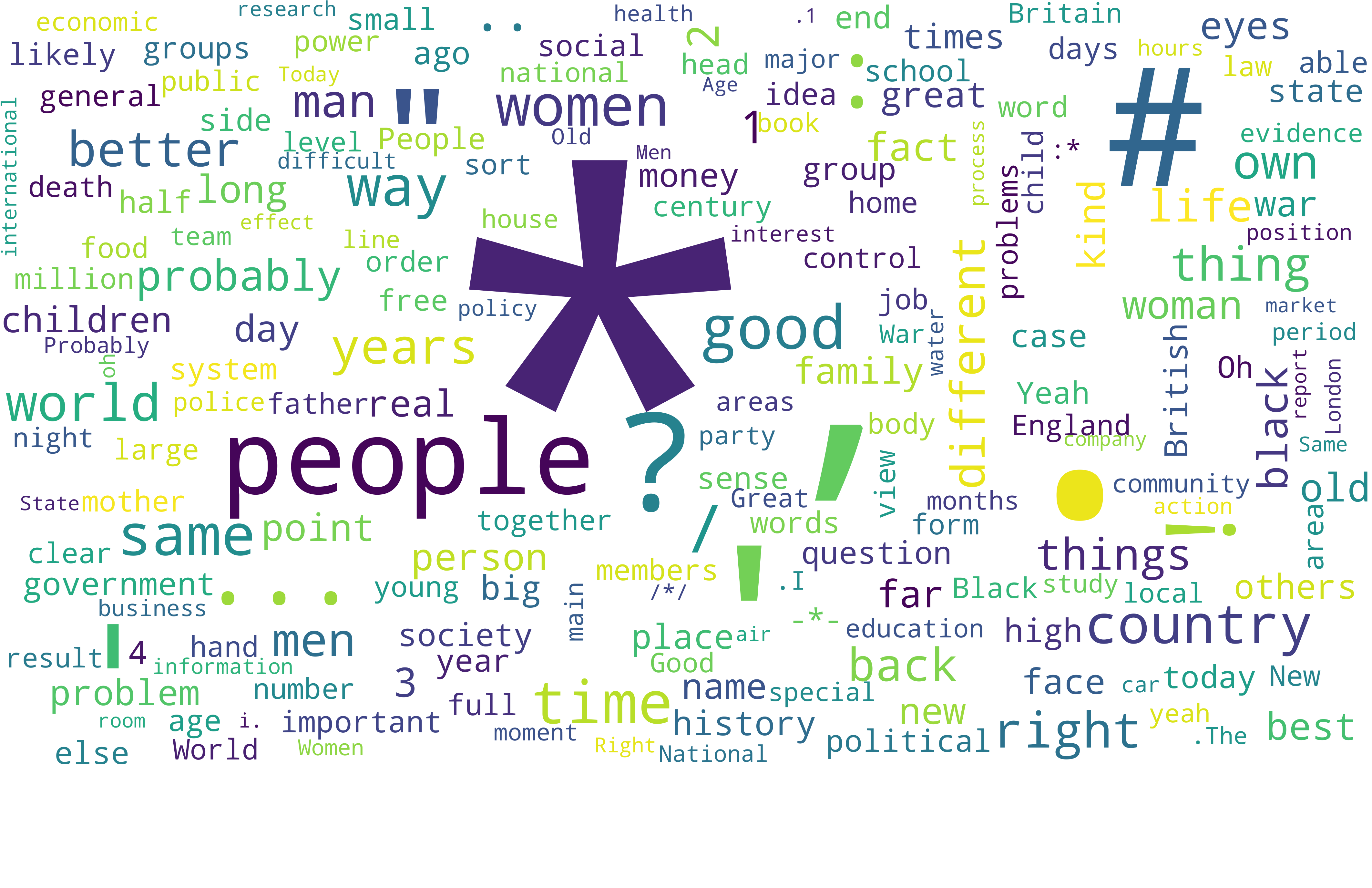} \end{minipage} & 
		\begin{minipage}{.142\textwidth} \includegraphics[width=0.9\textwidth,trim=0cm 3.4cm 0cm 0cm,clip]{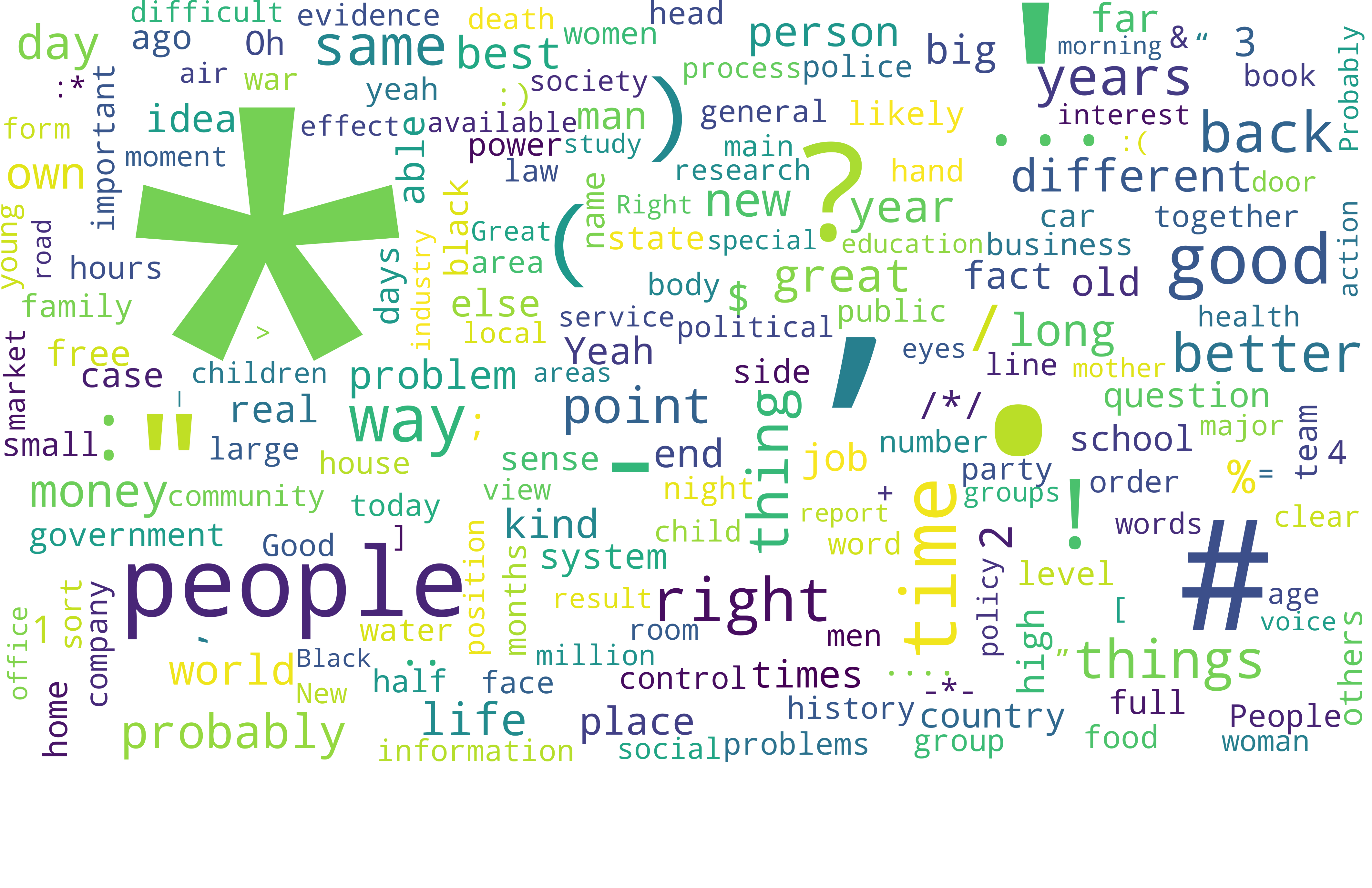} \end{minipage} 	\\		
		\midrule		
		1000 &
		\begin{minipage}{.142\textwidth} \includegraphics[width=0.9\textwidth,trim=0cm 3.4cm 0cm 0cm,clip]{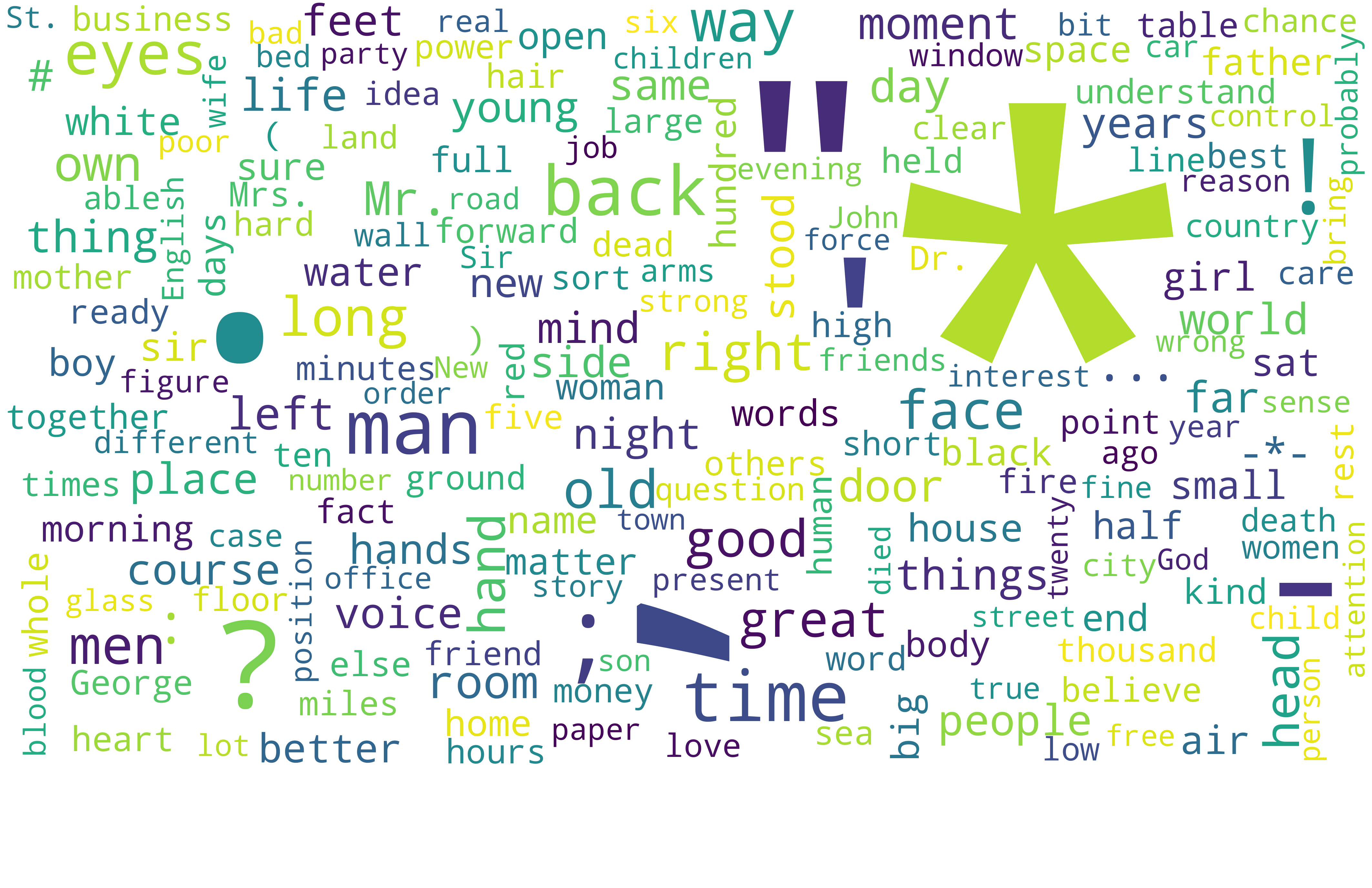} \end{minipage}  &
		\begin{minipage}{.142\textwidth} \includegraphics[width=0.9\textwidth,trim=0cm 3.4cm 0cm 0cm,clip]{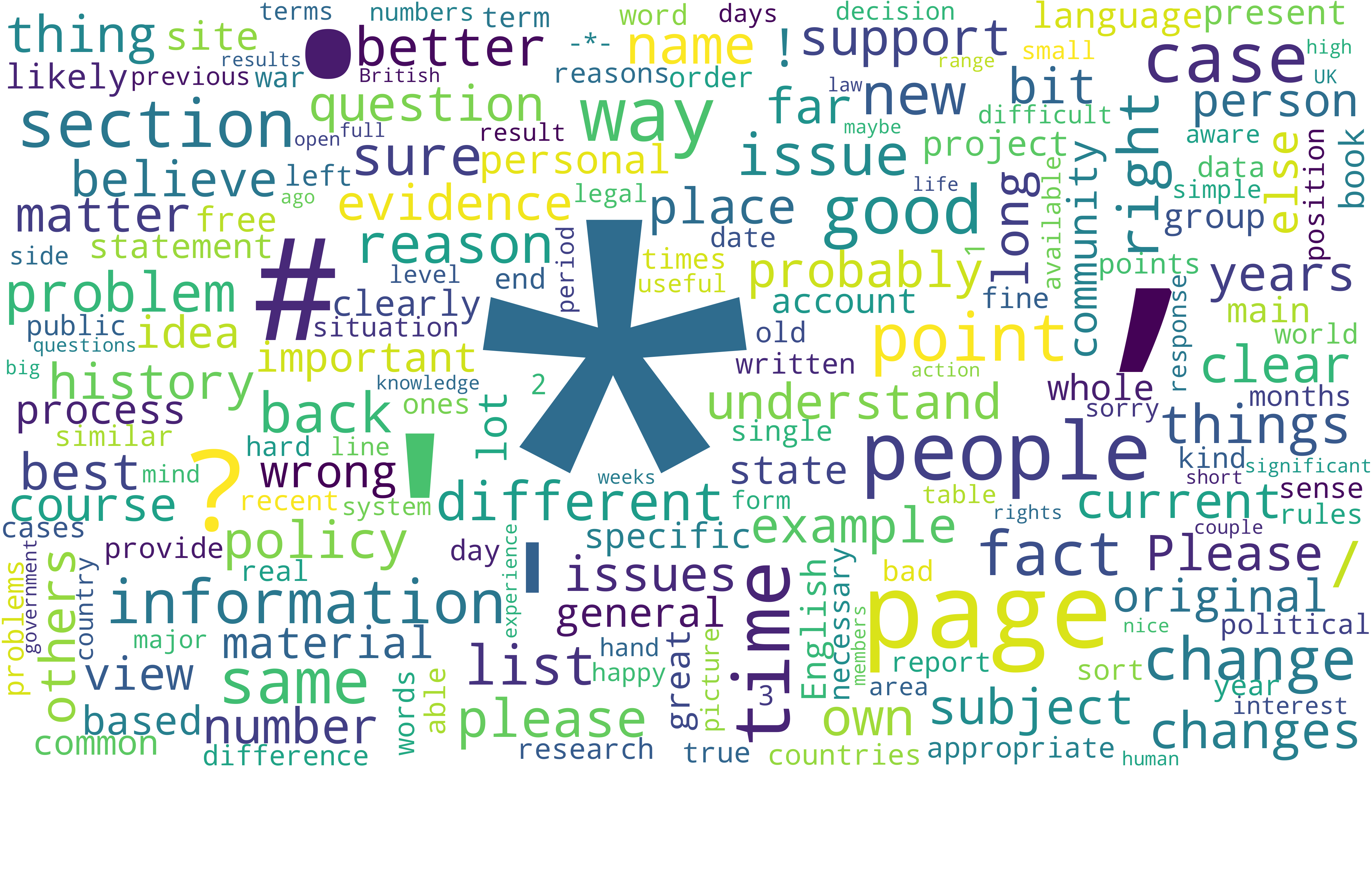} \end{minipage}  &	
		\begin{minipage}{.142\textwidth} \includegraphics[width=0.9\textwidth,trim=0cm 3.4cm 0cm 0cm,clip]{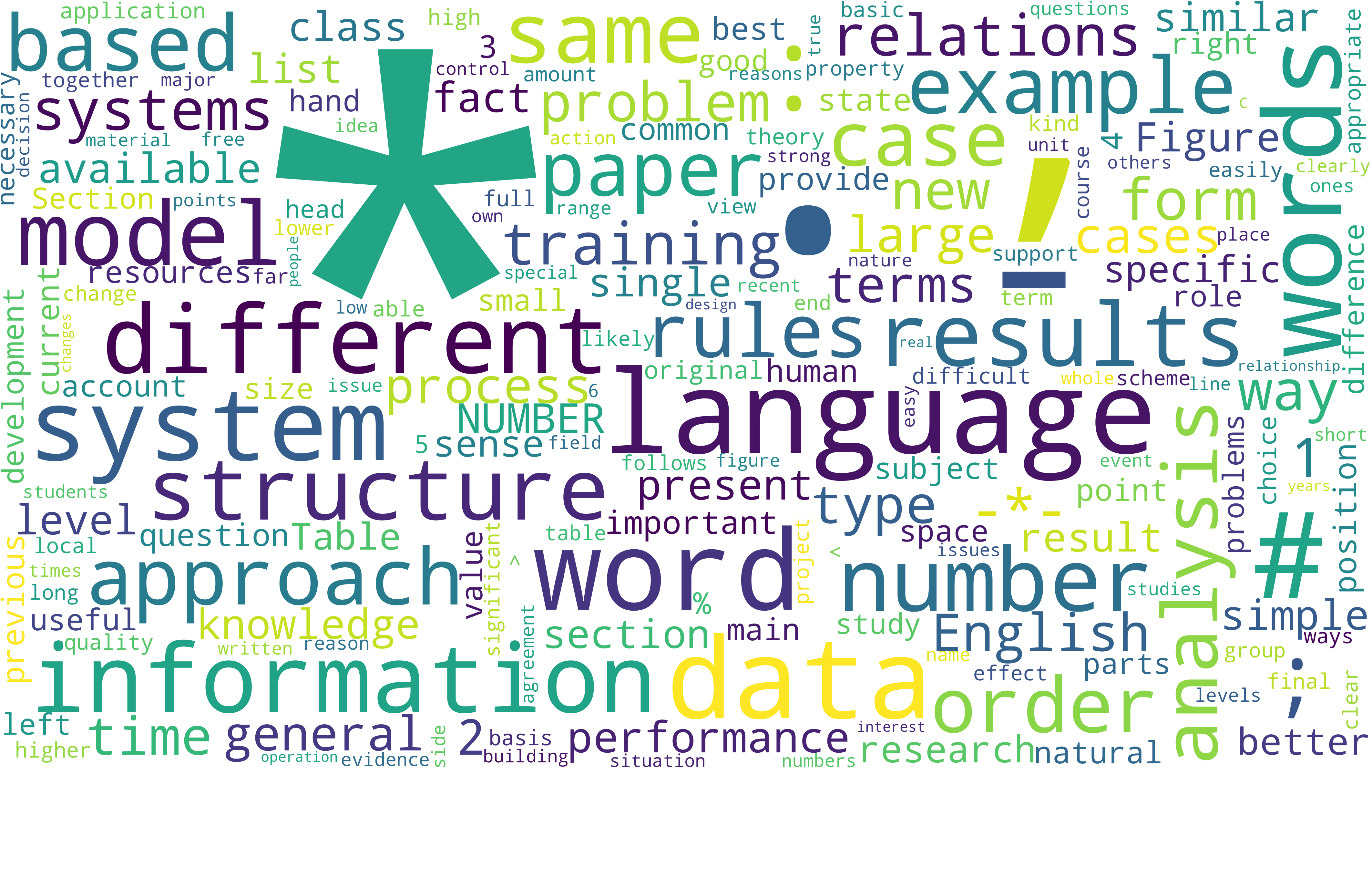} \end{minipage}  &
		\begin{minipage}{.142\textwidth} \includegraphics[width=0.9\textwidth,trim=0cm 3.4cm 0cm 0cm,clip]{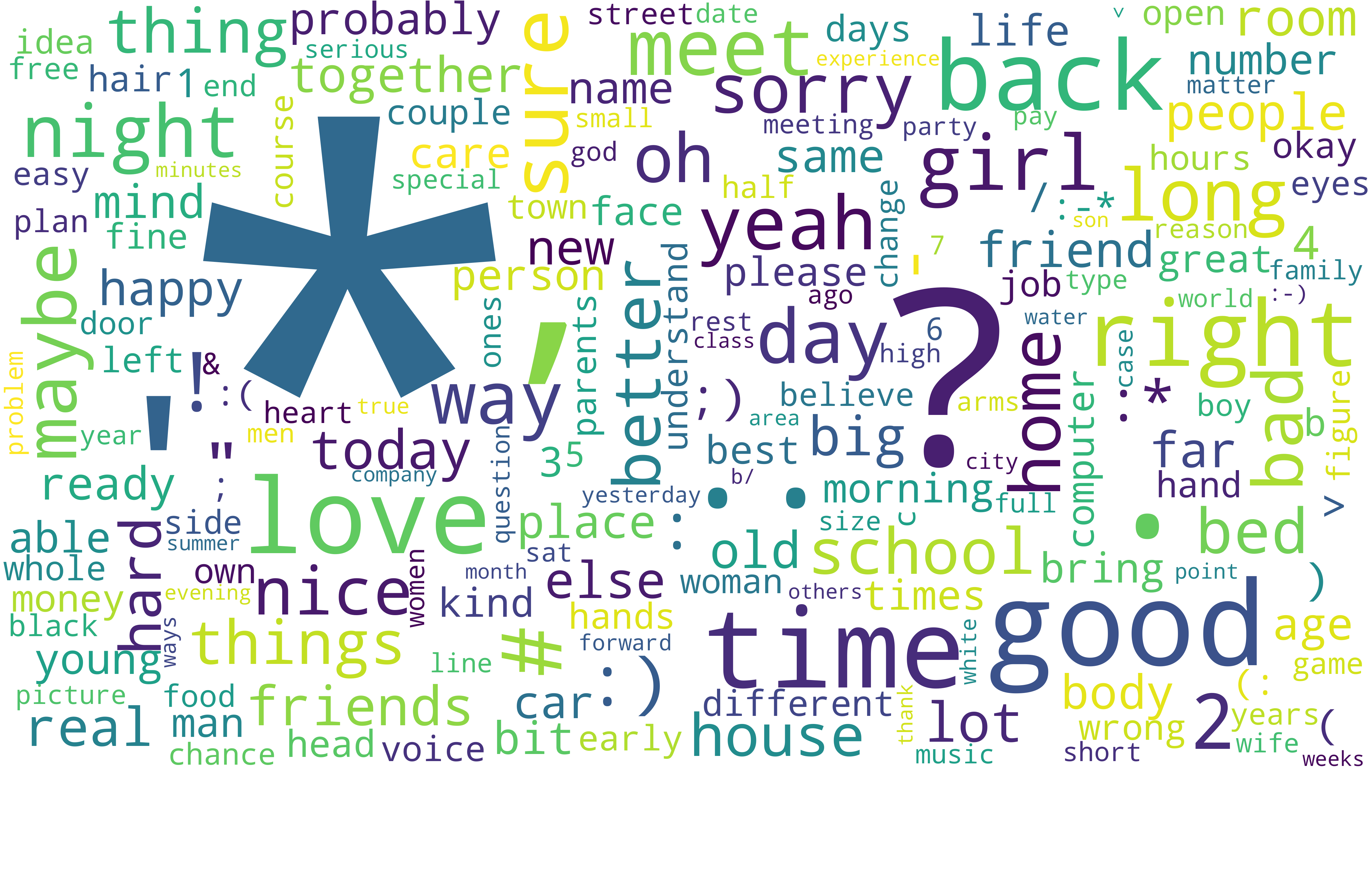} \end{minipage}  &         
		\begin{minipage}{.142\textwidth} \includegraphics[width=0.9\textwidth,trim=0cm 3.4cm 0cm 0cm,clip]{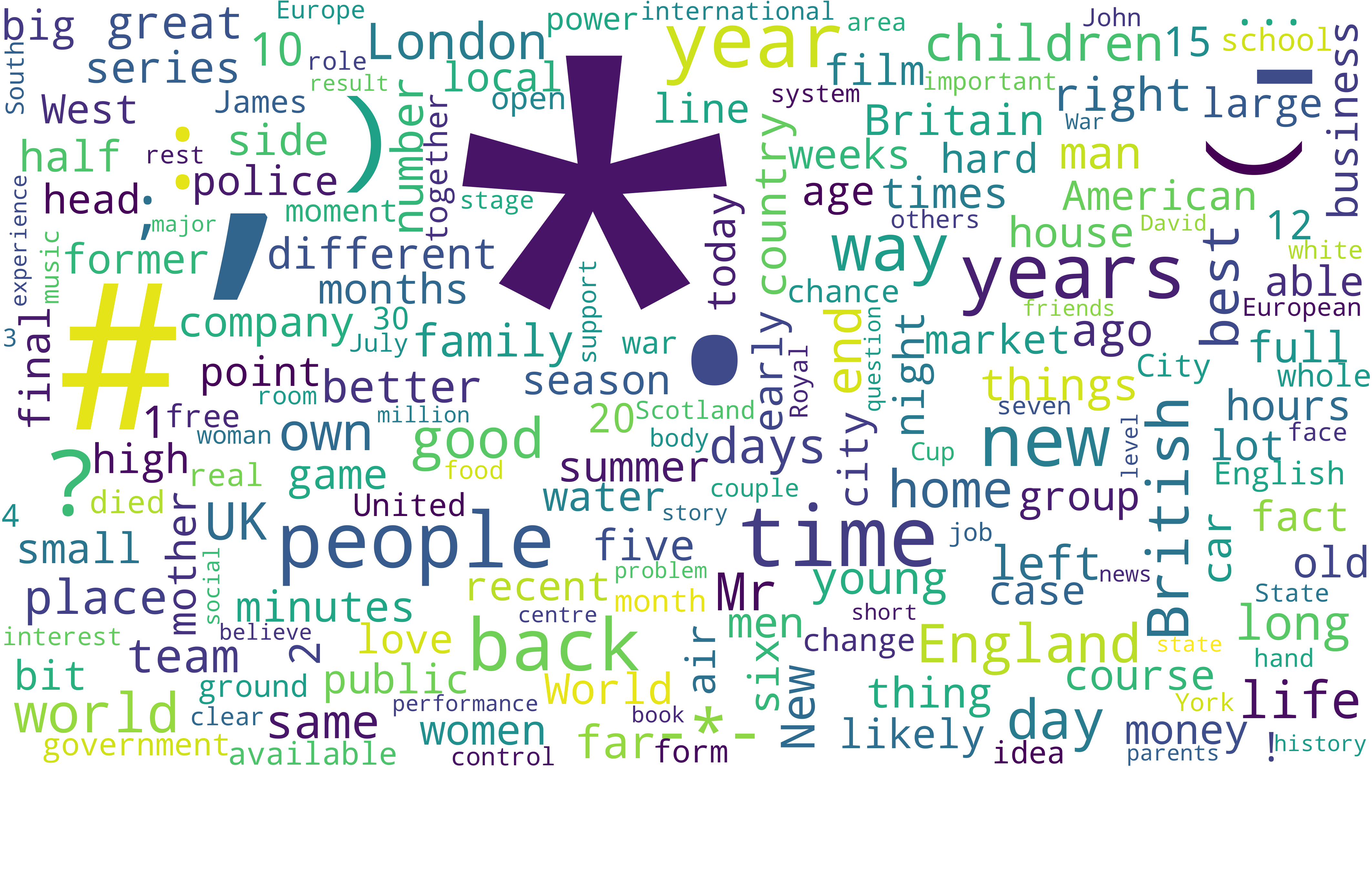} \end{minipage} & 
		\begin{minipage}{.142\textwidth} \includegraphics[width=0.9\textwidth,trim=0cm 3.4cm 0cm 0cm,clip]{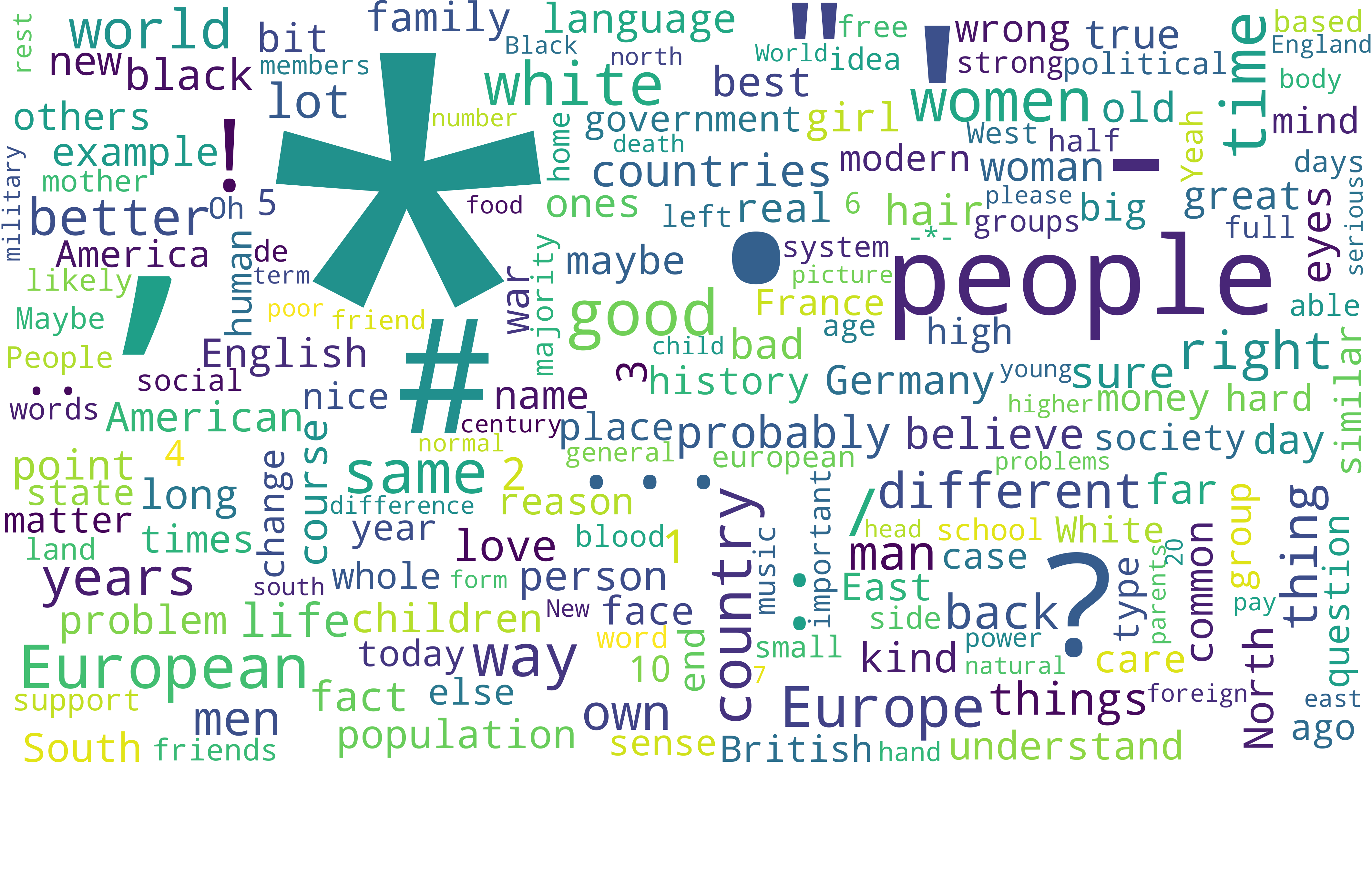} \end{minipage} & 
		\begin{minipage}{.142\textwidth} \includegraphics[width=0.9\textwidth,trim=0cm 3.4cm 0cm 0cm,clip]{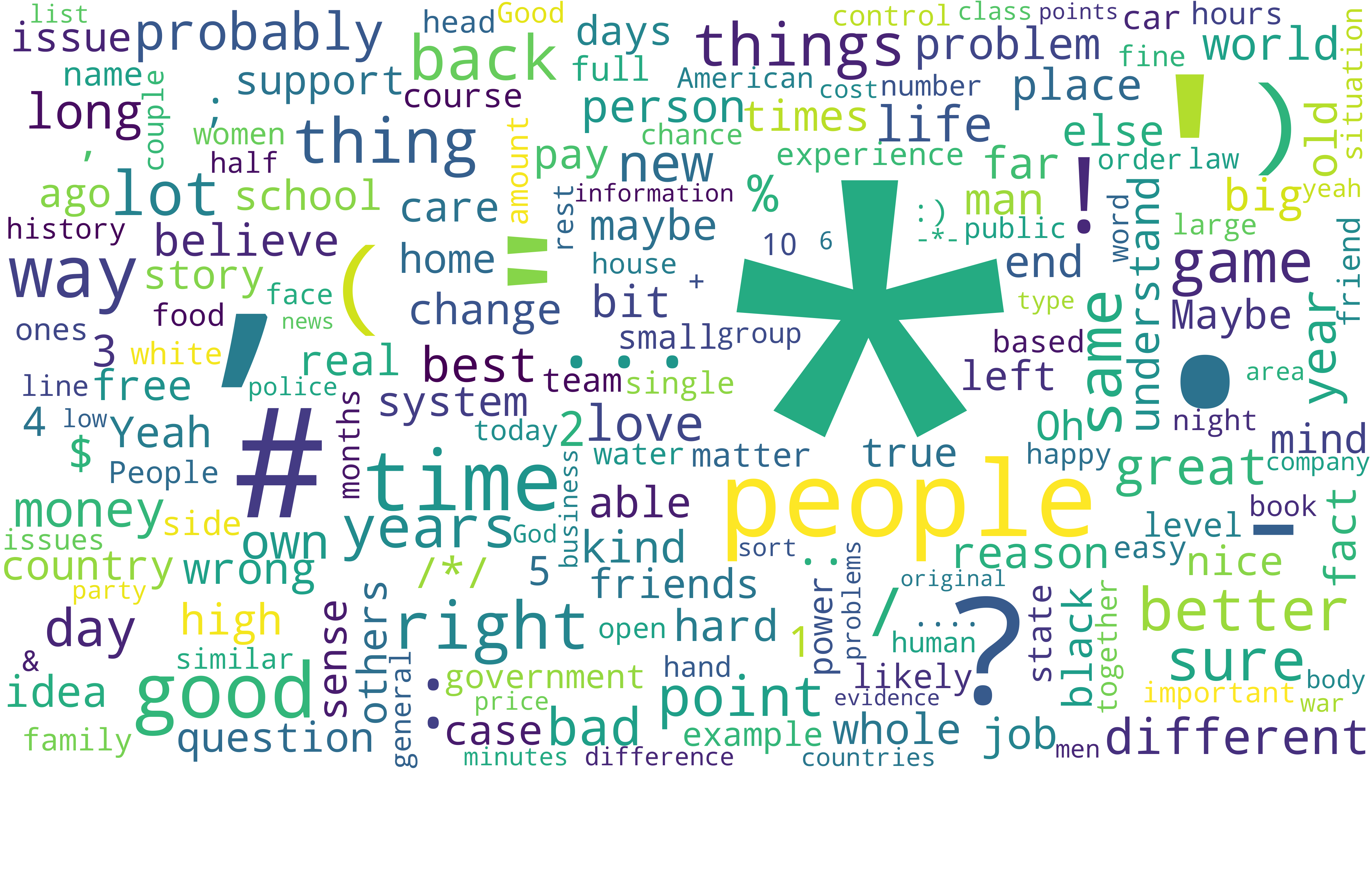} \end{minipage} \\	
		\bottomrule		
	\end{tabularx}
	\caption{Influence of the topic regularization hyperparameter $\bm{k}$ across all documents in each \textDistortion corpus. The word clouds illustrate those text units remaining in the corpora subtracted from all words and phrases present in $\listStylePatterns$. \label{tbl:TextDistortion_Influence_of_K}} 
\end{table*} 

\begin{table} 
	\centering\small 
	\begin{tabular}{l cccccc} \toprule 
		& \textbf{\coav}  & \textbf{\occav} & \textbf{\veenmanNNCD} & \textbf{\stamatatosProf} & \textbf{\spatium} & \textbf{\koppelUnmask} \\\midrule			
		\textbf{Average} & 3.74\% & 1.88\% & 4.30\% & 2.33\% & 1.80\% & 4.94\% \\ 
		\textbf{Median}  & 3.53\% & 3.33\% & 4.64\% & 2.11\% & 0.88\% & 6.41\% \\         
		\bottomrule	
	\end{tabular}
	\caption{Method-wise accuracy improvements with respect to \posNoise compared to \textDistortion for all test corpora.  \label{table:ImprovementsSummary}}
\end{table} 
\setlength{\aboverulesep}{0pt}
\setlength{\belowrulesep}{0pt}
\setlength{\extrarowheight}{.15ex}
\definecolor{white}{HTML}{ffffff}
\definecolor{lightyellow}{HTML}{fff7bc}
\definecolor{lightgray}{HTML}{f0f0f0}
\newcolumntype{y}{>{\columncolor{lightyellow}}r}
\newcolumntype{g}{>{\columncolor{lightgray}}r}
\begin{table*} [h!]	
	\centering\small 
	\begin{tabular}{p{0.21cm} l rr@{\hskip 0.25cm} yr@{\hskip 0.25cm} yr@{\hskip 0.25cm} gg@{\hskip 0.25cm} gg@{\hskip 0.25cm} gg@{\hskip 0.25cm} gg} \toprule
		&  & \multicolumn{2}{l}{\multirow{2}{*}{$\bm{\mathsf{Original}}$}} &  \multicolumn{2}{l}{\cellcolor{white}\multirow{2}{*}{$\bm{\mathsf{POSNoise}}$}}  &  \multicolumn{10}{c}{$\bm{\mathsf{TextDistortion}}$} \\[0.04cm] \cline{7-16} 
		&  & & & \multicolumn{2}{l}{} &  \multicolumn{2}{l}{$\bm{k=170}$} & \multicolumn{2}{l}{$\bm{k=100}$} & \multicolumn{2}{l}{$\bm{k=300}$} & \multicolumn{2}{l}{$\bm{k=500}$} & \multicolumn{2}{c}{$\bm{k=1000}$} \\[0.09cm]
		&  \textbf{Method} & \textbf{Acc.} & \textbf{AUC} & \textbf{Acc.} & \textbf{AUC} & \textbf{Acc.} & \textbf{AUC} & \textbf{Acc.} & \textbf{AUC} & \textbf{Acc.} & \textbf{AUC} & \textbf{Acc.} & \textbf{AUC} & \textbf{Acc.} & \textbf{AUC}     \\ \midrule		
		& \coav           & 0.744 & 0.853 & 0.731 & 0.828 & 0.731 & \underline{0.850} & 0.737 & 0.830 & 0.737 & 0.857 & 0.750 & 0.861 & 0.769 & 0.868 \\
		& \occav          & 0.500 & 0.754 & 0.513 & 0.780 & \textbf{0.519} & 0.772 & 0.513 & 0.742 & 0.519 & 0.782 & 0.699 & 0.785 & 0.532 & 0.804 \\
		& \veenmanNNCD    & 0.769 & 0.843 & \textbf{0.712} & 0.815 & 0.692 & 0.825 & 0.641 & 0.809 & 0.641 & 0.839 & 0.679 & 0.842 & 0.679 & 0.845 \\
		& \stamatatosProf & 0.718 & 0.788 & 0.686 & \underline{0.815} & 0.686 & 0.735 & 0.724 & 0.794 & 0.705 & 0.803 & 0.679 & 0.790 & 0.699 & 0.781 \\
		& \spatium        & 0.705 & 0.757 & \textbf{0.731} & 0.773 & 0.692 & 0.740 & 0.692 & 0.774 & 0.699 & 0.775 & 0.699 & 0.781 & 0.705 & 0.759 \\
		\multirow{-6}{*}{\large\rotatebox{90}{$\bm{\CorpusGutenberg}$}}
		& \koppelUnmask & 0.667 & 0.763 & \textbf{0.744} & 0.800 & 0.679 & 0.745 & 0.679 & 0.770 & 0.660 & 0.770 & 0.724 & 0.778 & 0.686 & 0.732 \\\midrule
		& \coav           & 0.898 & 0.949 & \textbf{0.823} & 0.908 & 0.770 & 0.857 & 0.726 & 0.814 & 0.801 & 0.891 & 0.805 & 0.898 & 0.836 & 0.918 \\
		& \occav          & 0.783 & 0.864 & \textbf{0.770} & 0.834 & 0.730 & 0.805 & 0.695 & 0.778 & 0.704 & 0.807 & 0.765 & 0.822 & 0.748 & 0.829 \\
		& \veenmanNNCD    & 0.832 & 0.965 & \textbf{0.770} & 0.957 & 0.704 & 0.927 & 0.690 & 0.937 & 0.748 & 0.954 & 0.717 & 0.954 & 0.770 & 0.965 \\
		& \stamatatosProf & 0.774 & 0.869 & 0.735 & 0.827 & \textbf{0.739} & 0.809 & 0.726 & 0.792 & 0.752 & 0.821 & 0.774 & 0.856 & 0.783 & 0.872 \\
		& \spatium        & 0.770 & 0.851 & \textbf{0.783} & 0.849 & 0.774 & 0.857 & 0.730 & 0.827 & 0.774 & 0.854 & 0.783 & 0.871 & 0.770 & 0.857 \\
		\multirow{-6}{*}{\large\rotatebox{90}{$\bm{\CorpusWikiSockpuppets}$}}
		& \koppelUnmask   & 0.708 & 0.793 & \textbf{0.743} & 0.844 & 0.721 & 0.790 & 0.704 & 0.749 & 0.708 & 0.821 & 0.726 & 0.809 & 0.712 & 0.806 \\\midrule
		& \coav           & 0.782 & 0.883 & \textbf{0.775} & 0.844 & 0.675 & 0.762 & 0.654 & 0.713 & 0.725 & 0.799 & 0.764 & 0.841 & 0.768 & 0.844 \\
		& \occav          & 0.500 & 0.713 & 0.500 & \underline{0.692} & 0.500 & 0.631 & 0.500 & 0.606 & 0.504 & 0.658 & 0.507 & 0.703 & 0.504 & 0.718 \\
		& \veenmanNNCD    & 0.736 & 0.996 & \textbf{0.739} & 0.994 & 0.693 & 0.996 & 0.657 & 0.997 & 0.686 & 0.995 & 0.718 & 0.995 & 0.725 & 0.994 \\
		& \stamatatosProf & 0.739 & 0.793 & \textbf{0.732} & 0.789 & 0.668 & 0.703 & 0.664 & 0.693 & 0.689 & 0.711 & 0.696 & 0.728 & 0.718 & 0.776 \\
		& \spatium        & 0.664 & 0.717 & \textbf{0.721} & 0.736 & 0.661 & 0.733 & 0.654 & 0.713 & 0.650 & 0.720 & 0.721 & 0.760 & 0.696 & 0.740 \\
		\multirow{-6}{*}{\large\rotatebox{90}{$\bm{\CorpusACL}$}}
		& \koppelUnmask   & 0.679 & 0.738 & \textbf{0.689} & 0.762 & 0.625 & 0.688 & 0.614 & 0.667 & 0.632 & 0.712 & 0.679 & 0.769 & 0.664 & 0.734 \\\midrule
		& \coav           & 0.939 & 0.988 & \textbf{0.910} & 0.970 & 0.885 & 0.936 & 0.830 & 0.917 & 0.885 & 0.946 & 0.904 & 0.960 & 0.913 & 0.975 \\
		& \occav          & 0.776 & 0.967 & 0.753 & 0.959 & \textbf{0.760} & 0.923 & 0.715 & 0.909 & 0.740 & 0.940 & 0.747 & 0.945 & 0.747 & 0.953 \\
		& \veenmanNNCD    & 0.978 & 1.000 & \textbf{0.946} & 1.000 & 0.923 & 1.000 & 0.907 & 1.000 & 0.936 & 1.000 & 0.946 & 1.000 & 0.952 & 1.000 \\
		& \stamatatosProf & 0.859 & 0.928 & \textbf{0.856} & 0.927 & 0.824 & 0.905 & 0.843 & 0.917 & 0.824 & 0.902 & 0.856 & 0.914 & 0.865 & 0.922 \\
		& \spatium        & 0.894 & 0.963 & \textbf{0.888} & 0.956 & 0.881 & 0.947 & 0.859 & 0.941 & 0.862 & 0.954 & 0.878 & 0.954 & 0.881 & 0.954 \\
		\multirow{-6}{*}{\large\rotatebox{90}{$\bm{\CorpusPeeJ}$}}
		& \koppelUnmask   & 0.910 & 0.971 & \textbf{0.885} & 0.958 & 0.798 & 0.895 & 0.824 & 0.873 & 0.827 & 0.906 & 0.824 & 0.926 & 0.856 & 0.909 \\\midrule
		& \coav           & 0.855 & 0.927 & \textbf{0.810} & 0.883 & 0.798 & 0.866 & 0.750 & 0.836 & 0.807 & 0.878 & 0.834 & 0.898 & 0.828 & 0.912 \\
		& \occav          & 0.732 & 0.828 & \textbf{0.684} & 0.790 & 0.648 & 0.752 & 0.636 & 0.744 & 0.678 & 0.768 & 0.699 & 0.785 & 0.732 & 0.814 \\
		& \veenmanNNCD    & 0.795 & 1.000 & \textbf{0.720} & 1.000 & 0.696 & 1.000 & 0.675 & 1.000 & 0.693 & 1.000 & 0.729 & 1.000 & 0.759 & 1.000 \\
		& \stamatatosProf & 0.795 & 0.877 & \textbf{0.735} & 0.828 & 0.714 & 0.823 & 0.717 & 0.820 & 0.717 & 0.828 & 0.747 & 0.843 & 0.768 & 0.862 \\
		& \spatium        & 0.765 & 0.856 & \textbf{0.774} & 0.853 & 0.762 & 0.849 & 0.741 & 0.825 & 0.753 & 0.850 & 0.774 & 0.862 & 0.777 & 0.867 \\
		\multirow{-6}{*}{\large\rotatebox{90}{$\bm{\CorpusTelegraph}$}}
		& \koppelUnmask   & 0.741 & 0.832 & \textbf{0.783} & 0.831 & 0.684 & 0.783 & 0.711 & 0.794 & 0.714 & 0.782 & 0.747 & 0.837 & 0.738 & 0.801 \\\midrule
		& \coav           & 0.832 & 0.921 & \textbf{0.859} & 0.924 & 0.824 & 0.906 & 0.800 & 0.891 & 0.844 & 0.910 & 0.844 & 0.909 & 0.850 & 0.914 \\
		& \occav          & 0.809 & 0.877 & \textbf{0.818} & 0.881 & 0.782 & 0.857 & 0.779 & 0.842 & 0.788 & 0.868 & 0.788 & 0.878 & 0.782 & 0.863 \\
		& \veenmanNNCD    & 0.832 & 0.998 & \textbf{0.803} & 1.000 & 0.738 & 0.999 & 0.735 & 0.998 & 0.759 & 1.000 & 0.765 & 1.000 & 0.776 & 1.000 \\
		& \stamatatosProf & 0.779 & 0.853 & \textbf{0.776} & 0.851 & 0.759 & 0.835 & 0.765 & 0.854 & 0.738 & 0.836 & 0.768 & 0.851 & 0.768 & 0.856 \\
		& \spatium        & 0.806 & 0.870 & \textbf{0.800} & 0.863 & 0.791 & 0.894 & 0.800 & 0.889 & 0.782 & 0.874 & 0.797 & 0.874 & 0.809 & 0.887 \\
		\multirow{-6}{*}{\large\rotatebox{90}{$\bm{\CorpusApricity}$}}
		& \koppelUnmask   & 0.735 & 0.831 & 0.774 & 0.858 & 0.774 & \underline{0.869} & 0.724 & 0.826 & 0.715 & 0.808 & 0.788 & 0.843 & 0.738 & 0.828 \\\midrule
		& \coav           & 0.836 & 0.909 & \textbf{0.818} & 0.876 & 0.782 & 0.836 & 0.753 & 0.823 & 0.787 & 0.851 & 0.796 & 0.867 & 0.818 & 0.892 \\
		& \occav          & 0.778 & 0.851 & \textbf{0.778} & 0.845 & 0.745 & 0.817 & 0.730 & 0.809 & 0.755 & 0.822 & 0.750 & 0.826 & 0.768 & 0.839 \\
		& \veenmanNNCD    & 0.774 & 0.999 & \textbf{0.761} & 1.000 & 0.703 & 1.000 & 0.692 & 1.000 & 0.709 & 1.000 & 0.728 & 1.000 & 0.745 & 1.000 \\
		& \stamatatosProf & 0.764 & 0.821 & \textbf{0.769} & 0.835 & 0.737 & 0.799 & 0.756 & 0.828 & 0.753 & 0.815 & 0.766 & 0.828 & 0.753 & 0.828 \\
		& \spatium        & 0.797 & 0.863 & 0.809 & 0.873 & \textbf{0.818} & 0.880 & 0.826 & 0.892 & 0.823 & 0.881 & 0.824 & 0.886 & 0.813 & 0.884 \\
		\multirow{-6}{*}{\large\rotatebox{90}{$\bm{\CorpusReddit}$}}
		& \koppelUnmask   & 0.719 & 0.785 & \textbf{0.731} & 0.796 & 0.722 & 0.797 & 0.716 & 0.801 & 0.738 & 0.815 & 0.758 & 0.822 & 0.755 & 0.812 \\\bottomrule		
	\end{tabular} 
	\caption{Comparison between the \numberOfExistingBaselines \AV methods applied to all \originalCorpus, \posNoise and \textDistortion test corpora. 		
		 Bold values indicate the best accuracy (\textbf{Acc.}) results with respect to the \posNoise and \textDistortion ($\bm{k=170}$) corpora. 
		 In case of ties, \textbf{\auc} serves as a secondary ranking option represented by underlined values. \label{tab:ExperimentComparisonResults}}
\end{table*}
\section{Conclusion and Future Work} \label{Conclusions}
We discussed a serious problem in \av, which affects many \AV methods that have no control over the features they capture. 
As a result, the classification predictions of the respective \AV methods may be biased by the topic of the investigated documents. 
To address this problem, we proposed a simple but effective approach called \posNoise, which aims to mask topic-related text units in documents. 
In this way, only those features in the documents that relate to the authors' writing style are retained, so that the actual goal of an \AV method can be achieved.  
In contrast to the alternative topic masking technique \textDistortion, our approach follows a two-step strategy to mask topic-related content in a given document $\D$. 
The idea behind \posNoise is first to substitute topic-related words with \posTags predefined in a set $\mathcal{S}$. 
In a second step, a predefined list $\mathcal{L}$ is used to retain certain categories of stylistically relevant words and phrases that occur in $\D$. 
In addition to this list, we also retain all remaining words in $\D$, for which their corresponding \posTags are not contained in $\mathcal{S}$. 
The result of this procedure is a topic-agnostic document representation that allows \AV methods to better quantify stylistically relevant features. 
Besides a POS tagger and a predefined list $\mathcal{L}$, no further linguistic resources are required. 
In particular, there is no hyperparameter that requires careful adjustment, as it is the case with the \textDistortion approach.  
To assess our approach, we performed a comprehensive evaluation with six existing \AV approaches applied to seven test corpora with related and mixed topics. 
Our results have shown that \AV methods based on \posNoise lead to better results than \textDistortion in 34 out of 42 cases with an accuracy improvement of up to 10\%. 
Also, we have shown that regardless of which $k$ is chosen for \textDistortion, our approach consistently leads to a better trade-off between style and topic. 
\\
\\
However, besides the benefits of our approach, there are also several issues that require further consideration. 
When considering languages other than English, \posNoise must be adjusted accordingly. 
First, a predefined list of topic-agnostic words and phrases must be compiled for the target language. 
Second, our approach relies on a POS tagger so that the availability of a trained model for the desired language must be ensured. 
Third, due to the imperfection of the tagger, incorrect \posTags may appear in the topic-masked representations. 
Although we have hardly noticed this issue with respect to documents written in English, it is likely to happen with documents written in other languages. 
Fourth, due to the underlying tagging process, \posNoise is slower than the existing approach \textDistortion, where the runtime depends on several factors such as the text length or the complexity of the trained model. 
Besides these issues, \posNoise leaves further room for improvement. One idea for future work, is to investigate automated possibilities to extend the compiled patterns list. 
This can be achieved, for example, by using alternative linguistic resources such as lexical databases that are also available in multiple languages (\eg WordNet and GermaNet). 
Another direction for future work is to investigate the question in which verification scenarios \posNoise is also applicable. 
One idea, for example, is to perform experiments under cross-domain \AV conditions, which often occur in real forensic cases 
(for example, how a model trained on forum posts or cooking recipes performs on suicide letters, for which no training data is available yet). 
Beyond the boundaries of \AV, we also aim to investigate the suitability and effectiveness of our approach in related disciplines of authorship analysis, such as author clustering and author diarization.
\section*{Acknowledgments}
This research work has been funded by the German Federal Ministry of Education and Research and the Hessian Ministry of Higher Education, Research, Science and the Arts within their joint support of the National Research Center for Applied Cybersecurity ATHENE.	

\clearpage
\section*{Appendix}
\subsection{Hyperparameters of Involved AV Methods}
\begin{table} [h!]
	\centering\small  
	\begin{tabular}{lrr} 
		\toprule
		\textbf{Hyperparameter}       & \textbf{Our grid search range} & \textbf{Original}  \\ \midrule
		$U_1$ = initial feature set sizes     & $\{ 5, 15, 25, 35, 50, 75, 100, 150 \}$ & 250 \\
		$U_2$ = \# eliminated features & $\{ 2, 3, 5 \}$ & 3 \\ 
		$U_3$ = \# iterations          & $\{ 3, 5, 7 \}$ & 10 \\
		$U_4$ = chunk sizes (in words) & $\{ 5, 15, 25, 35, 50, 75 \}$ & 500 \\
		$U_5$ = \# folds               & $\{ 3, 5, 7, 10 \}$ & 10 \\
		\bottomrule			
	\end{tabular}
	\caption{Original and adjusted hyperparameter ranges of the \AV method \koppelUnmask (\# stands for \quote{number of}). \label{tab:UnmaskingHyperparams}}
\end{table}
\begin{table} [h!]
	\centering\footnotesize	
	\begin{tabular}{ll rrcc l rrrrr} 
		\toprule
		$\bm{\Corpus}$ &  \textbf{Representation} & \multicolumn{4}{c}{\textbf{\stamatatosProf}} & & \multicolumn{5}{c}{\textbf{\koppelUnmask}}  \\ 
		& & $L_u$ & $L_k$ & $n$ & $d$  & & $U_1$ & $U_2$ & $U_3$ & $U_4$ & $U_5$   \\\midrule
		\multirow{3}{*}{$\CorpusGutenberg$} & \originalCorpus & 8,000 & 3,000 & 5 & $d_0$ && 150 & 5 & 5 & 5 & 10 \\
		& \posNoise       & 3,000  & 1,000  & 4 & $d_0$ & & 100 & 2 & 7 & 25 & 10   \\
		& \textDistortion & 1,000  & 2,000  & 4 & $d_0$ & & 150 & 5 & 5 & 5 & 10   \\\midrule											
		\multirow{3}{*}{$\CorpusWikiSockpuppets$} & \originalCorpus & 2,000 & 6,000 & 5 & $d_0$ && 75 & 3 & 7 & 15 & 10 \\
		& \posNoise       & 1,000 & 2,000 & 4 & $d_0$ && 100 & 3 & 3 & 15 & 10 \\
		& \textDistortion & 1,000 & 8,000 & 5 & $d_0$ && 75 & 3 & 5 & 5 & 10 \\\midrule																					
		\multirow{3}{*}{$\CorpusACL$}       & \originalCorpus & 1,000 & 4,000  & 5 & $d_0$ && 150 & 2 & 3 & 5 & 10 \\
		& \posNoise       & 2,000 & 2,000 & 5 & $d_1$ && 100 & 5 & 3 & 5 & 10 \\
		& \textDistortion & 9,000 & 8,000 & 5 & $d_0$ && 75 & 5 & 3 & 5 & 10 \\\midrule
		\multirow{3}{*}{$\CorpusPeeJ$}      & \originalCorpus & 2,000 & 3,000 & 5 & $d_0$ && 150 & 2 & 5 & 5 & 10 \\
		& \posNoise       & 1,000 & 2,000 & 4 & $d_0$ && 75 & 2 & 5 & 5 & 10 \\
		& \textDistortion & 1,000 & 2,000 & 4 & $d_0$ && 75 & 2 & 3 & 5 & 10 \\\midrule
		\multirow{3}{*}{$\CorpusTelegraph$} & \originalCorpus & 2,000 & 5,000 & 5 & $d_0$ && 100 & 5 & 7 & 25 & 10 \\
		& \posNoise       & 10,000 & 15,000 & 5 & $d_0$ && 150 & 3 & 7 & 5 & 10 \\
		& \textDistortion & 6,000 & 10,000 & 5 & $d_0$ && 100 & 2 & 3 & 5 & 10 \\\midrule
		\multirow{3}{*}{$\CorpusApricity$}  & \originalCorpus & 7,000 & 5,000 & 5 & $d_0$ && 75 & 5 & 5 & 50 & 10 \\
		& \posNoise       & 1,000 & 1,000 & 4 & $d_0$ && 150 & 5 & 7 & 5 & 10 \\
		& \textDistortion & 1,000 & 3,000 & 5 & $d_0$ && 100 & 3 & 5 & 50 & 10 \\\midrule
		\multirow{3}{*}{$\CorpusReddit$}    & \originalCorpus & 7,000 & 9,000 & 5 & $d_0$ && 50 & 3 & 5 & 5 & 10 \\
		& \posNoise       & 2,000 & 4,000 & 5 & $d_0$ && 35 & 2 & 3 & 5 & 10 \\
		& \textDistortion & 1,000 & 2,000 & 4 & $d_0$ && 35 & 2 & 3 & 5 & 10 \\																
		\bottomrule			
	\end{tabular}
	\caption{Hyperparameters of \stamatatosProf and \koppelUnmask. The hyperparameters of \stamatatosProf have the following notation: $L_u$ = profile size of the unknown document, $L_k$ = profile size of the known document, $n$ = order of \charNgrams and $d$ = dissimilarity function. The definitions of the three dissimilarity functions $d_0$, $d_1$ and \e{SPI} are described in detail in \cite[Section~3.1]{StamatatosProfileCNG:2014}. The five hyperparameters $U_1\,$--$\,U_5$ of \koppelUnmask are described in Table~\ref{tab:UnmaskingHyperparams}. \label{tab:BaselinesHyperparameters}}
\end{table} 

\clearpage
\bibliographystyle{plain}
\bibliography{Bibliography}
\end{document}